\documentclass[10pt,twocolumn,letterpaper]{article}

\usepackage{iccv}
\usepackage{times}
\usepackage{epsfig}

\usepackage{graphicx}
\usepackage{amsmath}
\usepackage{amssymb}
\usepackage{booktabs}
\usepackage{graphicx}
\usepackage{wrapfig}
\usepackage{makecell}
\usepackage[dvipsnames, svgnames, x11names]{xcolor}        
\usepackage{hyperref}
\hypersetup{pagebackref,breaklinks,colorlinks}
\usepackage{pifont}
\newcommand{\cmark}{\ding{51}}%
\newcommand{\xmark}{\ding{55}}%
\usepackage[font={small}]{caption}
\usepackage{mathtools, cuted}
\usepackage[misc]{ifsym}
\usepackage{mmstyle}
\usepackage{stfloats}
\definecolor{antiquefuchsia}{rgb}{0.57, 0.36, 0.51}

\newcommand{\datasetname}{DNA-Rendering}
\newcommand{\zjumocap}{ZJU-MoCap}
\newcommand{\humman}{HuMMan}
\newcommand{\genebody}{GeneBody}
\newcommand{\kptnerf}{KeypointNeRF}
\newcommand{\ibr}{IBRNet}
\newcommand{\pixel}{PixelNeRF}
\newcommand{\vision}{VisionNeRF}
\newcommand{\nhp}{NeuralHumanPerformer}

\newcommand{\ngp}{Instant-NGP}
\newcommand{\neus}{NeuS}

\newcommand{\nv}{NeuralVolumes}
\newcommand{\nb}{NeuralBody}
\newcommand{\anerf}{A-NeRF}
\newcommand{\aniN}{AnimatableNeRF}
\newcommand{\hn}{HumanNeRF}

\newcommand{\smplx}{SMPLX}
\newcommand{\twoD}{2D}
\newcommand{\threeD}{3D}

\usepackage{multirow, hhline}
\usepackage{balance}
\usepackage[ruled,vlined]{algorithm2e}
\usepackage{colortbl}
\definecolor{citecolor}{RGB}{65,105,225}
\usepackage{caption}
\usepackage{subcaption}
\usepackage{arydshln}
\makeatletter
  \renewcommand*\env@matrix[1][*\c@MaxMatrixCols c]{%
    \hskip -\arraycolsep
    \let\@ifnextchar\new@ifnextchar
  \array{#1}}
\makeatother

\iccvfinalcopy 

\usepackage[capitalize]{cleveref}
\crefname{section}{Sec.}{Secs.}
\crefname{section}{Section}{Sections}
\Crefname{table}{Table}{Tables}
\crefname{table}{Tab.}{Tabs.}

\definecolor{colorbest}{RGB}{255,179,179}
\definecolor{colorsecond}{RGB}{255,217,179}
\definecolor{colorthird}{RGB}{255,255,179}
\newcommand{\best}[0]{\cellcolor{colorbest} }
\newcommand{\second}[0]{\cellcolor{colorsecond}}
\newcommand{\third}[0]{\cellcolor{colorthird}}
\DeclareRobustCommand{\legendsquare}[1]{%
  \textcolor{#1}{\rule{2ex}{2ex}}%
}

\begin{document}

\title{\datasetname: A Diverse Neural Actor Repository for High-Fidelity Human-centric Rendering}
\author{
    Wei Cheng$^{1}$
    \quad
    Ruixiang Chen$^{2*}$
    \quad
    Wanqi Yin$^{2*}$
    \quad
    Siming Fan$^{1,2*}$
    \quad
    Keyu Chen$^{1*}$\\
    \quad
    Honglin He$^{1}$ 
    \quad
    Huiwen Luo$^{1}$ 
    \quad
    Zhongang Cai$^{3}$ 
    \quad
    Jingbo Wang$^{4}$
    \quad
    Yang Gao$^{2}$\\
    \quad
    Zhengming Yu$^{1}$
    \quad
    Zhengyu Lin$^{2}$
    \quad
    Daxuan Ren$^{3}$\\
    Lei Yang$^{1,2}$ 
    \quad
    Ziwei Liu$^{3}$
    \quad
    Chen Change Loy$^{3}$ 
    \quad
    Chen Qian$^{1}$ \\
    \quad
    Wayne Wu$^{1}$
    \quad
    Dahua Lin$^{1,4}$
    \quad
    Bo Dai$^{1\dagger}$
    \quad
    Kwan-Yee Lin$^{1,4\dagger}$ \\
    $^{1}$ Shanghai AI Laboratory
    \quad
    $^{2}$ SenseTime Research 
    \quad
    $^{3}$ S-Lab, NTU
    \quad
    $^{4}$ CUHK
}

\twocolumn[{
  \renewcommand\twocolumn[1][]{#1}
  \maketitle
  \begin{center}
  \includegraphics[width=0.95\textwidth]{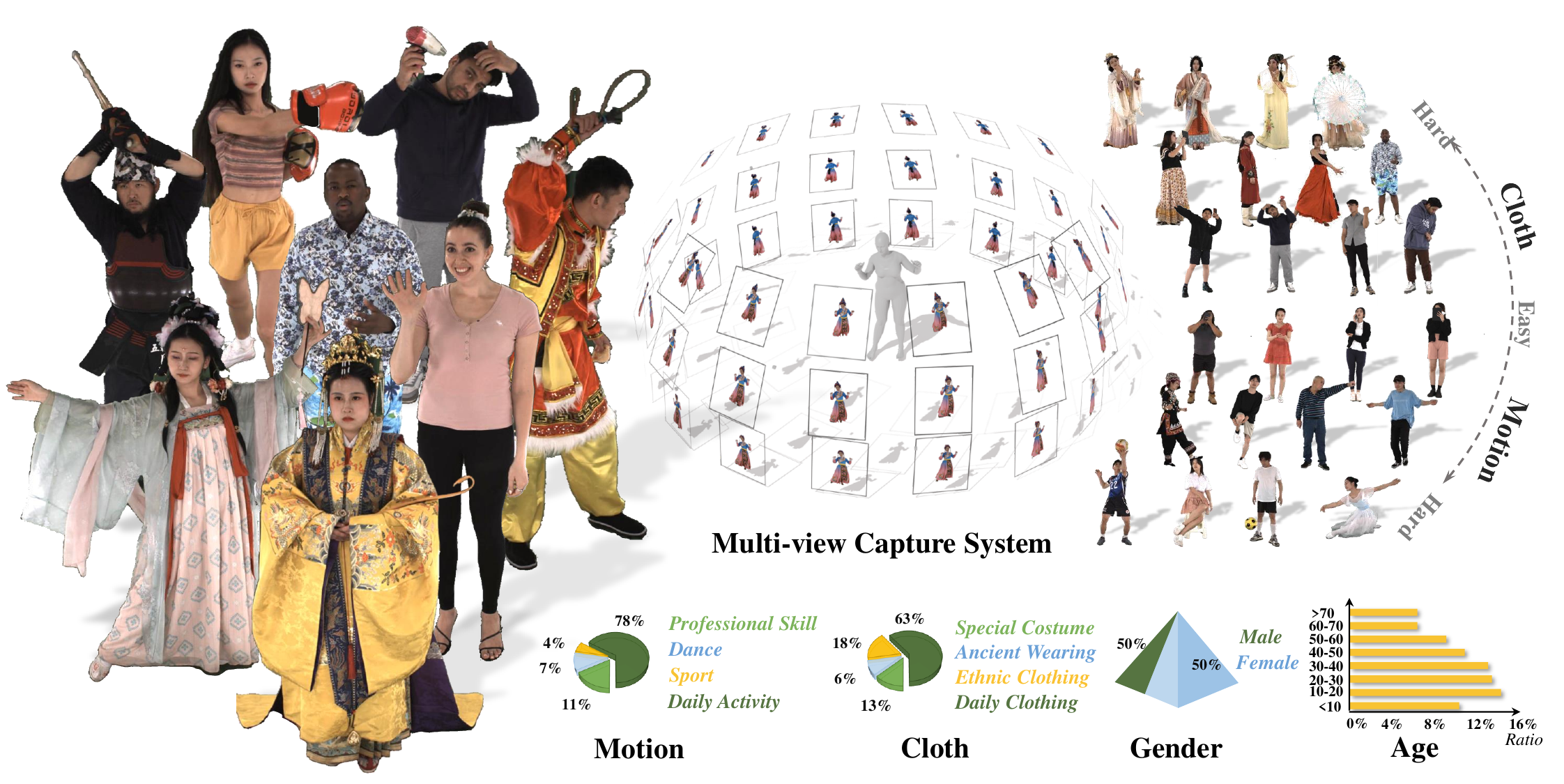}
  \vspace{-2ex}
  \captionof{figure}{\small{\textbf{Overview of our dataset}.} \datasetname~is a large-scale human-centric dataset, with high-quality multi-view images and videos for various human actors. The dataset comes with grand categories of motion, cloth, accessory, body shape, and human-object interaction. We hope it could boost the development of human-centric rendering and related tasks.}
  \label{fig:intro-large}
  \end{center}
}]

{\let\thefootnote\relax\footnotetext{\noindent\textsuperscript{*}Joint-first authors with W. Cheng.}}
\let\thefootnote\relax\footnotetext{\dag Equal advising.}

\begin{abstract}

Realistic human-centric rendering plays a key role in both computer vision and computer graphics. Rapid progress has been made in the algorithm aspect over the years, yet existing human-centric rendering datasets and benchmarks are rather impoverished in terms of diversity (\textit{e.g.,} outfit's fabric/material, body's interaction with objects, and motion sequences), which are crucial for rendering effect. Researchers are usually constrained to explore and evaluate a small set of rendering problems on current datasets, while real-world applications require methods to be robust across different scenarios. In this work, we present~\textbf{\datasetname}, a large-scale, high-fidelity repository of human performance data for neural actor rendering.
\datasetname~presents several alluring attributes. First, our dataset contains over $1500$ human subjects, $5000$ motion sequences, and $67.5M$ frames' data volume. Upon the massive collections, we provide human subjects with grand categories of pose actions, body shapes, clothing, accessories, hairdos, and object intersection,  which ranges the geometry and appearance variances from everyday life to professional occasions. Second, we provide rich assets for each subject -- \twoD/\threeD~human body keypoints, foreground masks,  \smplx~models, cloth/accessory materials, multi-view images, and videos. These assets boost the current method's accuracy on downstream rendering tasks. Third, we construct a professional multi-view system to capture data, which contains $60$ synchronous cameras with max $4096\times3000$ resolution, $15$ fps speed, and stern camera calibration steps, ensuring high-quality resources for task training and evaluation.

Along with the dataset, we provide a large-scale and quantitative benchmark in full-scale, with multiple tasks to evaluate the existing progress of novel view synthesis, novel pose animation synthesis, and novel identity rendering methods. In this manuscript, we describe our \datasetname~effort as a revealing of new observations, challenges, and future directions to human-centric rendering. The dataset, code, and benchmarks will be publicly available at \href{https://dna-rendering.github.io/}{https://dna-rendering.github.io/}.
\end{abstract}

\vspace{-1ex}
\section{Introduction}

\begin{table*}[htb]
\begin{center}
\vspace{-2ex}
\resizebox{\textwidth}{!}{

\begin{tabular}{c|ccccc|cccc|c} \hline
\multirow{2}*{\textbf{\small Dataset}} &\multicolumn{5}{c|}{\textbf{\small Attribute}} & \multicolumn{4}{c|}{\small \textbf{Scale}} & \multicolumn{1}{c}{\small \textbf{Realism}}  \\
\multirow{2}*{\textbf{~}} &\small  Ethnicity & \small   Age  & \small   Cloth   &  \small  Motion   &\small  Interactivity & \small \#ID $\times$ \#Outfit &\small  \#Motions & \small \#View  & \small \#Frames & \small HRes \\ \small \textit{Human3.6M}~\cite{ionescu2013human3}    & \small \color{DarkRed}{\xmark}    & \small \color{DarkRed}{\xmark} &  \small  \color{DarkRed}{\xmark}  &  \small  \color{OliveGreen}{\cmark}   & \small \color{OliveGreen}{\cmark}   & \small $11\times1$ & \small $17$ & \small $4$  &\small $3.6$M & \small $1000$P   \\
\small \textit{CMU Panoptic}~\cite{Joo_2017_TPAMI} & \small \color{OliveGreen}{\cmark}    & \small \color{OliveGreen}{\cmark} &  \small  \color{DarkRed}{\xmark}  &  \small  \color{OliveGreen}{\cmark}   & \small \color{OliveGreen}{\cmark}   & \small $97\times1$ & \small $65$ & \small $31+480^{*}$ & \small $15.3$M & \small $1080$P   \\
\small \textit{\zjumocap}~\cite{peng2020neural}    & \small \color{DarkRed}{\xmark}    & \small \color{DarkRed}{\xmark} &  \small  \color{DarkRed}{\xmark}  &  \small  \color{OliveGreen}{\cmark}   & \small \color{DarkRed}{\xmark}   & \small $10\times1$ & \small $10$ & \small $24$ & \small $180$K & \small $1024$P   \\
\small \textit{HUMBI}~\cite{yu2020humbi} & \small \color{OliveGreen}{\cmark}    & \small \color{OliveGreen}{\cmark} &  \small  \color{OliveGreen}{\cmark}  &  \small  \color{DarkRed}{\xmark}   & \small \color{DarkRed}{\xmark}   &\small $772\times1$ & $-$ &  \cellcolor{Gold!40} \small $107$ & \small $26$M & \small $1080$P  \\
\small \textit{AIST++}~\cite{aist-dance-db,li2021ai}              &\small \color{DarkRed}{\xmark} &  \small \color{DarkRed}{\xmark} & \small \color{DarkRed}{\xmark} & \small \color{DarkRed}{\xmark} & \small \color{DarkRed}{\xmark} &\small $30\times1$ & \small  $-$  & \small $9$ &\small $10.1$M & \small $1080$P   \\
\small \textit{THuman 4.0}~\cite{shao2022diffustereo}       & \small \color{DarkRed}{\xmark}    & \small \color{DarkRed}{\xmark} &  \small  \color{OliveGreen}{\cmark}  &  \small  \color{OliveGreen}{\cmark}   & \small \color{DarkRed}{\xmark}   & \small $3\times1$ & \small $-$ & \small $24$ & \small $10$K & \small $1150$P   \\
\small \textit{\humman}~\cite{cai2022humman}              &\small \color{OliveGreen}{\cmark} & \small \color{OliveGreen}{\cmark} & \small \color{OliveGreen}{\cmark} & \small \color{OliveGreen}{\cmark} &\small \color{DarkRed}{\xmark} & \cellcolor{LightGray!50}\small $1000\times1$& \cellcolor{LightGray!50}\small $500$ &\small $10$  & \small \cellcolor{LightGray!50}$60$M & \small $1080$P  \\
\small \textit{\genebody}~\cite{cheng2022generalizable}  & \small \color{OliveGreen}{\cmark}    & \small \color{OliveGreen}{\cmark} &  \small  \color{OliveGreen}{\cmark}  &  \small  \color{OliveGreen}{\cmark}   & \small \color{OliveGreen}{\cmark} &\small $50\times2$ &\small $61$ &\small $48$ & \small  $2.95$M & \small \cellcolor{LightGray!50} $2048$P  \\ \hline 
\small \textbf{\textit{\datasetname~(Ours)}}              & \small \color{OliveGreen}{\cmark}    & \small \color{OliveGreen}{\cmark} &  \small  \color{OliveGreen}{\cmark}  &  \small  \color{OliveGreen}{\cmark}   & \small \color{OliveGreen}{\cmark}   & \cellcolor{Gold!40} \small$500\times3$ & \small \cellcolor{Gold!40} $1187$ & \small \cellcolor{LightGray!50} $60$ & \small \cellcolor{Gold!40} $67.5$M & \small \cellcolor{Gold!40} $4096$P   \\ \hline
\end{tabular}

}
\vspace{-1ex}
\caption{\textbf{Dataset comparison on attributes and scales.} We compare the proposed dataset with previous human-centric multiview datasets in terms of attribute coverage, scale, and realism. `Ethnicity' denotes whether the dataset contains actors from multiple ethnic groups. `Age' means if there is a wide age range containing elders or infants. `Cloth' separates datasets with only daily costumes or with extra diverse clothing. `Attribute-Motion' denotes whether it has human motion in different scenarios. `Interactivity' tells whether there contains human-object interaction. We mark these attributes with {\color{OliveGreen}{\cmark}}~and {\color{DarkRed}{\xmark}}. 
In scale, we list the number of key factors with compared dataset, Note that `Scale-$\#$Motions' means the number of motion categories, and superscript $^{*}$ means low-resolution VGA cameras, we exclude them during `$\#$View' ranking and `$\#$Frames' calculation. 
Cells \legendsquare{Gold!40}~\legendsquare{LightGray!50} indicate the best, and second best in the specific category among all datasets. We abbreviate resolution at height as `HRes'.}
\vspace{-6ex}
\label{tab:dataset}
\end{center}
\end{table*}

Understanding humans is an everlasting problem in our research community, and extensive literature on perceiving and synthesizing humans shows great efforts toward this goal. Over the decades, multiple pioneers have constructed large-scale and diverse datasets, such as COCO~\cite{lin2014microsoft} for human pose estimation, and ActivityNet~\cite{caba2015activitynet} for analyzing human action. These datasets are the driving force behind flourishing developed human-centric perceiving algorithms.

Yet, when it comes to human-centric rendering, there is still a noticeable gap in comprehensive datasets. 
Capturing high-quality and massive 3D/4D human avatars is difficult due to the requirements of high-end equipment as well as an efficient data processing pipeline.  Existing datasets~\cite{ionescu2013human3,joo2015panoptic,peng2020neural,shao2022diffustereo,isik2023humanrf} partially narrow the gaps but have significant limitations on sample diversity (\textit{e.g.,} clothing, motion, body shape, and human-object interaction), or have insufficient realism (\textit{e.g.,} camera resolution, and capture speed). These factors are crucial to rendering effects.

To drive advancement in human-centric rendering, we contribute a large-scale multi-view human performance capture dataset, named \datasetname, which includes the factors that are important to rendering in great diversity and granularity. {\textit{On the hardware side}}, we build up a 360-degree indoor system equipped with $60$ calibrated RGB cameras and $8$ synchronized depth sensors.
The captured videos are under the fidelity of up to $12$MP ($4096 \times 3000$) resolution and recorded at $15$ fps. {\textit{From the dataset's footage design aspect}}, we intend to cover most attributes (\textit{e.g.,} age, ethnicity, shape, motion, cloth, accessory, and interactive objects) that could reflect the rendering differences with respect to texture, materials, primary/secondary motion deformation, and category priors. In practice, we design over $1500$ outfits and $1187$ motion types to ensure the comprehensive coverage of real-world scenarios. We invite $500$ actors to participate in the data capture process. We record each person with three different outfits and at least nine unique motion sequences. The full dataset contains $5000$ video sequences with over $67.5M$ frames. Compared with the existing human-centric dataset like CMU Panoptic~\cite{Joo_2017_TPAMI}, \zjumocap~\cite{peng2020neural}, THUman~\cite{shao2022diffustereo}, and Human3.6M~\cite{ionescu2013human3}, \datasetname~comprises the most multi-view body performance samples and reaches the highest image quality. The unfold  comparisons between \datasetname~and the others are given in Tab.~\ref{tab:dataset}.

Meanwhile, we provide essential annotations attached to each frame to facilitate the application of downstream tasks. The ultra-large scale of the \datasetname~dataset raises great challenges to the corresponding data processing steps. To this end, we develop an automatic annotation pipeline encompassing camera calibration, color correction, image matting, \twoD~/3D landmark estimation, and \smplx~model fitting. To ensure the labeling quality, we developed a series of technical refinements to the annotation toolchain. With these efforts, the automatic pipeline can generate faithful data annotations both effectively and efficiently. To further assist the community in the public use of external data, we will open-source the annotation tool used in the \datasetname~project.

The unprecedented richness of \datasetname~ dataset provides fertile data soil for researchers to develop, and dissect their rendering methods in depth. To set up a kickoff example, we further construct benchmarks upon the dataset with extensive experiments. We evaluate the performances of several state-of-the-art full-body rendering and animation approaches under three major tasks, {\textit{i.e.,}} {\textit{novel view synthesis}}, {\textit{novel pose animation}}, and {\textit{novel identity rendering}}. To better analyze current methods in terms of the model capacity, module necessity, and methodology generality, we set up multiple test set splits under different levels of challenging aspects. For instance, we divide the \textit{easy, medium,} and \textit{hard} subsets {\textit{w.r.t.}} the cloth looseness, the texture complexity, the motion difficulty, and the human-object interactivity, respectively. We conclude a series of key observations based on the benchmarks, such as how human prior influences the robustness of rendering, how sensitive the multi-view/frame relationship module design is to data volume/distribution, and how loss design affects the performance in terms of different rendering metrics.

In summary, we contribute \datasetname~project to fulfill the requirement of a high-fidelity human performance capture dataset for the research community. We establish by far the largest multi-view human body performance dataset for high-fidelity human-centric rendering research, with an emphasis on image quality and data attributes. The attached benchmarks provide baseline standards for three major tasks, with rigorous evaluations and dissections on multiple state-of-the-art methods. We believe the dataset, the attached benchmarks, and the tools will boost a wide range of digital human applications and inspire future research.

\vspace{-1ex}
\section{Related Works}

\subsection{Human-centric Datasets}

\noindent{\textbf{Perception Datasets.}}
Perceiving human is a long-standing problem. Over the decades, researchers have kept dedicating their efforts to building relevant datasets.  Earlier efforts in the computer vision community present large-scale datasets like COCO~\cite{lin2014microsoft} for human segmentation or keypoint detection from in-the-wild images. Later research works follow the inspiration to establish open-world datasets ~\cite{chen2014detect,gong2018instance}, while with emphasis on parsing more precise human body parts. Some researchers focus on constructing datasets~\cite{chen2015utd,ofli2013berkeley,shahroudy2016ntu} that capture daily activities and help the perception of human action recognition by using RGB-D cameras. Despite the wild variety of data samples, these datasets are not capable of human rendering tasks, due to a lack of multiview images as groundtruth references for evaluating methods' performance. 
In computer graphics society, there is another parallel branch that contributes datasets ~\cite{cmumocap,sfumocap} for avatar animation, with recording human motion via maker-based motion capture systems. AMASS~\cite{AMASS:ICCV:2019} further integrates these motion capture databases with fully rigged surface mesh representation. In the last decade, computer vision and graphics society fit in with each other in the field of perceiving 3D humans. Datasets like Human3.6M~\cite{ionescu2013human3} and MPI-INF-3DHP~\cite{mono-3dhp2017} capture humans under in-door multi-view environment with 3D marker label or multiview segmentation, which further encourage the applications in recovering human in \threeD. 3DPW~\cite{von2018recovering} dataset captures human motion in the wild and annotates \threeD~pose with the help of pre-scanned human model. These databases facilitate the development of numerous algorithms. These databases facilitate the development of numerous algorithms. 
However, due to the limits of the data sample, camera views, and resolution, they cannot reflect the pros and cons of rendering methods.

\noindent{\textbf{Rendering Datasets.}} Representing \threeD~/4D human appearances and performances are important in both research communities and commercial applications. The progress of relevant algorithms relies on human scan datasets with accurate geometry or dense views.  Such datasets are the foundational factor to close the gap between virtual avatars and real humans. THuman~\cite{zheng2019deephuman,yu2021function4d,deepcloth_su2022}, ClothCap~\cite{pons2017clothcap}, SIZER~\cite{tiwari2020sizer} and other commercial scan datasets~\cite{web:renderpeople,web:3dpeople} capture static human scan reconstructed by either depth sensors or camera array. ~\cite{Bogo:CVPR:2014,dfaust:CVPR:2017,Zhang_2017_CVPR,shao2022diffustereo,zheng2022structured} provide dynamic human scans with minimal clothing and daily costumes. These datasets are usually biased centering on standing poses due to the sophisticated capture process. With the emergence of neural rendering techniques, rendering realistic humans directly from images has become a trend. Such a setting usually requires the dataset equipped with high-quality dense view images and accurate annotations like human body keypoints and foreground segmentation.
CMU Panoptic~\cite{joo2015panoptic} uses a 30-HD-camera system and annotates the humans with \threeD~keypoints. HUMBI~\cite{yu2020humbi} focuses on local motions like gesture, facial expression, and gaze movements, rather than factors that have influences on rendering quality, such as cloth texture, object interactivity, \textit{etc}. \zjumocap~\cite{peng2020neural} is a widely used dataset for human rendering algorithms. It includes ten video sequences with $24$ cameras under $1K$ resolution and provides annotations of human segmentation and estimated SMPL model for each frame. However, \zjumocap~is limited in narrow motion and clothing diversity, which might lead the evaluation to great bias. AIST++~\cite{aist-dance-db,li2021ai} is a dance database with various dance motions while sticking in the one-fold scenario and lacking view density. Recently proposed~\humman~\cite{cai2022humman} and \genebody~\cite{cheng2022generalizable} datasets, expand the motion and clothing diversity, while the effective human resolution is still below $1$K. Differing with these efforts, we make a step further and contribute the largest high-fidelity multiview dataset for human-centric rendering tasks,  with $5000$ human performance sequences under a large variation in ethnicity, age, cloth, motion, and interaction. Concurrent works~\cite{zhou2022relightable,isik2023humanrf} also contribute datasets for human avatar tasks, while centering on detailed human geometry with long-lens cameras to film human body parts. In practice, \cite{zhou2022relightable} captures on low frame rate, and ~\cite{isik2023humanrf} enjoys dense multi-view capturing but limits to $16$ motion sequences. Moreover, their setups are vulnerable to the capture of aboriginal motion tracks at full-body levels.

\subsection{Implicit Neural Body Representation}
Different from previous works that represent humans with explicit representations (such as skeleton, parametric model, or mesh), recent work models human appearance as neural implicit function, \textit{e.g.}, neural radiance fields~\cite{mildenhall2020nerf} or neural signed distant functions. PIFu~\cite{saito2019pifu,saito2020pifuhd} presents the orthogonal camera space as an occupancy function conditioned by pixel-aligned features and depth. It learns to reconstruct the human body with single-view images, which inspires subsequent research on human reconstruction from a single image~\cite{lazova2019360,alldieck2019tex2shape,alldieck2019learning,xiu2022icon,xiu2023econ}. \nb~\cite{peng2020neural} learns a neural radiance field of dynamic humans conditioned by body structure and temporal latent code from sparse multi-view videos.  Recently, many category-agnostic implicit representations, \pixel~\cite{yu2020pixelnerf}, \ibr~\cite{wang2021ibrnet}, \vision~\cite{lin2023visionnerf}, {\textit{etc.}}, can generalize NeRF to arbitrary unseen scenes given a set of reference views.
The key insight behind these methods is that the novel views of new scenes can be recovered from visual clues of reference views and neural implicit functions that are derived by rendering equations. Rafting these two factors together can help models learn visual consistency on certain camera distributions. 
The intrinsic differences among these methods are the design of feature aggregation, which varies from average~\cite{yu2020pixelnerf}, max~\cite{qi2017pointnet} pooling to more adaptive weighted pooling~\cite{wang2021ibrnet} and vision transformer~\cite{lin2023visionnerf}. Given human rendering is more challenging due to the large variation in pose and appearance, recent generalizable human rendering methods~\cite{zhao2021humannerf,mihajlovic2022keypointnerf,cheng2022generalizable,kwon2021neural} condition such image feature-aligned NeRF with human priors. For example, \nhp~\cite{kwon2021neural} uses structured latent code and \kptnerf~\cite{mihajlovic2022keypointnerf} deploys human keypoints. These methods outperform category-agnostic methods on human rendering on existing human rendering datasets, while such improvement is not convincing due to the limited scale and coverage bias of these datasets.
We believe the proposed large-scale dataset with large variation overall dimensions will help both the robustness of generalizing humans and better generalization ability assessment. 

\subsection{Animatable Digital Human}
The challenge of creating realistic animatable human avatars from images is two folds -- (1) how to reconstruct the human body from motion sequences and (2) how to disentangle non-rigid deformation. 
Early seminal work \anerf~\cite{su2021nerf} learns dynamic body from sequences, it conditions the radiance field with relative pose coordinate of the query point, which fails to model the non-rigidity of clothed humans. To reconstruct the human body from sequences, \aniN~\cite{peng2021animatable} learns a static canonical radiance field together with a ray blending network from the current frame to canonical space. To further better disentangle motion deformation from pose recent works~\cite{saito2021scanimate,chen2021snarf} use a complementary forward blending network or root-finding algorithm to regularize the learned blending with cycle consistency loss. Other works~\cite{weng2022humannerf,jiang2022neuman,yu2023monohuman} learn animatable models from more challenging monocular video, with a tighter assumption of Gaussian distributed occupancy along bone or fixed SMPL motion weights.

\section{\datasetname}

In this section, we discuss the core features of our \datasetname~from the perspective of dataset construction, including the hardware systems setup, data collection, and relevant dataset statistics. By introducing these aspects, we present a way to ensure high fidelity and effectiveness under the massive amount of data capture.

\subsection{Dataset Capture}\label{sec:dataset_capture}
\noindent \textbf{System Setup.} 
Our capture system contains a high-fidelity camera array, with $60$ high-resolution RGB cameras and $16$ lighting boards uniformly distributed in a sphere with a radius of three meters. The cameras are adjusted to point at the sphere's center, where the participants perform. Concretely, the array consists of $48$ high-end $2448\times2048$ industrial cameras, and $12$ ultra-high resolution cameras with up to $4096\times3000$ resolution. We use these industrial cameras to capture subjects' body and facial performances, and the ultra-high resolution cameras to capture more detailed textures on clothing and accessory.  We additionally place eight Kinect cameras to capture additional depth streams as auxiliary geometric data. The high-fidelity video streams and depth streams are synchronized at $15$ frames per second. To guarantee the cameras can clearly capture subjects' performances and clothing patterns from all views, the lighting is adjusted to the setting of $5600$K $\pm$ $300$K color temperature and $4500$ Lux/m illuminance. The above designs ensure the system could record the sharp texture edges, fine-grained color changes of clothing patterns, and the reflection effects caused by different clothing materials. Please refer to Sec.~\ref{sec:sup:dataset-capture} for more details.

\noindent \textbf{Data Collection Protocol.} To enable subsequent research probing into the factors that have influences on rendering, we design a data collection protocol with both interlaced and hierarchical data attributes. Specifically, we ask each actor to wear three sets of outfits and perform at least three actions in different hallucinated scenarios for each outfit, which maximize the identity scale and diversity. Each motion sequence is recorded under specific action category instruction with a free-style performance lasting for $15$ seconds, which ensures the diversity of action performance. As an auxiliary feature, we also capture a static frame of A-pose for actors in each outfit for canonical pose recording, and a frame with only empty background for image matting. Multiview videos are inspected on-site with a quick preview generation tool, which ensures the movement fits best to the cameras' field of view. For accurate camera pose annotation, extrinsic calibration data are collected at a daily frequency. The color data and intrinsic calibration data are collected whenever system adjustments are made. Please refer to Sec.~\ref{sec:sup:dataset-capture} for details.

\subsection{Dataset Statistics} In order to cover diverse attributes that relate to rendering quality, we have carried out a detailed design from the selection of the actors' gender, age, and skin color, to their actions, clothing, and makeup. Compared with previous work, to our best knowledge, our dataset contains the largest number of human subjects, covering the most diverse action categories, clothing types, and human-object interaction scenarios. The key statistics of our dataset are shown at the bottom of Fig.~\ref{fig:intro-large}. Specifically, to preserve authenticity in action behavior, we invite $153$ professional actors to perform special scenes with corresponding costumes/makeup, and $347$ normal performers to act under footage of daily-life scenes. The special scenes constitute $153$ sub-categories, including sports, dances, and unique event performances such as typical costumes in ancient Chinese dynasties, traditional costumes around the world, cosplay, \textit{etc}. Common scenes can be divided into $269$ sub-categories, covering scenes such as daily indoor activities, communication, entertainment, and new trends. We describe the comprehensive distribution of data in Sec.~\ref{sec:sup:dataset-stats} and the limitation of data in Sec.~\ref{sec:sup:dataset-limitation}. Please refer to the corresponding sections for more visualized figures and detailed discussion. 

\begin{figure}[!t]
\vspace{-2ex}
    \centering
    \includegraphics[width=0.98\linewidth]{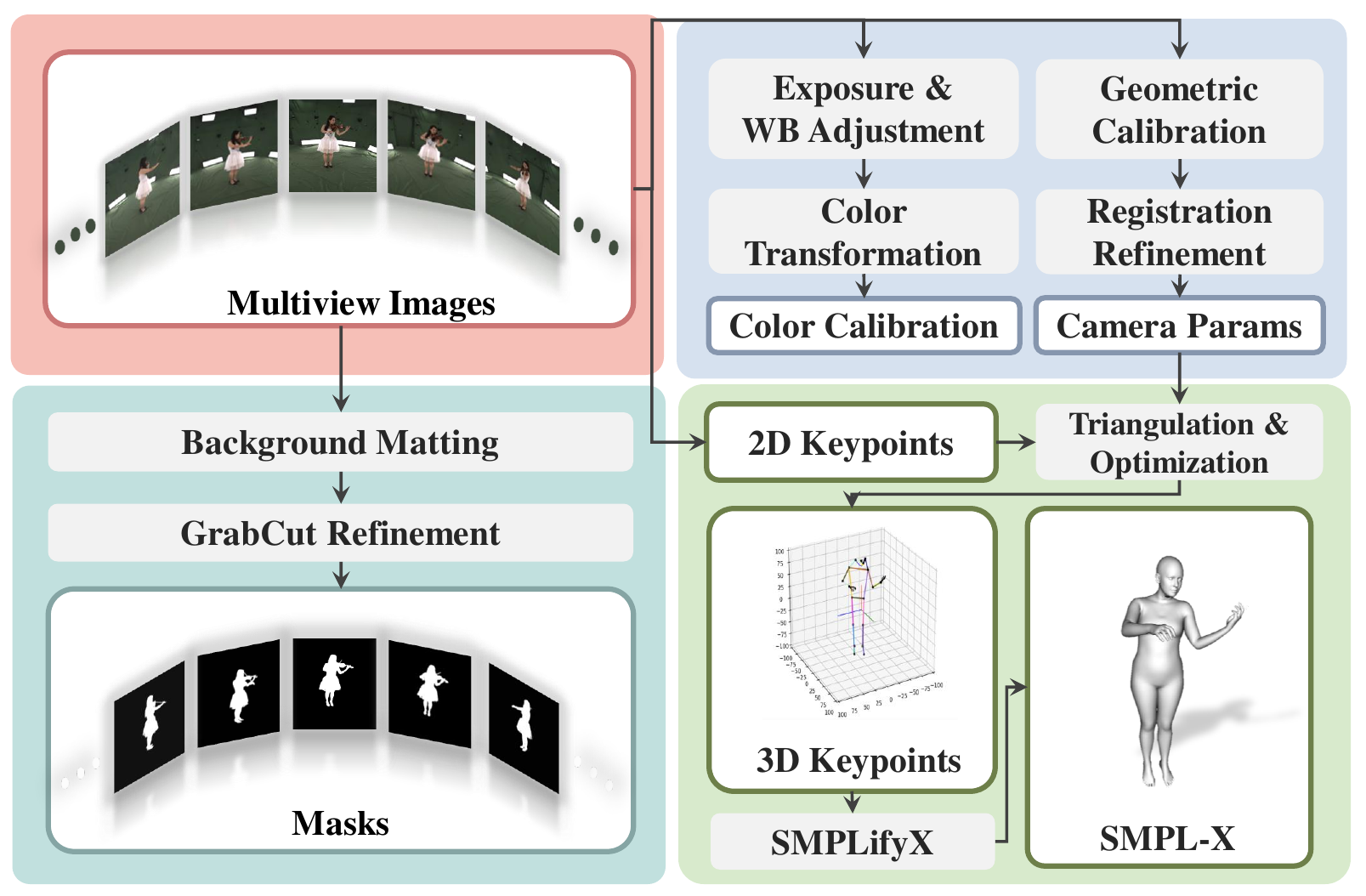}
    \vspace{-2ex}
    \caption{\small{\textbf{Annotation pipeline.} The illustration of annotation pipeline for camera calibration, camera color calibration, masks, keypoints, and parametric model.} }\label{fig:pipeline}
\vspace{-2ex}
\end{figure}

\subsection{Data Annotation}\label{sec:annotation}
To enable applications in human rendering and animation, \datasetname~provides rich annotations attached with the raw data, \textit{i.e.,} camera calibration, camera color calibration, image matting, and parametric model fitting. The overall annotation pipeline is shown in Fig.~\ref{fig:pipeline}.

\noindent \textbf{Camera Calibration.} First, we calibrate the intrinsic parameters of each camera individually. Specifically, we divide the camera's field of view into a $3\times3$ Sudoku, and capture images with $\pm30$ degree rotation in pitch, row, and yaw angle of checkerboard in all grids, referring to Fig.~\ref{fig:intrinsic}. Second, for extrinsic calibration, we deploy multiple ChArUco boards and spin the main board in the capture volume. We use open toolboxes~\cite{xrprimer,multical} to optimize intrinsic parameters, distortion coefficients, and extrinsic parameters with the captured data. To eliminate the depth camera pose error caused by the large resolution gap between industrial cameras and Kinect depth cameras, we further adopt a point cloud registration stage to refine the depth camera extrinsic parameters in the second stage. More concretely, for each depth camera, we project the partial point cloud and estimate a full point cloud from the MVS algorithm such as ~\cite{wang2020patchmatchnet} as a reference. We jointly optimize the pose graph of the depth camera for neighboring pointclouds with overlaps through a multi-way registration~\cite{choi2015robust} with MVS pointcloud as reference. 

For detailed camera calibration estimation, please refer to Section.~\ref{sec:sup:annot-calib} in the supplementary.

\noindent \textbf{Color Calibration.}
The identical color response across different cameras could be vital for a multi-view, mixed-type camera system to provide qualified data for rendering applications, as it is an essential data basis for algorithms to render realistic view-dependent effects. Different from other multi-camera datasets, \textit{e.g.}, Multiface~\cite{wuu2022multiface,Lombardi21} which uses a network to optimize the color transformation during model training, we pay attention to ensure the color consistency of data collection across different cameras. First, we conduct careful adjustments on hardware parameters such as exposure and white balance to make the captured color of the color checkerboard under the standard light as close as possible. Then, the 2-order polynomial correction coefficients could be optimized by least square regression of transforming the detected color to the true value on the color checkerboard. Please refer to Sec.~\ref{sec:sup:annot-calib} and Fig.~\ref{fig:colorcalib} for details. We also analyze the impact of color consistency of multi-camera datasets on generalizable rendering in Sec.~\ref{sec:sup:cross-dataset-view}.

\begin{figure}[t]
\vspace{-2ex}
\centering
\includegraphics[width=0.98\linewidth]{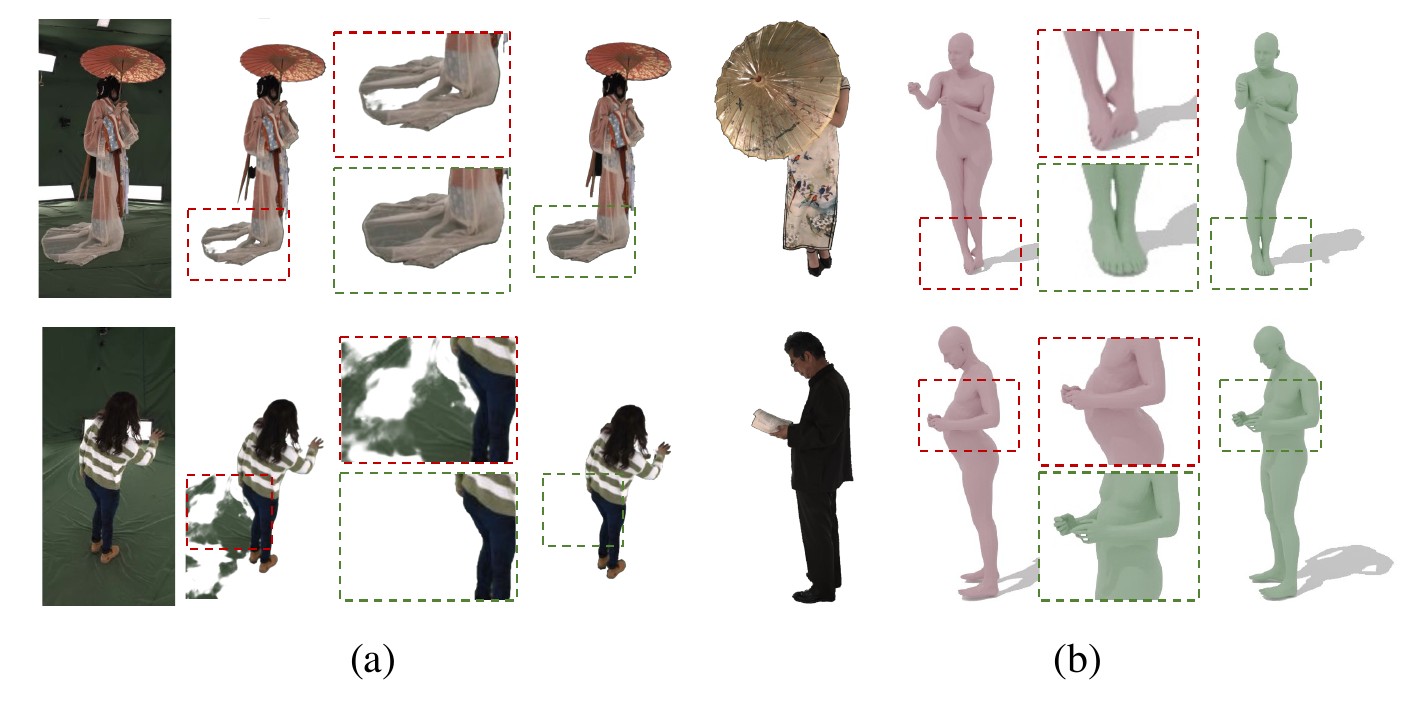}
\vspace{-2ex}
\caption{\small{\textbf{Annotation quality improvements.} The zoom-in boxes with red dot lines show the annotation quality before pipeline optimization. The green ones show quality improvements over (a) mask annotation and (b) \smplx~annotation with the optimized pipeline.}}\label{fig:mask_smplx_annotation}
\vspace{-4ex}
\end{figure}

\noindent \textbf{Matting.}
Considering the large quantities of the captured images, we develop an automatic matting pipeline to extract the foreground objects from the backgrounds. 
We first adopt an off-the-shelf background matting model~\cite{lin2021real} to eliminate most background pixels.
However, due to the complicated nature of the capture settings, the learning-based model inevitably generates unsatisfying results in some challenging cases, leaving some pieces of labeled data with artifacts such as broken holes or noisy patches (See Fig.~\ref{fig:mask_smplx_annotation}). Thus, we further propose a refinement strategy by applying the \textit{HSV} filtering and the grabcut~\cite{rother2004grabcut} algorithms to improve the matting quality. According to the manual assessment of the refined masks, the error rates of problematic cases are reduced from $11$\% to $2$\% on average. We compare matting with and without refinement, and visualize detailed manual assessment in Fig.~\ref{fig:matting_errors}.

\noindent \textbf{Keypoints and Parametric Model.}
Inspired by existing works \cite{cai2022humman, cai2021playing}, we develop an automatic pipeline to annotate keypoints and parametric model parameters. 
1) First, \twoD~keypoints in COCO-Wholebody \cite{jin2020whole} format (including body, hand, and face keypoints) are detected for each camera view, with pretrained model HRNet-w48~\cite{sun2019deep}. 
2) Then, we triangulate \threeD~keypoints with known camera intrinsic and extrinsic parameters from the multi-view \twoD~keypoints with optimization and post-processing strategies \cite{xrmocap} including keypoint selection, bone length constraint, as well as outlier removal. 
3) Finally, we register the \smplx, a commonly used parametric model, via \threeD~keypoints. Body shape $\beta \in \mathbb{R}^{n \times 10}$ (or $\beta \in \mathbb{R}^{n \times 11}$ for children \cite{hesse2018learning,patel2021agora}), pose parameters (body pose, hand pose, and global orientation) $\theta \in \mathbb{R}^{n \times 156}$, and translation parameters $t \in \mathbb{R}^{n \times 3}$ ($n$ is the number of frames) are estimated via a modified SMPLify-X~\cite{pavlakos2019expressive} for dynamic poses. 

Our annotation pipeline has proved effective and robust in getting natural \smplx~model, as shown in Fig.~\ref{fig:mask_smplx_annotation}. We evaluate the fitting error between \threeD~keypoints and corresponding regressed \smplx~joints. The mean and median `Mean Per Joint Position Error' (MPJPE) of our system is $30.20$ mm and $29.80$ mm. The error is on par with the oracle fitting accuracy of $29.34$ mm in Human3.6M~\cite{ionescu2013human3, loper2014mosh}, which includes data from an optical motion capture system. Detailed analysis is conducted in Sec.~\ref{sec:sup:annot-keypoint} and Sec.~\ref{sec:sup:annot-smpl}. A thorough comparison of our fitting pipeline with other fitting methods~\cite{shuai2022multinb,zheng2021pamir,cheng2022generalizable} is described in Sec.~\ref{sec:sup:annot-smplify}.

\section{Benchmarking Human-centric Rendering}\label{sec:benchmark}

Our~\datasetname~dataset could be used to unfold the reflections and boot the developments of research on high-fidelity human body rendering tasks, due to its \textit{large-scale volume, diverse scenarios, multi-level challenges, and high-resolution multi-view data} properties. To kick off an example of how to utilize this dataset, we set up benchmarks with exclusive experiments centered around three fundamental tasks of human body rendering, {\textit{i.e., novel view synthesis, novel pose animation, and novel identity rendering}}. In this section, we introduce the benchmark settings and key observations. 

\subsection{Data Splits}\label{sec:data_split}
To unfold each method in depth, and thoroughly evaluate the effectiveness of our dataset, we construct multiple training and testing data splits to conduct level tests for each method. We consider the four most influential factors of rendering quality for the benchmark test, {\textit{i.e.,}} the looseness of clothes, the texture complexity, the pose difficulty, and the interactivity between the human body and manipulated object.

\noindent \textbf{The Cloth Looseness.} 
We define the cloth's challenging levels by the deformation distance between the minimal-cloth human body and the clothing outline, and the softness of cloth materials.
For the \textit{Easy} level, we collect simple cases wearing tight-fitting clothes like yoga wear and sports t-shirts. 
The \textit{Medium} level includes the daily clothes such as coats, skirts, jeans, loose t-shirts, \textit{etc.} 
As for the \textit{Hard} level, we make a split containing ethical costumes, national clothing, and fancy decorations.

\noindent \textbf{The Texture Complexity.}
The texture distribution also plays an important role in the human body rendering tasks. 
To examine the correlations between texture complexity and rendering performance, we build three data splits for texture evaluation. The \textit{Texture-Easy} split is composed of single-color clothes. The \textit{Texture-Medium} split includes most daily clothes in a few colors and plain patterns.
The \textit{Texture-Hard} split contains the most complicated texture clothes with intricate patterns like dots, stripes, plaids, \textit{etc.}


\noindent \textbf{The Pose Difficulty.}
In the novel pose animation task, it is vital to probe if the trained models could handle different levels of motion sequences in terms of (non-rigid) difficulties and degree of out-of-distribution (OOD). 
Therefore, we prepare the \textit{Motion-Easy, Motion-Medium,} and \textit{Motion-Hard} splits for training and evaluation.
The \textit{Easy} data are simple motions with limited body parts involved, like shaking and waving hands.
The \textit{Medium} level refers to casual motions including full-body actions such as walking, eating, sitting, kneeling, stretching, {\textit{etc}}.
Moreover, the \textit{Hard} split is designed to cover the extremely challenging motion cases that are performed by professional sports players or actors, {\textit{e.g.,}} instrument playing, sports action, yoga, and dancing.

\noindent \textbf{The Human-Object Interactivity.}
Finally, we propose to evaluate the impact of human-object interactivity by novel view synthesis and novel pose animation tasks.
The data difficulty level is determined by the object size. Specifically, we define four data splits: the \textit{Interaction-No} split contains pure human motions with no interactive objects; the \textit{Interaction-Easy} split includes rigid small-size hand-held objects like cellphones, pencils, cigarettes, and cups. The \textit{Interaction-Medium} split has middle-size hand-held objects, \textit{e.g}., handbag, volleyball, newspaper, {\textit{etc}}. This split includes both rigid motions and non-rigid object motions; and the \textit{Interaction-Hard} split consists of large-size assets such as yoga mats, desks, chairs, and sofas. 

To sum up, we construct an overall train split consisting of $400$ sequences with even distribution on all human factors and difficulties, and $13$ test factor-difficulty splits in total with three sequences in each test split.

\subsection{Task Definition}\label{sec:task-definition}
Human body rendering and animation problems have been popular research topics in the digital human area for decades. With the development of implicit scene representations like NeRF~\cite{mildenhall2020nerf}, many researchers design algorithms upon the methodology. In conducting the benchmark experiment on the \datasetname~dataset, we intend to provide a thorough comparison of the state-of-the-art methods on data splits with different difficulty levels.
Depending on the generalizability of the state-of-the-art methods, we categorize the recently published works into two classes: case-specific methods and generalizable ones. We evaluate the methods under multiple problem settings according to their categories. Concretely, {\textit{we set up novel view synthesis and novel pose animation tasks for the case-specific methods, and the novel identity rendering task for the generalization approaches}}. In this section, we present the key observations of the benchmarks. We provide a detailed review of the benchmark methods and necessary modifications to apply them to our dataset in Sec.~\ref{sec:sup:methods} of the supplementary.

\noindent \textbf{Novel View Synthesis.}
Recent {\textit{dynamic}} human rendering works like \nb~\cite{peng2020neural}, \anerf~\cite{su2021nerf}, \aniN~\cite{peng2021animatable}, and \nv~\cite{lombardi2019neural} obtained impressive results by training on a single case with multi-view video data. \hn~\cite{weng2022humannerf} demonstrated the ability to render realistic novel view images of humans from monocular video sequences. In this task, we adopt the official architecture implementation of the case-specific methods and train each individual model for every single case in the DNA-Rendering test set. For a fair comparison, we unify the training setting of \nv~\cite{lombardi2019neural}, \anerf~\cite{su2021nerf}, \nb~\cite{peng2020neural}, \aniN~\cite{peng2021animatable}, and \hn~\cite{weng2022humannerf} with $42$ dense training views. we evaluate the image rendering quality of these methods on the other $18$ unseen testing camera poses. 
Meanwhile, we also train two general scene {\textit{static}} methods -- \ngp~\cite{muller2022instant} and \neus~\cite{wang2021neus}, in each testing frame with the same training views. These two methods' performances could serve as the per-frame static reconstruction baseline reference. The rendering results are analyzed based on the difficulty level of data splits. 
 
\noindent \textbf{Novel Pose Animation.}
Similar to the novel view synthesis task, we conduct novel pose animation benchmark on the four case-specific methods~\cite{peng2020neural,peng2021animatable,su2021nerf,weng2022humannerf}. 
For each test case, we split the sequence into two parts, where images from the first $80\%$ frames are used for training and the ones from the last $20\%$ are used for testing. Besides, for the SMPL-guided pose animation methods~\cite{peng2020neural,peng2021animatable,weng2022humannerf}, we provide the {\textit{SMPL parameters}} of test images for the models to infer rendering. 
As for the SMPL-free method~\cite{su2021nerf}, the trained models take the target {\textit{pose images}} as the input (\textit{i.e.,} the underlying skeletons), and optimize the novel pose representations for inference.

\noindent \textbf{Novel Identity Rendering.}
The other category of our benchmark methods is the generalizable algorithms that can be trained on multiple cases and infer across different unseen identities. Specifically, we probe three general scene generalizable methods -- \pixel~\cite{yu2020pixelnerf}, \vision~\cite{lin2023visionnerf}, \ibr~\cite{wang2021ibrnet}, and two human-centric methods -- \nhp~\cite{kwon2021neural}, and \kptnerf~\cite{mihajlovic2022keypointnerf}. To fairly compare their performances on unseen identities, we use the same training set (all training samples of the three splits, which results in $400$ sequences in total) to train the generalizable models. In the inference stage, we evaluate the image rendering quality on novel cases from each test split respectively.

\begin{figure}[h]
\vspace{-2ex}
\centering
\includegraphics[width=0.85\linewidth]{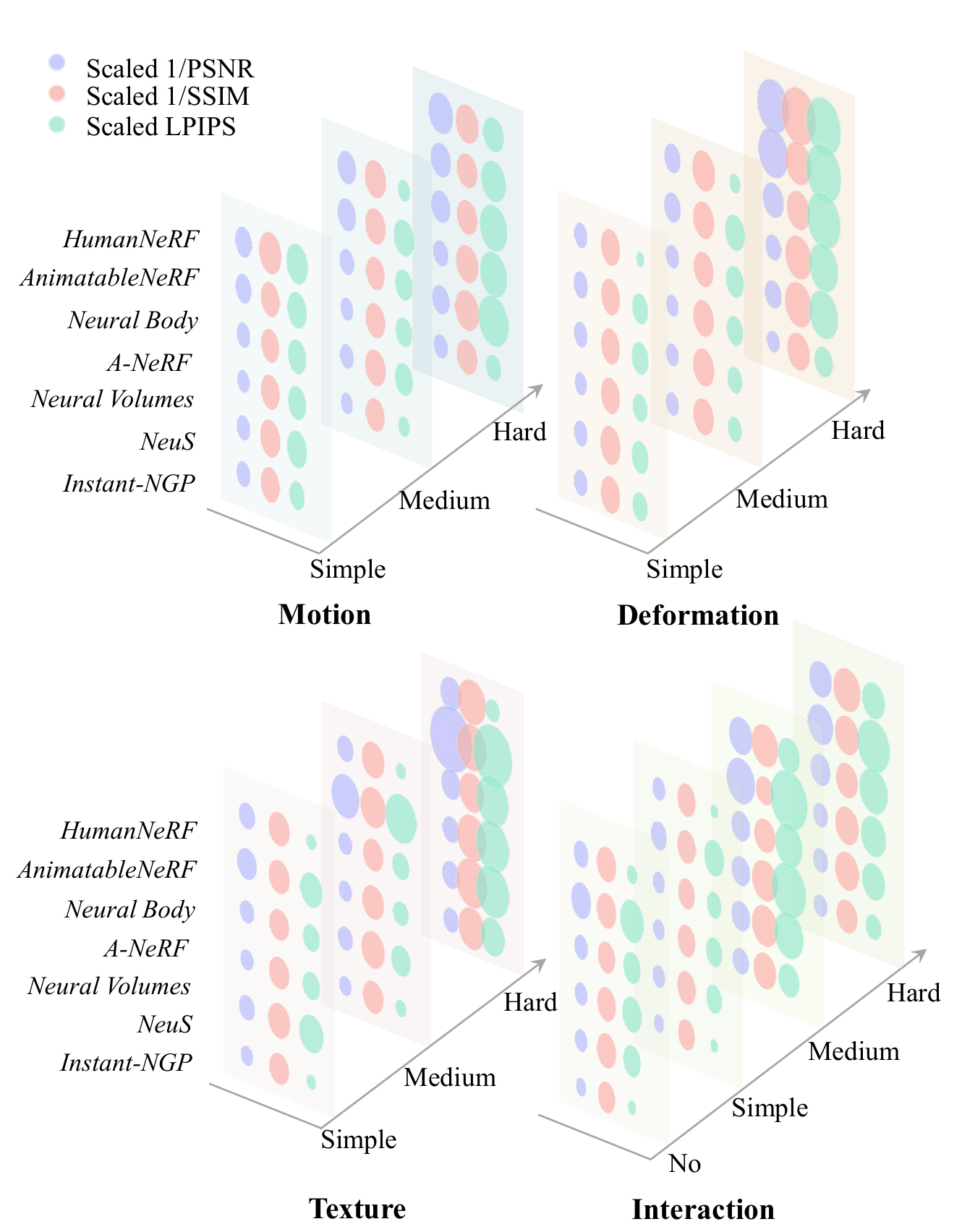}
\vspace{-1ex}
\caption{\textbf{Quantitative results visualization of novel view synthesis test across benchmarks splits and difficulties.} The colored circles denote different metrics, the smaller the circles indicate the better the novel view quality the method achieved. The numbers are reported in the appendix (Tab.~\ref{tab:novelview}).}
\label{fig:novel_view}
\vspace{-2ex}
\end{figure}

\begin{figure*}[ht]
\vspace{-4ex}
\centering
\includegraphics[width=0.95\linewidth]{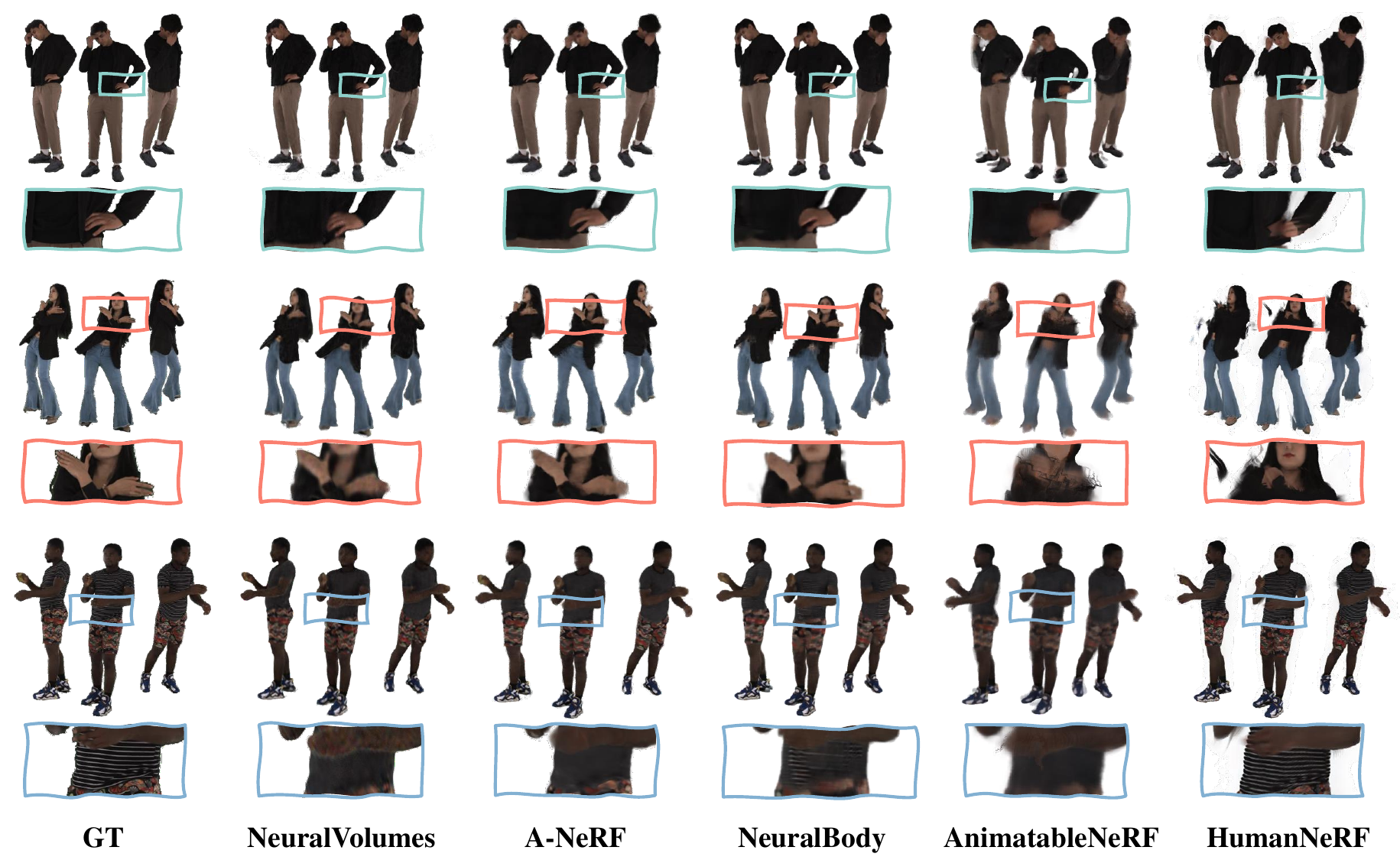}
\vspace{-2ex}
\caption{\textbf{Visualization of novel view synthesis result samples.} We qualitatively compare the novel views of the test frames in our test splits. More qualitative novel results in each split are shown in Fig.~\ref{fig:novelview_supp}.
}\label{fig:novelview}
\vspace{-3ex}
\end{figure*}


\begin{table*}[b]
\begin{center}
\resizebox{\linewidth}{!}{
\begin{tabular}{l|ccccc|ccccc|ccccc} 
\hline
\multirow{2}*{\textbf{Splits}} & \multicolumn{5}{c|}{\textbf{PSNR$\uparrow$}} & \multicolumn{5}{c|}{\textbf{SSIM$\uparrow$}} & \multicolumn{5}{c}{\textbf{LPIPS*$\downarrow$}} \\ 
& \multicolumn{1}{c}{NV} & \multicolumn{1}{c}{AN} & \multicolumn{1}{c}{NB} & \multicolumn{1}{c}{AnN} & \multicolumn{1}{c|}{HN} & \multicolumn{1}{c}{NV} & \multicolumn{1}{c}{AN} & \multicolumn{1}{c}{NB} & \multicolumn{1}{c}{AnN} & \multicolumn{1}{c|}{HN} & \multicolumn{1}{c}{NV} & \multicolumn{1}{c}{AN} & \multicolumn{1}{c}{NB} & \multicolumn{1}{c}{AnN} & \multicolumn{1}{c}{HN} \\ \hline
Motion-Simple & 22.05 & \best{26.65} & \second{25.84} & 22.78 & \third{24.65} & 0.947 & \second{0.965} & \best{0.974} & \third{0.958} & 0.953 & 78.30 & \best{58.04} & \second{58.33} & 74.33 & \third{62.76}  \\  
Motion-Medium & 19.30 & \second{21.73} & \best{21.84} & \third{21.41} & 21.14 & 0.941 & 0.951 & \best{0.969} & \second{0.957} & \third{0.952} & 92.80 & \third{71.81} & \second{65.46} & 80.97 & \best{54.46}  \\  
Motion-Hard & 19.17 & \second{21.49} & \third{20.43} & 19.64 & \best{22.48} & 0.938 & \third{0.952} & \best{0.965} & 0.949 & \second{0.964} & 105.46 & \third{83.58} & \second{82.85} & 97.98 & \best{51.18}  \\ \hline 
Deformation Simple  & 20.42 & \second{25.44} & \third{24.57} & 23.62 & \best{26.15} & 0.939 & 0.957 & \best{0.968} & \third{0.958} & \second{0.967} & 84.65 & \second{53.30} & \third{59.04} & 61.12 & \best{30.18}  \\  
Deformation-Medium & 23.09 & \best{27.26} & \second{27.05} & 23.52 & \third{24.97} & 0.945 & \second{0.963} & \best{0.974} & \third{0.961} & 0.958 & 61.09 & \second{48.41} & \third{49.91} & 65.43 & \best{33.94}  \\  
Deformation-Hard & \third{20.11} & \best{20.88} & \second{20.27} & 19.41 & 19.70 & 0.925 & \third{0.926} & \best{0.956} & \second{0.943} & 0.924 & 117.31 & 108.89 & \second{102.84} & \third{103.22} & \best{102.67}  \\ \hline 
Texture-Simple & 20.99 & \best{26.21} & \third{25.54} & 23.12 & \second{25.65} & 0.954 & \third{0.974} & \best{0.982} & 0.970 & \second{0.974} & 77.68 & \second{49.88} & \third{50.48} & 67.40 & \best{28.81}  \\  
Texture-Medium & 25.44 & \best{27.94} & \third{25.77} & 23.15 & \second{27.19} & 0.959 & \third{0.966} & \best{0.977} & 0.962 & \second{0.969} & 56.68 & \second{43.94} & \third{48.44} & 67.00 & \best{24.04}  \\  
Texture-Hard & 20.95 & \second{23.22} & \third{22.05} & 18.45 & \best{23.78} & 0.916 & 0.927 & \best{0.951} & \third{0.943} & \second{0.945} & 117.93 & \third{98.43} & \second{96.01} & 101.09 & \best{41.84}  \\ \hline 
Interaction-No & 22.64 & \best{26.32} & \third{25.41} & 22.44 & \second{25.93} & 0.957 & \third{0.968} & \best{0.980} & 0.967 & \second{0.968} & 71.98 & \third{62.55} & \second{54.71} & 69.29 & \best{31.61}  \\  
Interaction-Simple & 24.28 & \best{27.57} & \third{26.42} & 23.18 & \second{27.18} & 0.965 & \second{0.976} & \best{0.983} & 0.968 & \third{0.975} & 55.54 & \third{48.35} & \second{45.86} & 65.36 & \best{23.31}  \\  
Interaction-Medium & 20.37 & \best{23.67} & \third{21.96} & 20.81 & \second{23.20} & 0.934 & 0.950 & \best{0.965} & \second{0.953} & \third{0.951} & 95.79 & \second{84.74} & \third{87.41} & 97.32 & \best{52.10}  \\  
Interaction-Hard & 21.14 & \best{25.00} & \third{22.10} & 21.29 & \second{22.29} & 0.931 & \third{0.949} & \best{0.961} & \second{0.953} & 0.940 & 94.04 & \second{79.40} & \third{89.63} & 91.82 & \best{70.54}  \\ \hline 
Overall & 21.53 & \best{24.88} & \third{23.79} & 21.76 & \second{24.18} & 0.942 & 0.956 & \best{0.970} & \second{0.957} & \third{0.957} & 85.33 & \third{68.56} & \second{68.54} & 80.18 & \best{46.73} \\ \hline 
 
\end{tabular}
}
\vspace{-1ex}
\caption{\textbf{Benchmark results on novel pose task.} State-of-the-art methods' rendering performance on novel poses for each benchmark split. We abbreviate \nv~\cite{lombardi2019neural} as `NV', \anerf~\cite{su2021nerf} as `AN', \nb~\cite{peng2020neural} as `NB', \aniN~\cite{peng2021animatable} as `AnN' and \hn~\cite{weng2022humannerf} as `HN'. Cell color \legendsquare{colorbest}~\legendsquare{colorsecond}~\legendsquare{colorthird} indicate the best, second best, and third best performance in the same split respectively.}~\label{tab:animation}
\end{center}
\end{table*}

\begin{figure*}[ht]
\centering
\includegraphics[width=0.95\linewidth]{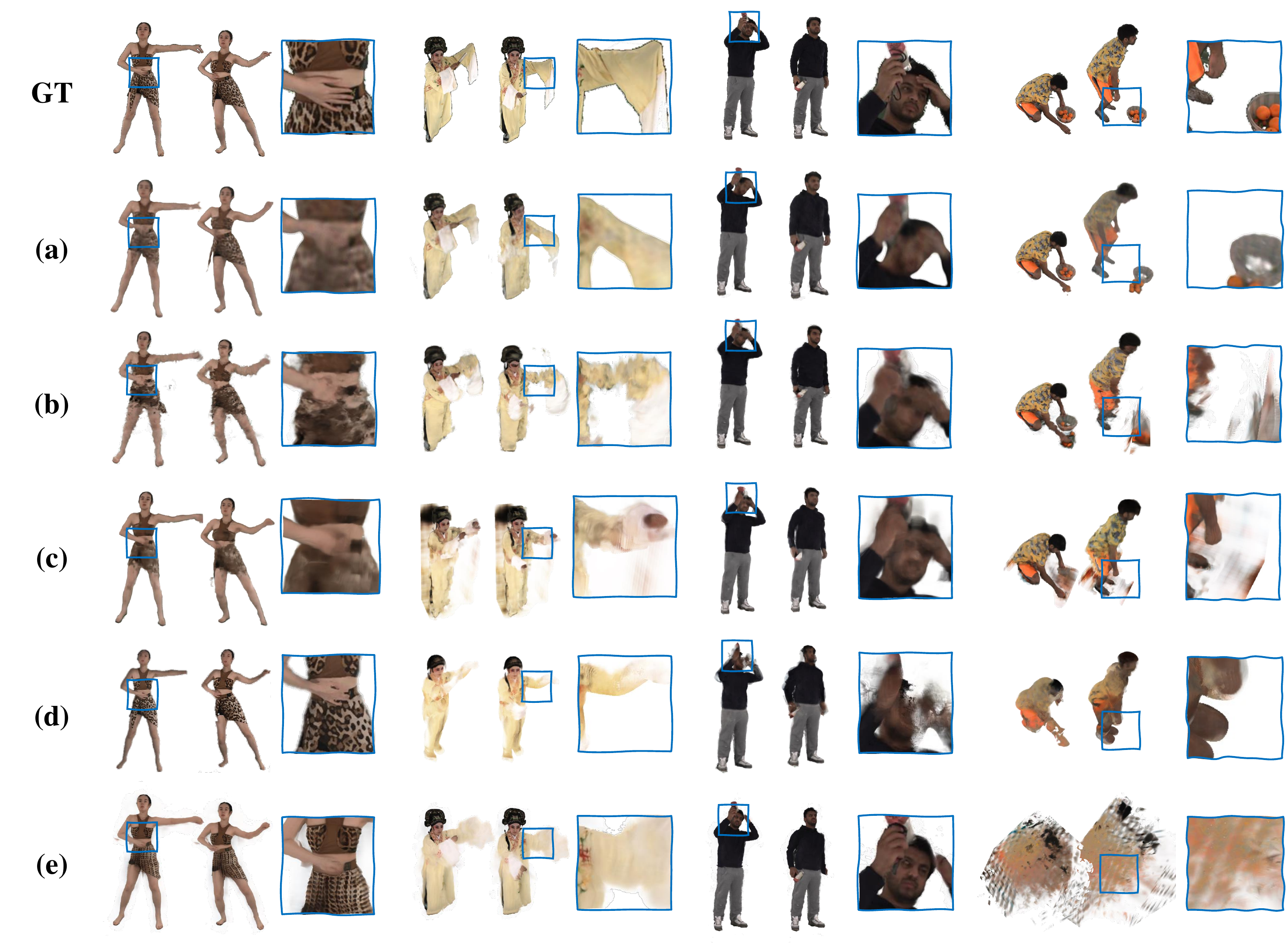}
\vspace{-0.5cm}
\caption{\textbf{Visualization of novel pose animation result samples.} From top to bottom, we illustrate the reposing results generated by \textbf{(a-e)}: \nv~\cite{lombardi2019neural}, \anerf~\cite{su2021nerf}, \nb~\cite{peng2020neural} \aniN~\cite{peng2021animatable}, and \hn~\cite{weng2022humannerf}. }\label{fig:novelpose}
\vspace{-0.5cm}
\end{figure*}

\subsection{Benchmark Results}

As introduced in Sec.~\ref{sec:data_split}, we construct a test set with $13$ sub-splits according to the four most concerned attributes (\textit{Deformation, Motion, Texture}, and \textit{Interaction}) in different difficulty levels (\textit{Easy, Medium}, and \textit{Hard}), and an extra \textit{No} level for \textit{Interaction}. This results in a data volume of $39$ motion sequences for testing.  For all rendered images, three metrics are computed -- PSNR, SSIM~\cite{wang2004image}, and LPIPS~\cite{johnson2016perceptual} (LPIPS* denotes LPIPS$\times1000$). We evaluate more than 10 state-of-the-art methods on these splits and analyze their performances under the same metrics. The experiment analysis is given below. Noted that due to limited space in the main paper, we provide the detailed setting, thorough discussions, and additional results in Sec.~\ref{sec:sup:benchmark} in the supplementary.

\noindent \textbf{Novel View Synthesis.} We visualize the bubble diagram of quantitative results across all benchmark splits in Fig.~\ref{fig:novel_view}. The precise numbers of the quantitative results are reported in Tab.~\ref{tab:novelview} in the supplementary. We conclude three key observations in the main paper: $(1)$ Generally speaking, the rendering quality is inversely proportional to split difficulties, as reported in Fig.~\ref{fig:novel_view}, where the circles get bigger when the difficulty grows. 
$(2)$ Among all case-specific dynamic methods, \anerf~\cite{su2021nerf} achieves the best PSNR performance, and \nb~\cite{peng2020neural} and \hn~\cite{weng2022humannerf} gets the best SSIM and LPIPS respectively. Qualitative results are shown in Fig.~\ref{fig:novelview}, \nb~\cite{peng2020neural} and \anerf~\cite{su2021nerf} could render novel view image with fewer background artifacts than other methods, while \hn~\cite{weng2022humannerf} can better preserve high fidelity textures, especially in high-frequency texture regions. $(3)$ When it comes to rendering novel views for trained human action frames, the difficulties increase on \textit{Motion} and \textit{Interaction} dimensions will not separate the performances among dynamic methods to a large extent, while hard \textit{Texture} cases enlarge the performance gap among dynamic methods (refer to T-shirt case with stripe pattern in Fig.~\ref{fig:novelview}). Meanwhile, dynamic methods' performances on~\textit{Texture} degrade the most when difficulty rises compared to the static baselines (refer to bubbles in Fig.~\ref{fig:novel_view}).  
\textit{Texture-Hard} and \textit{Deformation-Hard} shows the high texture and non-rigidity are still challenging for simple dynamic modeling with bone-coordinates~\cite{su2021nerf} or SMPL-skinning based blending~\cite{peng2021animatable}, even in seen-poses.
More qualitative results in each benchmark split are shown in Fig.~\ref{fig:novelview_supp}, and we analyze the conceptual difference of these methods in Sec.~\ref{sec:sup:benchmark-nvs} in the appendix.

\begin{table*}[b]
\resizebox{\textwidth}{!}{
\begin{tabular}{l|ccccc|ccccc|ccccc}
\hline
\multirow{2}{*}{\textbf{Splits}} &
\multicolumn{5}{c|}{\textbf{{PSNR$\uparrow$}}}&
\multicolumn{5}{c|}{\textbf{{SSIM$\uparrow$}}} &
\multicolumn{5}{c}{\textbf{{LPIPS*$\downarrow$}}} \\
& {IBR} & {PN} & {VN} & {NHP} & {KN} & {IBR} & {PN} & {VN} & {NHP} & {KN} & {IBR} & {PN} & {VN} & {NHP} & {KN} \\ 
 \hline
Motion-Simple & \best{26.13} & \second{26.04} & \third{25.90} & 25.61 & 24.67 & \second{0.964} & 0.957 & 0.959 & \third{0.961} & \best{0.964} & \second{65.48} & 72.68 & 72.29 & \third{65.53} & \best{44.77}  \\  
Motion-Medium & \second{25.56} & \best{25.84} & \third{25.22} & 24.63 & 24.34 & \second{0.966} & 0.960 & 0.961 & \third{0.963} & \best{0.971} & \second{59.71} & 63.80 & 64.83 & \third{61.08} & \best{38.22}  \\  
Motion-Hard & \third{23.78} & \best{24.49} & 23.72 & 23.43 & \second{24.36} & \second{0.959} & 0.950 & 0.949 & \third{0.956} & \best{0.973} & \second{79.18} & 89.93 & 93.04 & \third{80.20} & \best{43.79}  \\ \hline 
Deformation Simple  & \second{26.72} & \best{26.85} & 26.31 & \third{26.41} & 26.01 & \second{0.965} & 0.960 & 0.960 & \third{0.963} & \best{0.965} & \third{53.73} & 61.92 & 63.48 & \second{48.45} & \best{34.68}  \\  
Deformation-Medium & \third{27.46} & \second{27.55} & \best{27.90} & 27.28 & 25.83 & \third{0.965} & 0.961 & \second{0.965} & 0.963 & \best{0.972} & \third{57.30} & 66.82 & 62.82 & \second{56.02} & \best{38.00}  \\  
Deformation-Hard & \best{23.98} & 20.77 & 20.05 & \third{22.64} & \second{23.00} & \second{0.942} & 0.841 & 0.838 & \third{0.936} & \best{0.943} & \second{87.04} & 282.36 & 264.57 & \third{92.22} & \best{67.76}  \\ \hline 
Texture-Simple & \second{25.70} & \best{26.27} & \third{25.62} & 25.43 & 22.72 & \second{0.973} & 0.967 & 0.968 & \best{0.973} & \third{0.971} & \third{63.66} & 69.16 & 70.93 & \second{58.70} & \best{50.07}  \\  
Texture-Medium & 26.15 & \best{26.76} & \second{26.40} & \third{26.25} & 25.14 & \second{0.965} & 0.960 & 0.963 & \third{0.963} & \best{0.968} & \third{54.45} & 56.66 & 56.58 & \second{53.95} & \best{35.10}  \\  
Texture-Hard & 23.34 & \best{24.08} & \second{23.61} & \third{23.45} & 22.91 & \third{0.932} & 0.921 & 0.922 & \second{0.933} & \best{0.935} & \third{98.77} & 106.05 & 104.58 & \second{90.84} & \best{72.87}  \\ \hline 
Interaction-No & \second{26.08} & \third{25.91} & \best{26.39} & 25.46 & 23.43 & \second{0.968} & 0.958 & 0.963 & \third{0.966} & \best{0.968} & \second{61.99} & 72.67 & 66.91 & \third{64.35} & \best{44.95}  \\  
Interaction-Simple & \best{27.60} & 25.76 & \second{27.54} & \third{26.67} & 26.12 & \second{0.976} & 0.944 & 0.973 & \third{0.974} & \best{0.977} & \second{50.50} & 82.30 & 53.38 & \third{50.77} & \best{28.60}  \\  
Interaction-Medium & \third{24.04} & \best{24.44} & \second{24.09} & 23.61 & 22.70 & \second{0.950} & 0.937 & 0.941 & \third{0.947} & \best{0.950} & \second{84.64} & 96.53 & 93.92 & \third{86.29} & \best{63.72}  \\  
Interaction-Hard & \third{24.78} & \best{25.76} & \second{24.93} & 24.02 & 24.12 & \second{0.951} & 0.944 & 0.942 & \third{0.946} & \best{0.953} & \third{79.06} & 82.30 & 82.54 & \second{78.82} & \best{58.43}  \\ \hline 
Overall & \best{25.49} & \second{25.42} & \third{25.21} & 24.99 & 24.26 & \second{0.960} & 0.943 & 0.946 & \third{0.957} & \best{0.962} & \third{68.89} & 92.55 & 88.45 & \second{68.25} & \best{47.77} \\ \hline 
 
\end{tabular}
}
\caption{\textbf{Benchmark results on novel identity task.} State-of-the-art methods' performance on novel identity rendering in each benchmark split. We abbreviate \ibr~\cite{wang2021ibrnet} as `IBR', \pixel~\cite{yu2020pixelnerf} as `PN', \vision~\cite{lin2023visionnerf} as `VN', \nhp~\cite{kwon2021neural} as `NHP' and \kptnerf~\cite{weng2022humannerf} as `KN'. } 
\label{tab:novel-id-sup}
\end{table*}

\noindent \textbf{Novel Pose Animation.}
Similar to novel view synthesis, when split difficulty increases the rendering quality decreases as shown in Tab.~\ref{tab:animation}. Among all data factors, we found that \textit{Deformation} and \textit{Interaction} are insurmountable factors for current methods to model in novel poses. Qualitative results are displayed in Fig.~\ref{fig:novelpose}, none of the methods can generate reasonable deformation in the case of the Peking opera costume. \nb~\cite{peng2020neural} and \aniN~\cite{peng2021animatable} can not model the interactive objects, and the objects are stretched when given large poses in \anerf~\cite{su2021nerf}. Conclusively, current state-of-the-art methods can learn relatively reasonable human avatars with even hard \textit{Motion} and \textit{Textures}, while stuck in the imperfectness of modeling hard \textit{Deformation} and \textit{Interaction}. These two animation challenges should stimulate the communities for further investigation. More detailed analysis is provided in Sec.~\ref{sec:sup:benchmark-animation} in the appendix.

\noindent \textbf{Novel Identity Rendering.}
We report the quantitative metrics of all $39$ novel identities in Tab.~\ref{tab:novel-id-sup}. Generally, generalizable methods with human piror~\cite{mihajlovic2022keypointnerf,kwon2021neural} performs better with higher robustness than category-agnostic methods~\cite{yu2020pixelnerf,wang2021ibrnet,lin2023visionnerf}. Among category-agnostic methods, \ibr~\cite{wang2021ibrnet} directly blends pixel color from source views, and it outperforms \pixel~and \vision~ that predict radiance color only from image features.  We draw the conclusion that, in generalizable human rendering, human prior and appearance references from observation could help boost the generalization ability on data with large variations of poses and appearances.
We illustrate the qualitative results in Fig.~\ref{fig:figure-novel-id-exp-classes}. We provide additional results and analysis in Sec.~\ref{sec:sup:benchmark-generalize} in supplementary.

\subsection{Cross-dataset Comparison}
Apart from the benchmark experiments conducted on different data splits of \datasetname~dataset, we also evaluate the {\textit{data generalizability}} provided by our dataset and the other competitive ones, {\textit{i.e.,}} \genebody\cite{cheng2022generalizable}, \zjumocap~\cite{peng2020neural} and \humman~\cite{cai2022humman}. 

\noindent{\textbf{Setting and Implementations.}} To eliminate the scale and annotation differences across all datasets, we train three general scene generalizable rendering methods~\cite{yu2020pixelnerf,wang2021ibrnet,lin2023visionnerf} on these datasets with the same pixel batch per-iteration and stop training with the same $200K$ global iterations. For each method, we train each individual model on each dataset mentioned above, with a fixed image resolution $512\times512$ and four balanced views. To thoroughly evaluate the datasets' generalizability, we cross-verify the rendering images of {\textit{novel identities}} on each dataset. 

\begin{figure}[t]
\vspace{-0.5cm}
\centering
\includegraphics[width=0.8\linewidth]{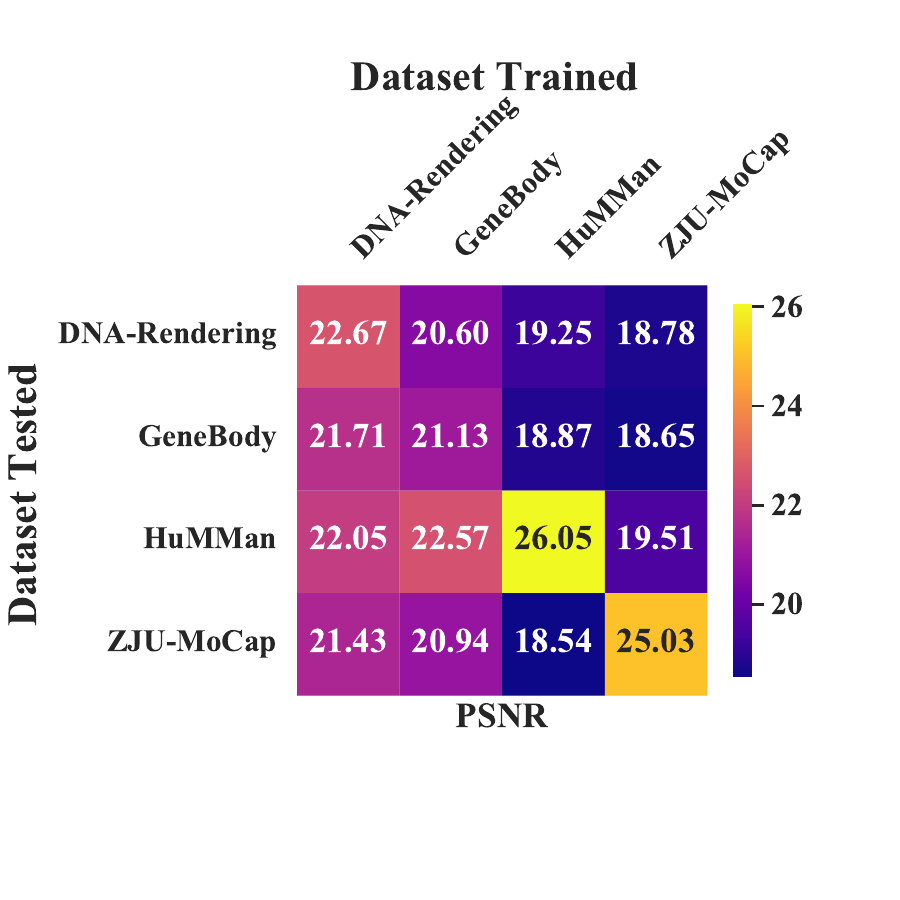}
\vspace{-0.6cm}
\caption{\small{\textbf{Results of cross-dataset experiments}. We visualize the `affinity' matrix of cross-dataset evaluation results.} }\label{fig:cross_dataset_exp}
\vspace{-0.5cm}
\end{figure}

\noindent{\textbf{Results.}} The experimental results are presented in Fig.~\ref{fig:cross_dataset_exp} in terms of the average PSNR of all three methods. From this colored error map, we conclude that training on \datasetname~dataset is beneficial for generalizing to the other datasets. Generally, due to the existence of domain gaps, a model would perform better in the situation of an in-domain setting, where the training set and test set follow the same distribution, see diagonal elements in Fig.~\ref{fig:cross_dataset_exp}. The off-diagonal numbers report the cross-domain performances of models trained on one dataset and tested directly on other datasets' test sets. We observe an interesting phenomenon that, compared to datasets with limited data diversity and high data bias (like \zjumocap~\cite{peng2020neural} and \humman~\cite{cai2022humman}), the proposed dataset enables generalization methods to achieve more plausible results even with large domain gaps. Moreover, opposite to \datasetname, \humman~\cite{cai2022humman} generalize poorly on other datasets even on cases with simple motions and appearances in \zjumocap~\cite{peng2020neural}, despite the fact that both \humman~\cite{cai2022humman} and our \datasetname~have large data volume. From a data engineering perspective, this demonstrates the construction of the proposed dataset benefits the community not merely with the amount of data, more importantly, the significant improvement in data completeness and richness. Due to space limit, we provide the detailed setup and additional results in Sec.~\ref{sec:sup:cross} of the supplementary. It is worth motioning that, we also unfold the generalization performance across testing cameras and reveal the impact of color consistency for multi-camera datasets in Sec.~\ref{sec:sup:cross-dataset-view}.

\section{Conclusion}\label{sec:conclusion}
We have presented \datasetname, a large-scale and high-fidelity repository for human-centric rendering. It is a multiview human body capture dataset that covers many diverse factors like ethnicity, age, body shape, clothing, motion, and interactive objects with faithful annotations. We have also presented benchmarks to evaluate state-of-the-art approaches on the \datasetname~dataset with in-depth discussions and compared our dataset with the others via cross-dataset experiments on generalization capability. 
We hope our \datasetname~project could boost the development of human-centric rendering and related domains with new reflections, challenges, and opportunities.

\noindent
\textbf{Acknowledgements.} This study is supported under the RIE2020 Industry Alignment Fund Industry Collaboration Projects (IAF-ICP) Funding Initiative, as well as cash and in-kind contributions from the industry partner(s). It is also partially supported by Singapore MOE AcRF Tier 2 (MOE-T2EP20221-0011, MOE-T2EP20221-0012), NTU NAP.

{\small
\bibliographystyle{ieee_fullname}
\bibliography{egbib}
}

\clearpage
\appendix

\noindent
\textbf{\LARGE Appendix}
\vspace{5ex}

\setcounter{equation}{0}
\setcounter{figure}{0}
\setcounter{table}{0}
\setcounter{section}{0}
\makeatletter
\renewcommand{\theequation}{S\arabic{equation}}
\renewcommand{\thefigure}{S\arabic{figure}}
\renewcommand{\thetable}{S\arabic{table}}

In the appendix, we provide detailed information about the proposed \datasetname~dataset and the attached benchmarks. We first provide dataset statistics, hardware design, and data collection protocol in Sec.~\ref{sec:sup:dataset}. Then, we discuss the additional information about the annotations, as well as a comparison to other publicly released toolchains in Sec.~\ref{sec:sup:annot}. Moreover, we conduct compact discussions on the benchmarks by introducing more detailed settings, additional results, and unfolded comparisons of benchmark methods' conceptual differences in Sec.~\ref{sec:sup:benchmark}. We provide an in-depth discussion on competing datasets and highlight our comparative contributions to society in Sec.~\ref{sec:sup:cross}. Finally, we discuss our future work in Sec.~\ref{sec:sup:future}.

\section{Dataset Details}\label{sec:sup:dataset}
\subsection{Dataset Statistics}\label{sec:sup:dataset-stats}
\label{section:datadistribution}


\begin{figure*}[hb]
\vspace{-4ex}
\centering
\includegraphics[width=\linewidth]{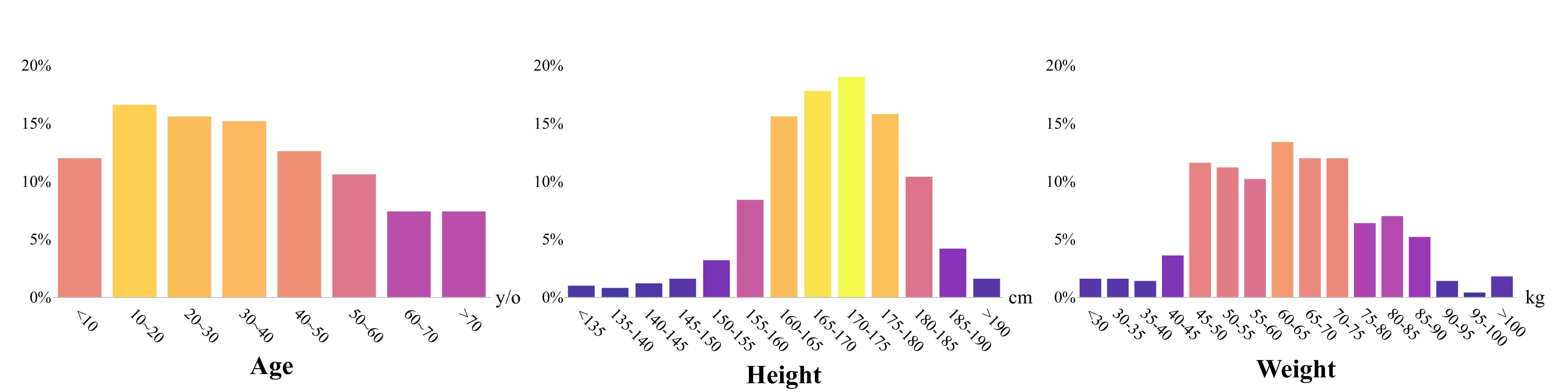}
\vspace{-5ex}
\caption{{\textbf{The distribution of actors' attributes.} We record the age, height, and weight of our invited actors. The statistical results reflect the wide range of the actors' personal attributes.} }
\label{fig:id_dis}
\vspace{-3ex}
\end{figure*}

\datasetname~has a wide distribution over ethnicity, clothing, actions, and human-object-interaction scenarios. In this section, we present the detailed data distribution in key data aspects, namely ethnicity, age, shape, actions, clothing, and interactive objects.

\noindent \textbf{Ethnicity, Age and Shape.}
We invite $500$ actors with a uniform distribution of gender and a ratio of $4:3:2:1$ for Asian, Caucasian, Black, and Hispanic individuals, respectively. The quota has a wide coverage of age and body shape. We visualize the distribution of actors' age, height, and weight in Fig.~\ref{fig:id_dis}.

\begin{figure*}[ht]
   \centering
   \begin{subfigure}[b]{0.48\textwidth}
     \centering
     \includegraphics[width=\textwidth]{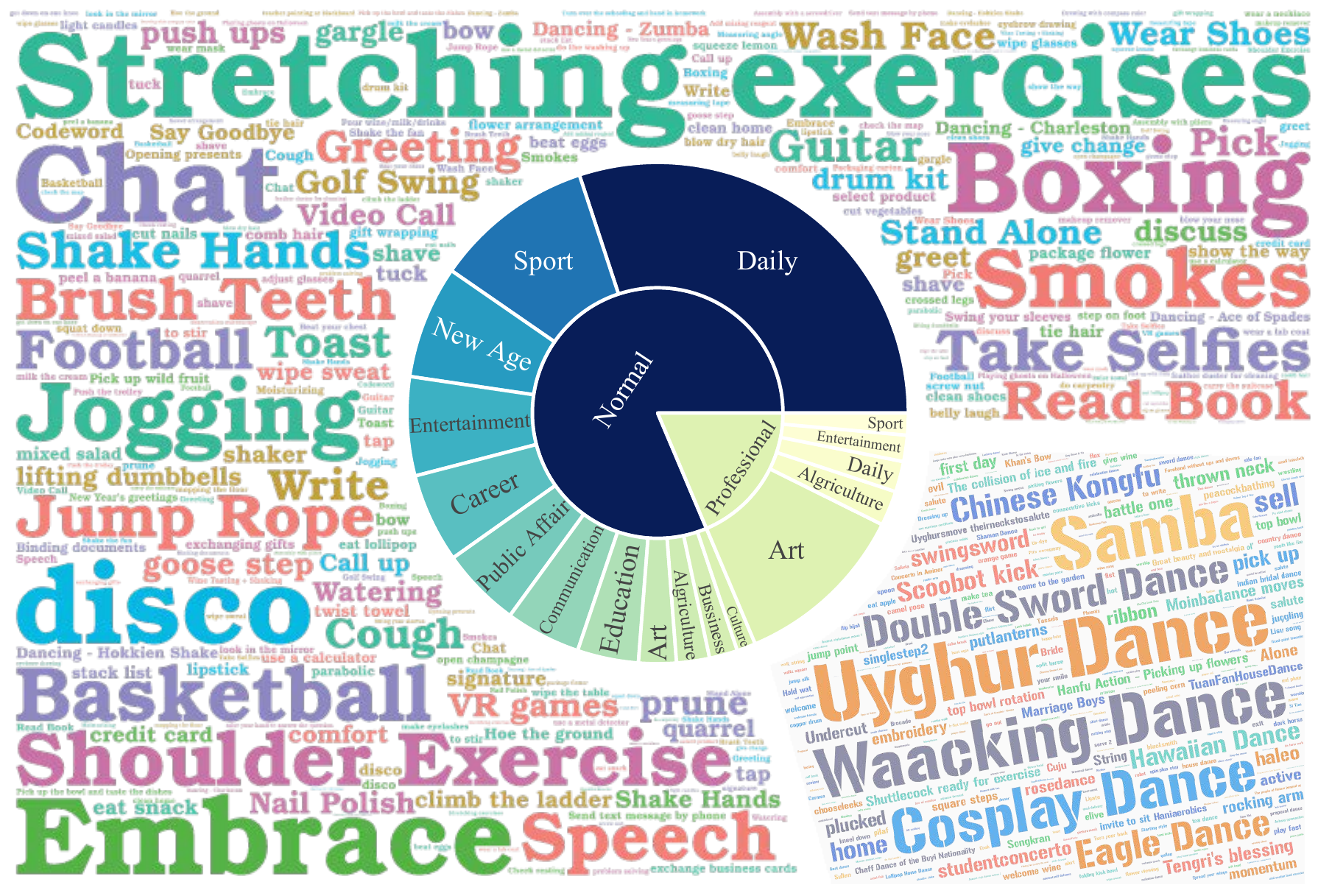}
     \caption{\textbf{Action distribution and labels}}
     \label{fig:stats-action}
   \end{subfigure}
   \hfill
   \begin{subfigure}[b]{0.48\textwidth}
     \centering
     \includegraphics[width=\textwidth]{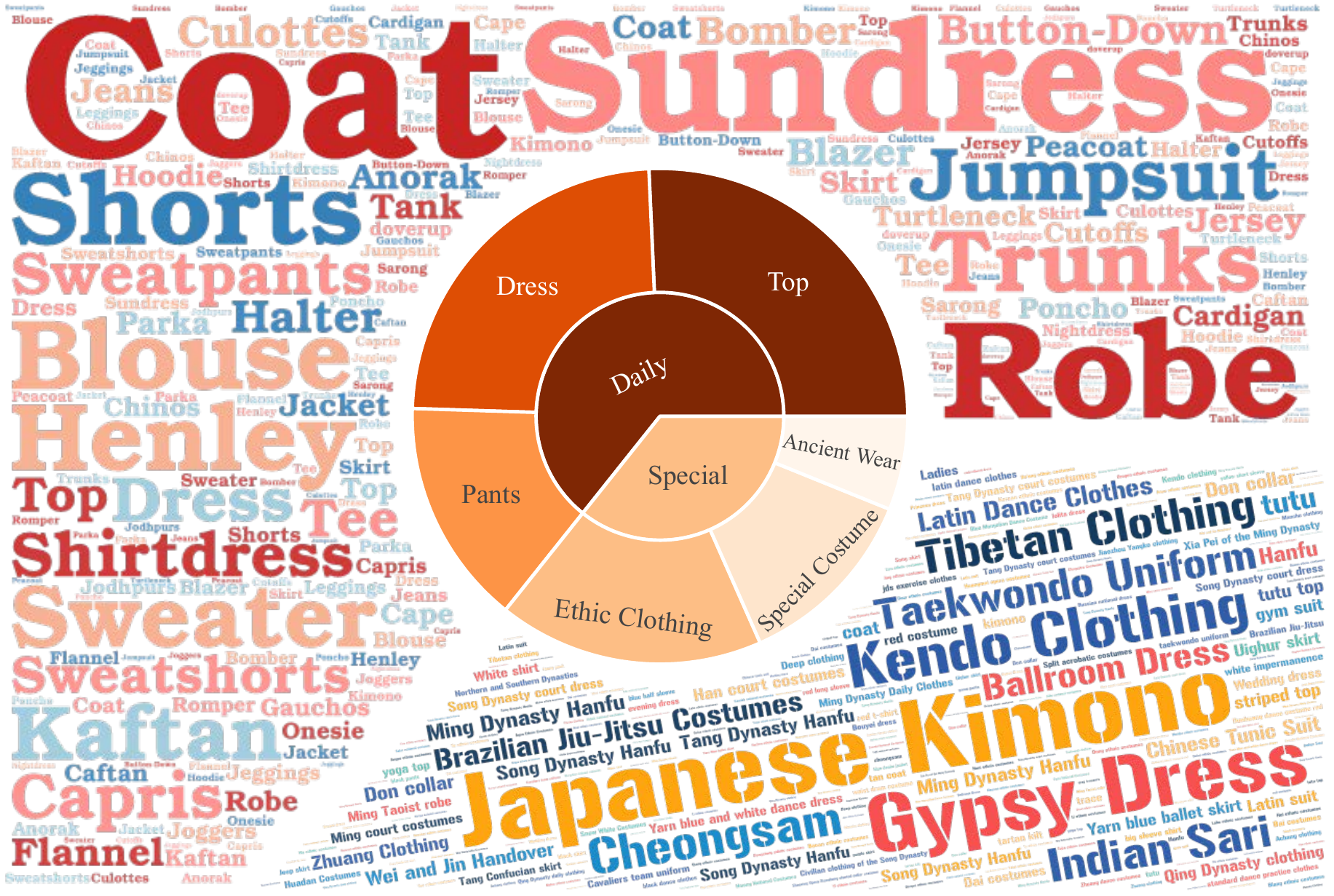}
     \caption{\textbf{Cloth distribution and labels}}
     \label{fig:stats-cloth}
   \end{subfigure}
   \hfill
    \caption{\textbf{Illustration of the action and clothes label distribution.} 
{(a)}. The distribution of action categories and sub-categories is visualized by a sunburst chart in the middle which is surrounded by the word cloud of normal and professional action labels. {(b)}. The distribution of clothes categories and labels are visualized in the same form with {(a)}.}\label{fig:cloact_dis}
\vspace{-3ex}
\end{figure*}

\noindent \textbf{Human Actions.}
\datasetname covers both normal actions and professional actions. We maintain a library of $269$ human action definitions, including daily-life activities, simple exercises, and social communication. All normal performers are asked to select $9$ actions from the action library and perform the picked actions in a free-style manner. There are $153$ professional actors among the total $500$ performers. These professional actors are asked to dress in their special costumes and perform $6$ unique professional actions with skills, including special costume performances, artistic movements, sports activities, {\textit{etc.}}. Note that different from the intuitive visualization in Fig.~\ref{fig:intro-large}, we visualize fine-grain categories of professional and normal action in Fig.~\ref{fig:stats-action}. These labels are classified in terms of a standard human activity subcategory definition\footnote{\href{https://en.wikipedia.org/wiki/Wikipedia:Contents/Human_activities}{https://en.wikipedia.org/wiki/Wikipedia:Contents/Human\_activities}}. The sunburst chart of distribution is visualized in the middle, and samples of specific categories of labels are visualized in the outer word cloud.

\noindent \textbf{Clothing and Interactive Objects.}
We create a clothing repository with $527$ items, which covers all $50$ clothing types in DeepFashion~\cite{liuLQWTcvpr16DeepFashion} while with a random distribution of color, material, texture, and looseness for each clothing type. We ask each performer to wear three sets of outfits, where one comes from the performer's self-prepared outfit (for both special and normal actors), and the other two are randomly coordinated from our clothing repository. The distribution of cloth statistical distribution on all action sequences and samples of cloth labels is illustrated in Fig.~\ref{fig:stats-cloth}. 


\subsection{Meta Attributes}\label{sec:sup:annot-meta}
\begin{figure}[hb]
\centering
\includegraphics[width=0.8\linewidth]{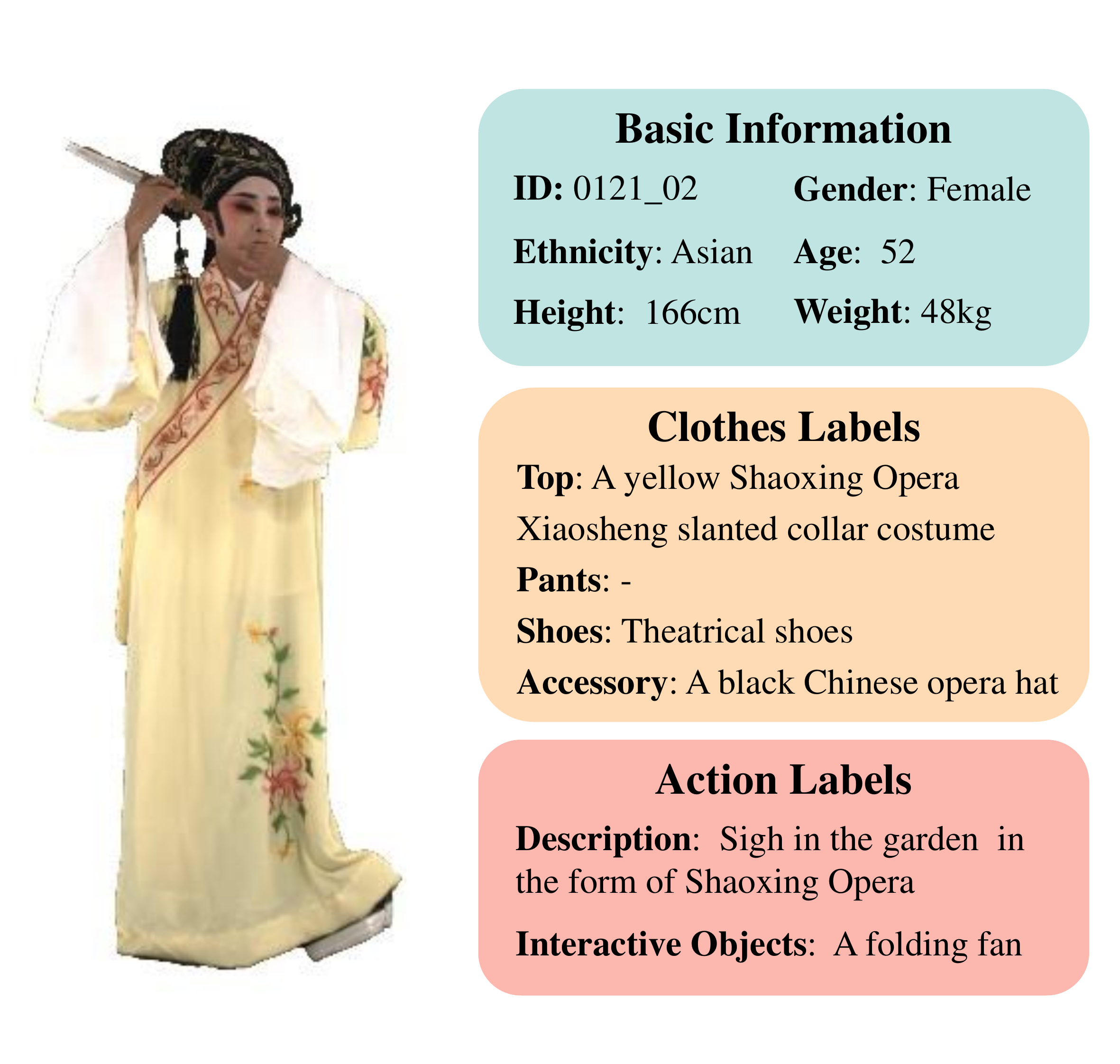}
\caption{\small{\textbf{An example of our meta attribute system.} We record the actors' basic information, costumes, and actions.}}\label{fig:MetaAttri}
\end{figure}

We have designed an attribute system for each dimension of the collected data, including basic information about the actors, clothing, and action information for each action sequence. Fig.~\ref{fig:MetaAttri} shows an example of meta attribute information of an action sequence for a professional actor. In terms of actor information, we record the actor's name, gender, ethnicity, age, height, and weight. For clothing information, we describe the upper and lower clothing, and shoe information. These descriptions include information on color, type, and other significant visual features. For action information, we describe the overall content of the action and, if there are any interactive objects, we also describe the type and other significant visual features of those objects.

\subsection{Hardware Construction}\label{sec:sup:dataset-system}
\label{section:data_system}

The main structure of \datasetname's capture system is a dome with a radius of three meters. The camera array built upon the doom consists of multiple types of cameras -- ultra-high-resolution $12$MP cameras, $5$MP industrial cameras, and Azure Kinect cameras. The lighting system provides natural lighting conditions. All cameras are triggered and synchronized by hardware, and synchronized multi-view data are transferred and recorded through our data streaming system.

\noindent \textbf{Camera System.} 
\datasetname~has $68$ cameras, including $12$ ultra-high resolution cameras with $12$MP resolution (short for $4096\times3000$ resolution), $48$ industrial cameras at $5$MP resolution (\textit{i.e.,} $2448\times2048$ resolution), and $8$ RGB-D Kinect cameras with depth resolution of $576\times640$. Specifically, the
$5$MP cameras are mounted on three cycles on the dome skeleton with $1$, $2$, and $3$ meters in height, each circle in height has $16$ balanced $5$MP cameras with $22.5$\textdegree~angle interval. $12$MP cameras are placed uniformly on another two intermediate height level circles, $1.5$ and $2.5$ meters height respectively. $12$MP cameras are installed with a $60$\textdegree~angle interval and interlaced with $5$MP cameras. The Kinect cameras are mounted close to the middle level of $5$MP cameras, providing the best RGB texture references for depth maps. Such construction of the camera array achieves dense coverage of the human body at multiple heights and angles. $5$MP cameras and $12$MP cameras are equipped with lenses of $8$ mm and $6$ mm respectively to achieve the best trade-off between full body proportion-in-view and size of capture volume. Note that, we use data captured from $12$MP and $5$MP cameras to construct our rendering dataset. The data captured from depth cameras comprise the auxiliary data which provide coarse geometry of human. Noted that we abandon the Kinect RGB cameras during the entire process, due to the bad color consistency.

\noindent\textbf{Lighting System.} 
Our lighting system consists of 16 flat light sources with a color temperature of $5600$K $\pm$ $300$K and an illuminance of $4500$ Lux/m. The lighting scale of each light source is $700\times500$ mm. There are eight flat lights on the ground installed with a $45$\textdegree~tilt towards actors in the middle to provide the best lighting on actors. There are extra eight flat lights hung on the roof to strengthen the lighting of upper body parts, especially for human heads. These uniformly distributed flat lights irradiate the whole scene with strong, natural, and balanced illumination.

\noindent\textbf{Data Streaming.} 
To collect, transfer and store the multi-view camera data, we construct a data streaming system that consists of two pieces of equipment for data synchronization -- a $10$ Giga-byte network, and a high data throughput workstation.
The camera system is synchronized by Kinect's trigger signal. First, eight Kinects are configured in a daisy chain and the out-trigger signal is converted to the TTL signal, and the other $60$ cameras are triggered by synchronization equipment. The $5$MP cameras are connected to six workstations via USB-$3.0$ ports and four-channel USB cards with PCI-E interfaces. The $12$MP cameras are connected to the other three workstations via $10$ GigE networks, capture cards, and PCI-E interfaces. 
To reduce active light interference of Kinect depth cameras, we adopt a $160$ $\mu$s time delay for each slave device on the chain. The maximum synchronization error of Kinect is $1.12$ ms in theory. The maximum synchronization error among all industrial cameras is less than $2$ ms, we measure this error by utilizing the image of high-speed flashing LED timer arrays and computing the displayed time differences.

\subsection{Data Collection Protocol}\label{sec:sup:dataset-capture}
We discuss the detailed data collection protocol from five aspects, {\textit{i.e.,}} data content, system check, core data collection, auxiliary data collection, and post-processing.

\noindent\textbf{Data Content.} 
During everyday data collection, we gather a comprehensive set of data sources, including action data, background data, actors' A-pose data for each outfit, extrinsic calibration data, and the record of performance attributes.  We collect intrinsic data and color calibration data only when we apply any modification to the system.

\noindent \textbf{System Check.} We conduct a daily system check before formal data collection. The process focuses on the verification of camera parameters and synchronization. Concretely, we will check $1)$ if the camera parameters remain the same with recorded optimal values (\textit{e.g.,} white balance, gamma, focal length, the valid field of view (FOV), {\textit{etc.}}). Checking these factors ensures capturing under excellent image quality and valid capture volume. $2)$ We monitor the network's condition and check the synchronization via a system probe using high-speed flashing LEDs. $3)$~Finally, we collect extrinsic camera calibration data via a standard data collection process that records the checkerboard rotating as described in Sec.~\ref{sec:annotation} in the main paper.

\noindent \textbf{Core Data Collection.} 
We invite 4-6 actors per day to perform actions in our studio in different appointed time slots. Once the actors arrive, we will briefly introduce the collection procedure and ask them to sign the authorization agreements first. If the actors agree, they are asked to prepare their outfits, makeup, and actions. Meanwhile, we record the basic information for each actor, including the height, weight, age, ethnicity, and other appearance attributes like the type, color, and material of his/her self-prepared outfit. After the preparation, we ask each actor to perform several actions in his/her self-prepared outfit. Specifically, a normal actor will pick at least three actions from our pre-defined daily activities and perform them in a free-style manner. A professional actor will wear a special outfit and perform at least six unique sets of footage that fit with the professional skill or costume. Then, we ask each actor to change his/her outfits with another two sets that are randomly coordinated from our clothing library. For each new outfit, the actor will perform another three different normal actions. To ensure the performed motion is rational and authentic enough, we will ask each performer to rehearse outside the studio before the formal shooting. After our staff confirms that the action is performed correctly, the actor will perform in the studio again for the formal data collection.

\noindent \textbf{Auxiliary Data Collection.} 
Aside from the core performance data collection, we also record auxiliary data, including the blank background data for the matting process, and A-pose data as a record for the canonical actor model before each round of new outfit recording. To record A-pose data, we require $1)$ the actor's hands tilt 45\textdegree downward the legs with clear distance; $2)$ the hands should slightly open without clenched fingers or put them together; $3)$ the farcical expression should keep expressionless with the eyes open and looking straight ahead. 

\noindent \textbf{Post-processing.}
After the action is completed, the center workstation generates fast multi-view preview videos for all cameras, and we check whether the performance content or the filming on each camera view meets the requirement. 
Actors are asked to re-play the performance if the recorded data is invalid. After collecting all qualified data, we post-process the data in per-day shooting volume  Image sequences in RAW format will be converted to the lossless BMP format, and then compressed into a video with a low constant rate factor with the x$264$ library. The processed data are then uploaded to the cloud server for subsequent dataset processing.

\subsection{Limitations on Data Collection}
\label{sec:sup:dataset-limitation}

\noindent \textbf{Data Content}.
To achieve high-fidelity data collection, we set the lighting with invariant and uniform illumination and set the acquisition frame rate to $15$ frames per second. We also constrain the field of view of the cameras, to ensure each one can capture the {\textit{full-body}} movements of a single actor (including the interacted object if there is one) while maintaining the FOV as max as possible. This allows us to capture details such as facial makeup and clothing textures.  In future work, we would update our hardware systems and upgrade our capture processes to accommodate different lighting conditions, multiple FOV ranges for multi-person scenes, high-speed capture of subtle movements, and multi-sensory (\textit{e.g.,} auditory, and tactile data) collection.

\noindent \textbf{Failure Cases}.
During the data collection process, various factors can lead to failure, such as large movements that exceed the field of view of the multi-camera system, or loss of frames due to large volume data transmission fluctuations. Following the standardized capture process, our operators will manually inspect the completeness and effectiveness of all camera data and actor movements after each capture cycle. If any issues are found, the hardware will be instantly checked and the data will be re-captured. Most failure cases are identified and promptly resolved at this stage.

\section{Data Annotation Details}\label{sec:sup:annot}

\subsection{Camera Intrinsic Calibration}\label{sec:sup:annot-calib}
\begin{figure}[h]
\centering
\includegraphics[width=\linewidth]{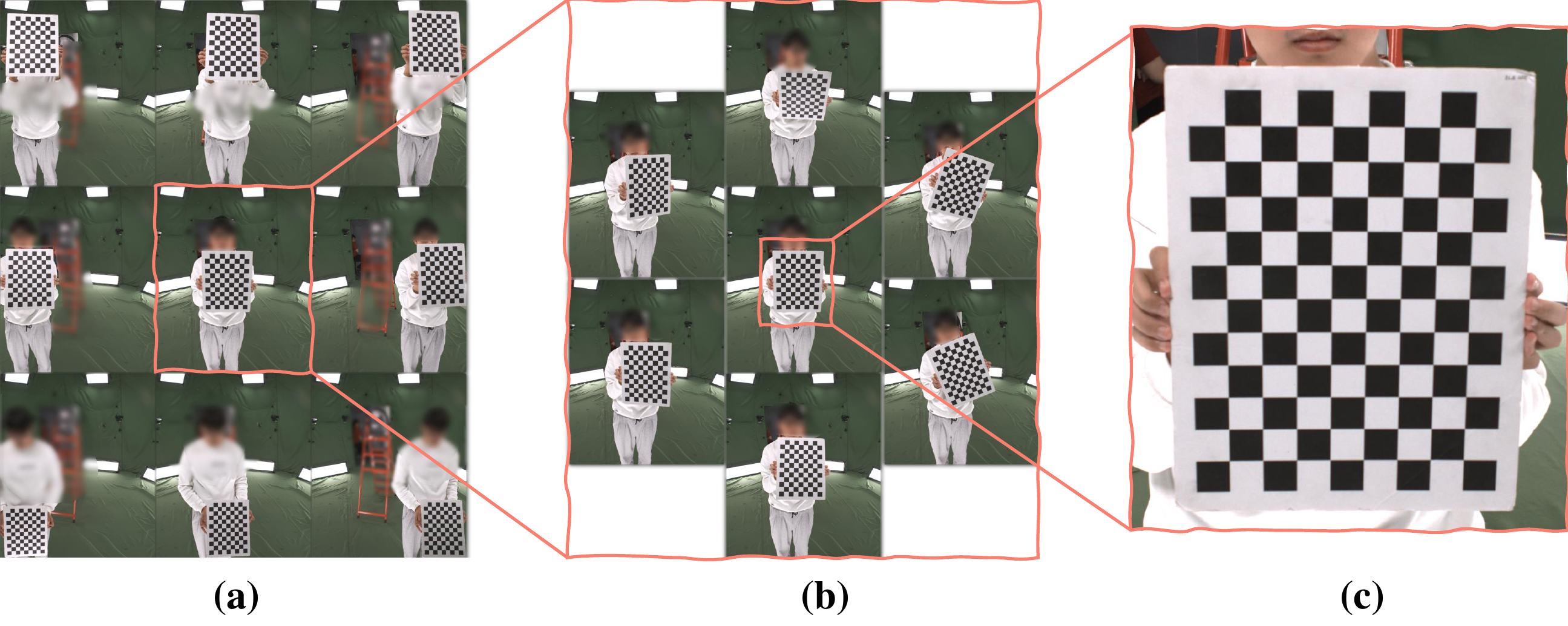}
\caption{\small{\textbf{Intrinsic calibration.} To ensure better distortion coefficient estimation, \textbf{(a)} we separate the camera view into $3\times3$ Soduku and capture the images of a checkerboard (about $1/4$ size of the one used in estimating extrinsic parameters) in every Soduku grid. \textbf{(b)} For each grid, we rotate the checkerboard at pitch, row, and yaw angles. This calibration step forces the checker to appear in every corner of the camera view. \textbf{(c)} Zoom-in for small-size checkboard for intrinsic calibration.
} }
\vspace{-2ex}
\label{fig:intrinsic}
\end{figure}

As this project targets capturing high-fidelity {\textit{whole-body}} data, we adopt a long lens that enlarges the human proportion in the camera view. This setup requires a high-quality estimation of camera distortion since some subjects' body parts might appear on the image boundary region. Thus, we use a $3\times3$ Soduku data collection protocol for intrinsic calibration, as illustrated in Fig.~\ref{fig:intrinsic}. Specifically, to maximize the checkboards' coverage across the whole image space, we separate the image into a $3\times3$ Soduku and capture the images of the checkerboards' movement in grids. For each grid, we  rotate the checkerboard in pitch, row, and yaw direction to enlarge the angle of the boards.

\subsection{Camera Color Calibration}\label{sec:sup:annot-colorcalib}
\begin{figure}[hb]
\centering
\includegraphics[width=0.98\linewidth]{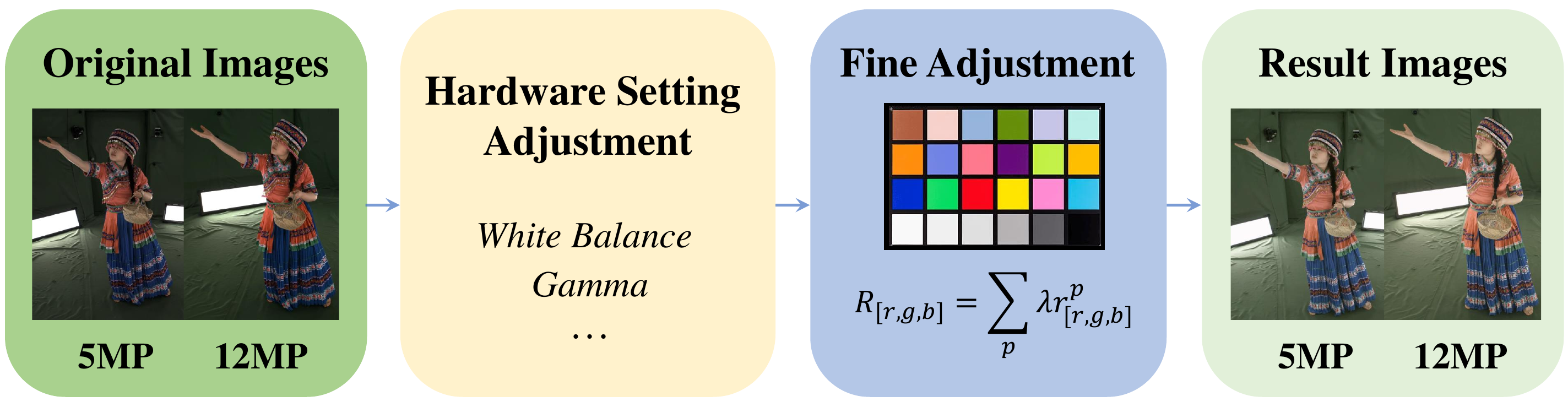}\\ \vspace{-0.5ex}
\includegraphics[width=0.98\linewidth]{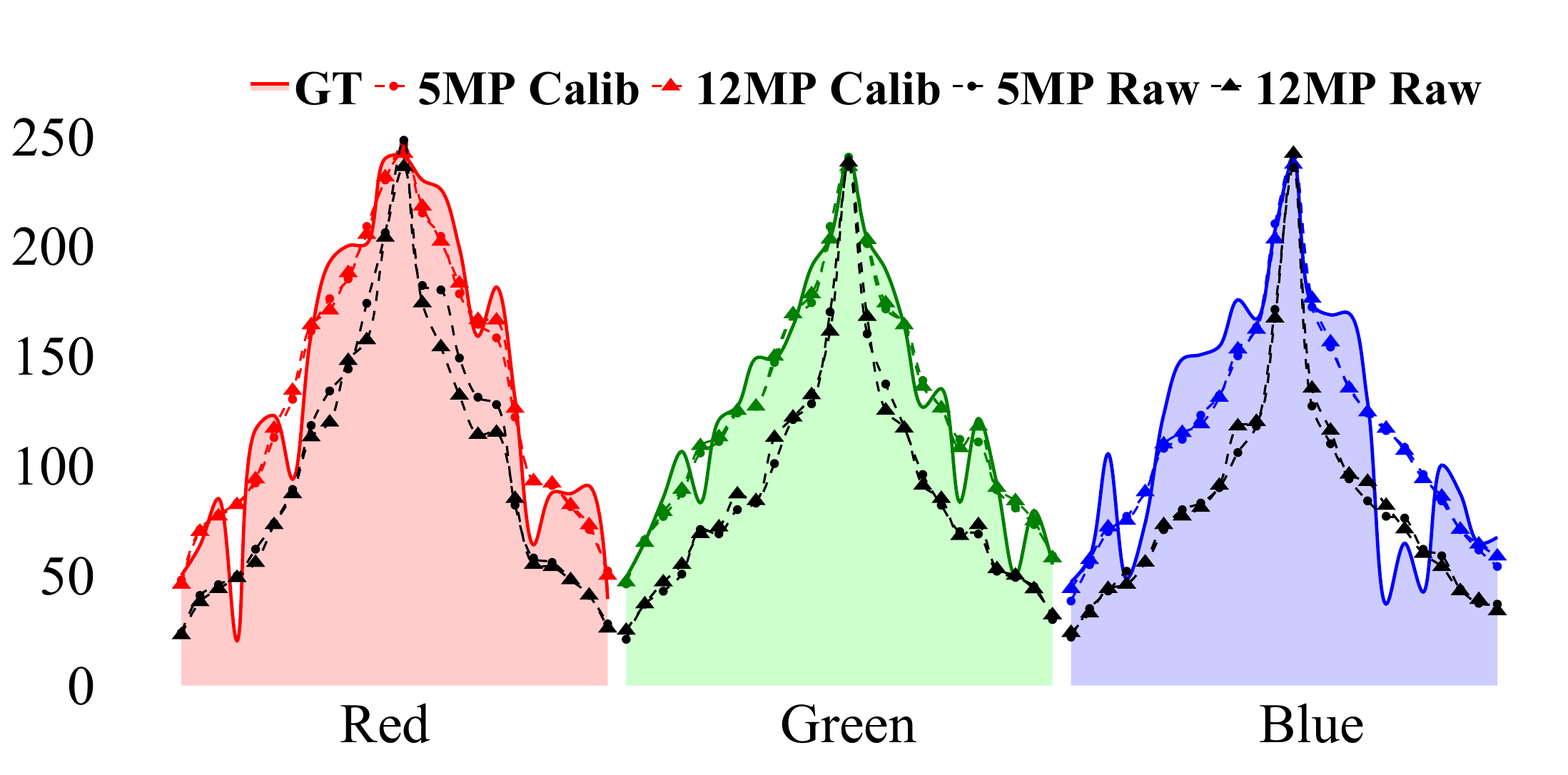}
\caption{\small{\textbf{Color calibration pipeline and calibrated color response.}The color-calibrated images are listed on the right of the flow chart (at the top of the figure). The color responses of two calibrated cameras (Camera $25$ which is a $5$MP camera, and Camera $51$ which is a $12$MP camera) compared with groundtruth color value of the color checkerboard are plotted below the flow chart, and smoothed spline curve is used. We show the `Raw responses' after hardware setting adjustment for reference. With the help of the color correction process, the average RGB value consistency between these two cameras $\Delta E_{00}$~\cite{luo2001development} is improved from $37.79$ to $4.15$. 
} }
\label{fig:colorcalib}
\vspace{-0.5cm}
\end{figure}

To ensure color consistency across multiple cameras, we inject a color calibration process into our data collection. A standard color board could be used as the criterion for f color calibration, and the fixed lighting condition in the dome could be treated as a standard condition during calibration. Specifically, the calibration lies in two aspects: $1)$ Hardware parameter adjustment. We make a rough adjustment on the hardware parameters to make the white balance and color balance of each camera as consistent as possible by human eyes; $2)$ Fine adjustment. Under a standard light source, we make the standard color board face straightforwardly to the camera to be calibrated at a constant distance, and a single image under this setting is collected; the corner detection algorithm is used to automatically identify the position of the color board in the image, and the color sampling is performed with the center radius $p=10$ pixels of each color square. The average color value is taken as the color sampling value. We carry out the polynomial projection of the color sampling value to the standard value via least squares. Note that, we calibrate in RGB form and take $n=2$ to prevent overfitting. The overall procedure and illustrated results are presented in Fig.~\ref{fig:colorcalib}.

\subsection{Keypoints}\label{sec:sup:annot-keypoint}

\begin{figure*}[ht]
\vspace{-3ex}
\centering
\includegraphics[width=\linewidth]{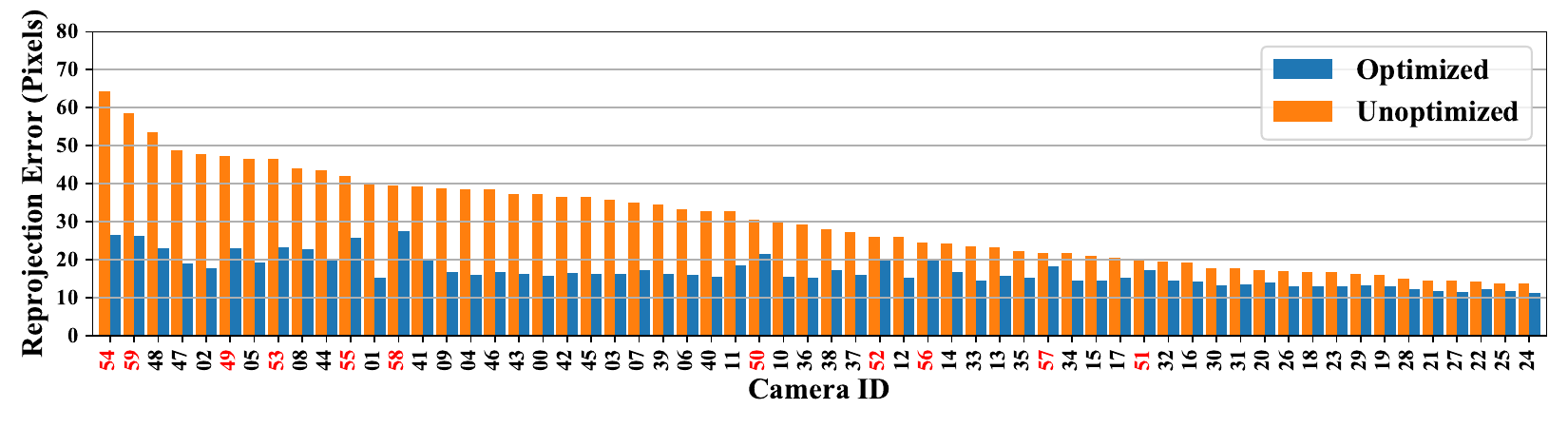}
\vspace{-5ex}
\caption{\small{\textbf{Evaluation of keypoint quality from every camera view.} We compute the mean reprojection error of \threeD~keypoints with \twoD~detection results. Optimization effectively reduces the error to below 30 pixels. Note that camera IDs 0-47 are $5$MP cameras, and camera IDs 48-59 are $12$MP cameras, high-lighted with red x-ticks. } }
\label{fig:reprojection_errors}
\vspace{-3ex}
\end{figure*}

We highlight that having a large number of camera views allows us to rectify the occasional failures of single-view \twoD~keypoint detection. For the more natural and stable \threeD~keypoints, 
we adopt the following optimization and post-processing strategies:
1) \emph{Keypoint selection.} We dynamically select views for each keypoint in the data sequence, which with the most confident score to the keypoint while ensuring the keypoint can be triangulated. 2) \emph{Bone length constraint.} The bone length is constrained with a fixed length. We use the median bone length after initial triangulation as the target in the optimization. Only the lengths of the main limbs are considered in this step. 3) \emph{Outlier removal.} As a post-processing pipeline, filter modules are designed based on human priors, including a \threeD~bounding box filter, a movement filter, and a relative position filter. \threeD~keypoints outliers, which are too far from the body trunk, move too fast between frames, or lead to an inconsistent relative position between frames are removed. An interpolation is applied to recover the missing keypoints. 
Such a post-processing scheme can assure reliable and consistent face and hand keypoints, even with large-scale occlusions.
As shown in Fig.~\ref{fig:reprojection_errors}, these optimization and post-processing strategies effectively reduce reprojection error compared to triangulation with all available \twoD~keypoints.

\subsection{Parametric Model}\label{sec:sup:annot-smpl}
 In our automatic parametric model annotation pipeline, body shape $\beta \in \mathbb{R}^{n \times 10}$ (or $\beta \in \mathbb{R}^{n \times 11}$ for children \cite{hesse2018learning,patel2021agora}) is first estimated based on the bone length calculated from \threeD~keypoints with the static and less challenging A-pose sequence. We use the estimated body shape parameters as initial values and optimize the full parametric model parameters including pose parameters (body pose, hand pose, and global orientation) $\theta \in \mathbb{R}^{n \times 156}$, and translation parameters $t \in \mathbb{R}^{n \times 3}$ ($n$ is the number of frames) via a modified SMPLify-X for other sequences with dynamic poses. The main energy terms in the optimization are keypoint energy $E_{\mathcal{P}}$, full-body joint angle prior energy $E_{a}$, bone length energy $E_{\mathcal{B}}$, and body shape prior energy $E_{\beta}$ \cite{pavlakos2019expressive,loper2015smpl, cai2022humman}. The main modification of SMPLify-X in our annotation pipeline is the decoupling body shape optimization and pose optimization, which we empirically find to produce more stable results.

Concretely, we employ bone length energy $E_{\mathcal{B}}$ and body shape prior energy $E_{\beta}$ to fine-tune body shape parameters for each sequence of a subject, with the same shape initialization from the A-pose static sequence. Body shape values are kept consistent throughout all the frames in a sequence.

\begin{equation}
    E_{shape}(\theta, \beta, t) = \lambda_{1} E_{\mathcal{B}} + \lambda_{2} E_{\beta}
\end{equation}

We then leverage keypoint energy $E_{\mathcal{P}}$ and full-body joint angle prior energy $E_{a}$ for pose optimization with body shape fixed.

\begin{equation}
    E_{pose}(\theta, \beta, t) = \lambda_{3} E_{\mathcal{P}} + \lambda_{4} E_{a}
\end{equation}

\begin{figure}[h]
\vspace{-2ex}
\centering
\includegraphics[width=0.98\linewidth]{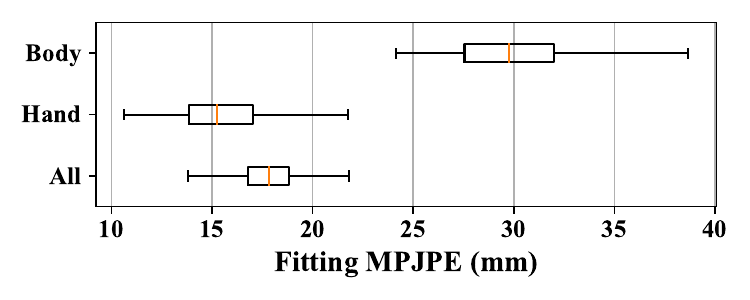}
\vspace{-3ex}
\caption{\small{\textbf{Evaluation of parametric model registration quality.} We evaluate body-only keypoints, hand-only keypoints, and all keypoints separately. The orange line indicates the median value, the box indicates the lower to the higher quartile, and the whiskers indicate the range of data.} }\label{fig:fitting_errors}
\vspace{-2ex}
\end{figure}

As shown in Fig.~\ref{fig:fitting_errors}, we evaluate the fitting error between \threeD~keypoints and corresponding regressed \smplx~joints. Body-only keypoints, hand-only keypoints, and all keypoints are evaluated separately. With our multi-view capture system and annotation pipeline, the MPJPE of body-only keypoints is on par with the optical motion capture system in Human3.6M \cite{ionescu2013human3, loper2014mosh}, MPJPE of hand-only keypoints is $15.87$mm.


\subsection{Comparison to Other SMPLX Fitting Methods}\label{sec:sup:annot-smplify}

\noindent \textbf{Baselines.}
To analyze the effectiveness of the proposed \smplx~fitting pipeline, we evaluate the accuracy of \smplx~fitting and compare it with three publicly available pipelines,{~\textit{i.e.,}}~the baseline \href{https://github.com/ZhengZerong/MultiviewSMPLifyX}{MultiviewSMPLifyX}~\cite{zheng2021pamir,pavlakos2019expressive}, \href{https://github.com/zju3dv/EasyMocap}{EasyMoCap}~\cite{easymocap,shuai2022multinb} used in~\zjumocap~\cite{peng2020neural,shuai2022multinb} dataset, and \href{https://github.com/generalizable-neural-performer/bodyfitting}{BodyFitting} used in~\genebody~\cite{cheng2022generalizable} dataset. Specifically, MultiviewSMPLify~\cite{zheng2021pamir} and BodyFitting~\cite{cheng2022generalizable} directly optimize the error of reprojected \threeD~SMPLX keypoints to \twoD~detections. 
Such a naive strategy is straightforward but lacks outlier robustness (might stuck in absolutely wrong detections or detection flip between left and right), and it is also computationally expensive. On the contrary, both EasyMoCap and the proposed pipeline adopt another strategy that separates the SMPLify process by a triangulation process.  This strategy optimizes \threeD~keypoints from \twoD~detection and then fits \smplx~from directly on optimized \threeD~keypoints. As  robust designs could be adapted during the triangulation process to eject outliers caused by flipping or occlusion, such a two-step strategy is faster and more robust to outliers. Whereas, one drawback is that the final \smplx~totally rely on the results of triangulation in the first stage by hand-crafted optimization and filtering. Compared to EasyMoCap, incorporate a more sophisticated designed 2D keypoints postprocessing phase, where movement filtering and relative position filtering are used when the given 2D keypoints are not accurate.

\begin{table*}[ht]
\begin{center}
\resizebox{0.9\linewidth}{!}{
\begin{tabular}{l||cccc|cccc|c}
\hline
\multirow{2}*{\textbf{Methods}}    & \multicolumn{4}{c|}{\textbf{\twoD~Reprojection Error (pixel)}} & \multicolumn{4}{c|}{\textbf{\threeD~MPJPE (mm)}} & \multirow{2}*{\textbf{Run Time (s)}} \\ 
& \multicolumn{1}{c}{\textbf{Body}} & \multicolumn{1}{c}{\textbf{Hand}} & \multicolumn{1}{c}{\textbf{Face}} & \multicolumn{1}{c|}{\textbf{Overall}} & \multicolumn{1}{c}{\textbf{Body}} & \multicolumn{1}{c}{\textbf{Hand}} &\multicolumn{1}{c}{\textbf{Face}} & \multicolumn{1}{c|}{\textbf{Overall}} & \\ \hline
MultiviewSMPLifyX~\cite{zheng2021pamir} & 
\third42.27 & \second33.36 & \second23.91 & \second28.77 &
55.46 & \second23.25 & \second22.24 & \second27.54 & 81.33 \\
BodyFitting (\genebody)~\cite{cheng2022generalizable} &
45.68 & \third33.43 & 34.17 & 35.67 & 
\third42.37 & 32.19 & \third30.67 & \third32.89 &\third29.50 \\
EasyMoCap (\zjumocap)~\cite{easymocap} &
\second32.71 & 33.75 & \third32.72 & \third33.64&
\second36.04 & \third25.37 & 38.10& 33.96 & \best0.69 \\
Ours & 
\best29.63& \best31.41& \best19.08& \best24.08 &
\best30.20& \best15.87& \best16.46& \best17.52 & \second3.23 \\
\hline
\end{tabular}
}
\end{center}
\vspace{-3ex}
\caption{\textbf{Comparison among multi-view \smplx~fitting methods.} Cell color \legendsquare{colorbest}~\legendsquare{colorsecond}~\legendsquare{colorthird} indicates the best, second best, and third best performance, respectively. Runtime in seconds indicates the average time required for the fitting process for one multi-view frame.}
\vspace{-3ex}
\label{tab:smplx_methdos}
\end{table*}

\noindent \textbf{Settings.}
For fairness, we use the same \twoD~keypoints consisting of human-inspected body labels and hands and faces auto-detection results. We also force all \smplx~models to have $10$ facial expression coefficiency and $45$ hand PCA components. We run the \smplx~fitting on our benchmark test data with their default \smplx~setting, namely with default penalty energies, their coefficient, and other settings except for the aforementioned modifications. We quantitatively evaluate the MPJPE of \threeD~keypoints and the reprojected \twoD~error across all views.

\noindent \textbf{Results.} The quantitive results are listed in Tab.~\ref{tab:smplx_methdos}, we also separately evaluate the accuracy on body, hand, face, and whole body. \threeD~MPJPE is computed by regressed \smplx~\threeD~keypoints to human-inspected \threeD~keypoints. \twoD~reprojection error is compared by reprojected \smplx~\threeD~keypoint to input \twoD~keypoints.
The runtime performances are also recorded, indicating the average fitting time usage (excluding other time namely, data IO, \textit{etc.}) for one multi-view frame. Our proposed pipeline outperforms other fitting methods in all categories in terms of both~\twoD~and \threeD~metrics, and has a more acceptable runtime requirement than EasyMocap~\cite{easymocap}.  Moreover, MultiviewSMPLify~\cite{zheng2021pamir,pavlakos2019expressive} achieves the second-best performance, while its time consumption is exploded by an order of magnitude. In a nutshell, our pipeline ensures the best trade-off between performance and efficiency.

\subsection{Matting and Segmentation Refine}\label{sec:sup:annot-matting}
\begin{figure}[ht]
\vspace{-2ex}
\centering
\includegraphics[width=0.9\linewidth]{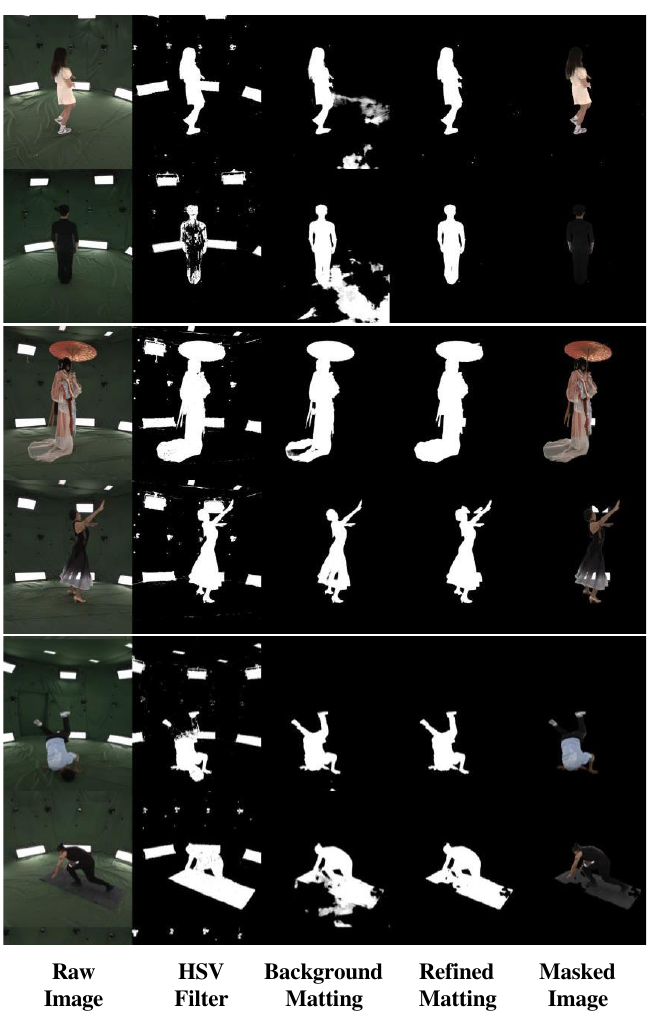}\\ \vspace{-1ex}
\includegraphics[width=0.9\linewidth]{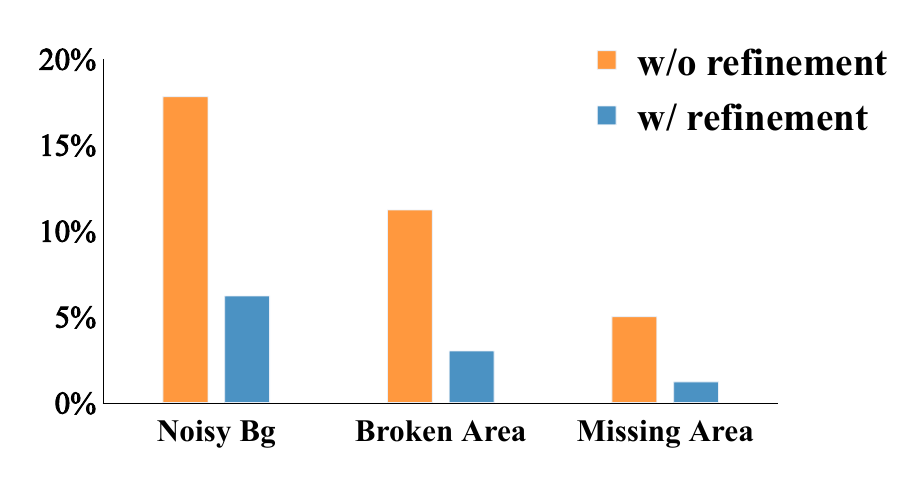}
\vspace{-2ex}
\caption{\textbf{Examples and statistics on matting refinement.} In the upper image, we show three kinds of common challenging cases and the comparisons among color filtering only, background matting only, and our optimized solution (\textit{Refined Matting}). From top to bottom cases, the problems (before optimization) are noisy backgrounds, broken body areas, and missing areas like body parts and incomplete objects; In the bottom figure, we show the error rates from the sampling survey of the three categories. }
\label{fig:matting_errors}
\vspace{-3ex}
\end{figure}

As described in the main text, despite the state-of-the-art background matting method~\cite{lin2021real} achieving impressive matting performance in the majority of our data, there are still several corner cases that fail to extract the foreground correctly, \textit{e.g}., noisy backgrounds, broken bodies, and missing bodies and objects. We demonstrate these most common corner cases in Fig.~\ref{fig:matting_errors}. To further improve the matting quality, we adopt the traditional computer vision algorithm- GraphCut~\cite{gortler1996lumigraph} to refine the predicted masks, and we find such a classical method plays a good fit to the CNN-based method which generates good results on these failure cases. 

In order to quantify the improvement, we introduce a manual inspection process to grade the results generated by CNN-based method~\cite{lin2021real} only and by a subsequent refinement procedure. Noted that due to the large scale of data, generating masks with manual labeling is impractical. More specifically, we ask three annotators to conduct such grading surveys on $500$ random multiview sequences. We report the error rates in terms of the type of corner cases in the bar chart in Fig.~\ref{fig:matting_errors}. From the human grading probe, we can conclude that with our proposed hybrid strategy, the error rates in all category decrease by a large margin compared with ~\cite{lin2021real} only, with the overall error rate reduced from $11$\% to $2$\%.

\subsection{Quality Control of Auto Annotation Results}\label{sec:sup:annot-quality}
To ensure the quality of annotation data, we conduct manual quality checks on the auto-annotated results. Specifically, we perform a human-in-the-loop quality evaluation for the \smplx~and matting results generated by the annotation pipeline. 

For \smplx~quality control, we overlay each \smplx~result on the original action data to create a multi-view video and manually verify the quality with the labeling task that requires our human annotator to grade the \smplx~quality. We subdivide the process into three stages.
$1)$ Binary filtering. If the \smplx~human body completely overlaps with the human body in the image or is within the natural shape range of the human body throughout the entire video, it is considered as a qualified \smplx~ annotation; otherwise, if there is severe misalignment or distortions on the main body, it is considered as an unqualified one. 
$2)$ Quality Grading. For qualified data cases, we further evaluate their subdivision quality, dividing them into five scores based on the unnaturalness of fingers or faces, the alignment of the head and shoulders with the image, {\textit{etc}}.
$3)$ Keypoints re-annotation. For unqualified cases, we ask the annotators to re-annotate the main skeleton in views with large errors by auto-annotators. The new annotation results are used to rerun the \smplx~results. We repeat the whole process until we achieve valid \smplx~models in all cases.

For Matting quality control, we manually evaluate the quality of the annotated video after matting by grading each video's quality. Quality is divided into three levels: A-level, where the entire human body is fully displayed without occlusion, and any interactive objects are fully shown, with no excess areas; B-level, where a small part of the human body or object is missing, or there are a few extra cutouts, with the erroneous pixel area not exceeding one-third of the human body area; and C-level, where there are serious problems and the erroneous pixel area exceeds one-third of the effective human body area. We treat A-level and B-level as acceptable mask annotations, while C-level as failure annotations. Noted that the cases in training and testing data split in our benchmark were manually selected to ensure high-quality annotations. For the sake of rigor, we will release the mask rating as confidence for mask annotation.

\section{Benchmark Details}\label{sec:sup:benchmark}
\label{sec:sup:benchmark}

\subsection{Methods Overview and Modifications}\label{sec:sup:methods}
In this subsection, we review the state-of-the-art methods benchmarked in this paper, and describe the major modification we made to the default implementation for adapting to the proposed dataset.

\subsubsection{Static Methods}\label{sec:sup:methods-static}
For static methods, we target to anchor the performances of novel view rendering on static test frames, which could be used as the baseline reference for dynamic methods on certain frozen times.

\noindent\textbf{\ngp}~\cite{muller2022instant} as an alter of NeRF~\cite{mildenhall2020nerf}, which utilizes the multi-resolution hash embedding and smaller network to accelerate the training and evaluation cost without loss of quality. Given the original implementation of \ngp~is under the underlying assumption of a moving single camera input or cameras sharing the intrinsic parameter across all camera positions, we modify it to suit multi-camera data with different intrinsic parameters.

\noindent\textbf{\neus}~\cite{wang2021neus} is a hybrid representation that combines neural radiance field with neural SDF, which produces better \threeD~reconstruction ability than NeRF-based methods~\cite{hedman2018instant,mildenhall2020nerf,peng2020neural} on existing datasets, while the rendered images of \neus~are typically not as sharp as NeRF-based methods. When adapting \neus~\cite{wang2021neus} on the proposed dataset, no special modification is required.

\subsubsection{Dynamic Methods}\label{sec:sup:methods-dynamic}
To construct the novel view synthesis and novel pose animation benchmark, we select the most recent state-of-the-art dynamic neural human rendering methods which can learn a neural body avatar from video sequences. 

\noindent\textbf{\nv}~\cite{lombardi2019neural} formulates a category-agnostic dynamic scene by a canonical voxel-grid decoder, and models the per-frame deformation as a mixture of affine warps that are parameterized by an auto-encoder with image input. 
Due to this property, we feed the network with $4$ balanced views of images from the training views. We also center and scale the camera system by $0.3$ to fit the voxel-grid system. During testing on novel pose, novel pose images of the $4$ view are input to the auto-encoder.

\noindent\textbf{\anerf}~\cite{su2021nerf} learns a human NeRF by conditioning the field with coordinates in each bone's local system. Note that its default setting only trains the network in the foreground, which usually leads to artifacts on the floor, we improve this by forcing pixel sampling on non-foreground space which helps to reduce the artifacts. 

\noindent\textbf{\nb}~\cite{peng2020neural} conditions a dynamic NeRF by time codes as well as structural latent features by sparsely convolving parametric model's vertices in \threeD. To run \nb~\cite{peng2020neural}, we transform our standard definition of the parametric model to EasyMoCap~\cite{easymocap} style, and we train the network using $42$ dense views. Note that during novel pose estimation, we fed the network with novel pose SMPLs and linearly extrapolate the time step.

\noindent\textbf{\aniN}~\cite{peng2021animatable} introduces neural blend weights with \threeD~human skeletons to generate observation-canonical correspondences in dynamic human NeRF. We do the same transformation like \nb~\cite{peng2020neural}.

\noindent\textbf{\hn}~\cite{weng2022humannerf} learns a dynamic neural human model from monocular video. It decouples the motion field by a corrected skeleton movement and non-rigid motion. Different from its original setting, we train the model with dense views by stacking multiview video sequences. It is~\textit{important} to point out that \hn~\cite{weng2022humannerf} models may collapse on certain data sequences producing meaningless images like the left bottom case in Fig.~\ref{fig:novelpose}. Such a phenomenon is consistent even with multiple trials of random initialization. Considering to deliver a more straightforward metric meaning in the benchmark, the report numbers of \hn~in Tab.~\ref{tab:novelview} and Tab.~\ref{tab:animation} only include the valid models.

\subsubsection{Generalizable Methods}\label{sec:sup:methods-generalizable}
For novel identity generalization, we evaluate three category-agnostic methods~\cite{yu2020pixelnerf,wang2021ibrnet,lin2023visionnerf} and two methods with human structure priors~\cite{kwon2021neural,mihajlovic2022keypointnerf}. 

\noindent\textbf{\pixel}~\cite{yu2020pixelnerf} is one of the first generalizable NeRFs that generalize novel objects' color and opacity by pixel-aligned feature-conditioned NeRF. We train it on our dataset with 4 selected views and fuse the multiview image feature with average pooling.

\noindent\textbf{\ibr}~\cite{wang2021ibrnet} predicts the radiance color of novel objects by blending observed color from source views, and inference the opacity from multiview feature fusion.

\noindent\textbf{\vision}~\cite{lin2023visionnerf} upgrades \pixel's~\cite{yu2020pixelnerf} image encoder with a global transformer~\cite{vaswani2017attention} and fuse the multi-level features with \twoD~CNN features. Like \pixel~\cite{yu2020pixelnerf}, we fuse the multiview feature with average pooling.

\noindent\textbf{\nhp}~\cite{kwon2021neural} combines key components of \pixel~\cite{yu2020pixelnerf} and \nb~\cite{peng2020neural}, and fuse them with multiview transformer and predict the radiance of human body. Noted there is a slight contradiction between the technical paper and the released implementation on the window size of the temporal transformer. We follow the open-sourced implementation and set the window size to 1 to avoid memory explosion which means the temporal transformer is a dummy module.

\noindent\textbf{\kptnerf}~\cite{mihajlovic2022keypointnerf}  use \ibr~\cite{wang2021ibrnet} as the backbone, and tailors  \threeD~keypoints as human prior into the framework. It conditions the radiance field with relative depth to every \threeD~keypoint in each source camera coordinate. We train the network with $24$ SMPL main skeleton keypoints. Noted that different from other generalizable methods which allow arbitrary resolution rendering, \kptnerf~\cite{mihajlovic2022keypointnerf} is suited to render square-sized images with $2^n$ width and height. We render out a minimum squared image that can cover the desired resolution then crop out the valid part.

\noindent

\subsection{Benchmark Details}\label{sec:sup:benchmark-detail}

As a supplement to the benchmark part in the main paper, we describe the detailed benchmark settings and additional analysis of results.
\subsubsection{Novel View Synthesis}\label{sec:sup:benchmark-nvs}
\noindent \textbf{Detailed Settings.}
As we described the task in Sec.~\ref{sec:task-definition} and reviewed the methods in Sec.~\ref{sec:sup:methods}, we evaluate the dynamic methods' novel view synthesis ability on the benchmark test set, which consists of $13$ splits with $39$ performance sequences. During training, we train models on each sequence separately, with the $42$ multiview (training views) image sequences of the first $80\%$ frames. Evaluations are performed on the same seen human poses but with every $45$ frame skip and only calculated on $18$ unseen camera poses. For static methods, we train separate models on multi-view images of each single frame. To evaluate the high-fidelity rendering of these benchmark methods, we train and test the models on \textit{half of the origin resolution}, namely $1024\times1224$ and $1500\times2048$ for 5MP and 12MP images respectively.

\begin{table*}[hb]
\begin{center}
\resizebox{\linewidth}{!}{
\begin{tabular}{l|cc:ccccc|cc:ccccc|cc:ccccc} \hline
\multirow{2}*{\textbf{Splits}} & \multicolumn{7}{c|}{\textbf{PSNR$\uparrow$}} &  \multicolumn{7}{c|}{\textbf{SSIM$\uparrow$}} &  \multicolumn{7}{c}{\textbf{LPIPS*$\downarrow$}} \\  
& \multicolumn{1}{c}{NGP} & \multicolumn{1}{c:}{NS} & \multicolumn{1}{c}{NV}  & \multicolumn{1}{c}{AN} & \multicolumn{1}{c}{NB} & \multicolumn{1}{c}{AnN}  & \multicolumn{1}{c|}{HN}   & \multicolumn{1}{c}{NGP} & \multicolumn{1}{c:}{NS} & \multicolumn{1}{c}{NV}  & \multicolumn{1}{c}{AN} & \multicolumn{1}{c}{NB} & \multicolumn{1}{c}{AnN}  & \multicolumn{1}{c|}{HN} & \multicolumn{1}{c}{NGP} & \multicolumn{1}{c:}{NS} & \multicolumn{1}{c}{NV}  & \multicolumn{1}{c}{AN} & \multicolumn{1}{c}{NB} & \multicolumn{1}{c}{AnN}  & \multicolumn{1}{c}{HN}
 
 \\ \hline
Motion-Simple & 30.97 & 27.49 & \second{27.85} & \best{29.15} & \third{27.84} & 25.89 & 25.49 & 0.979 & 0.973 & 0.966 & \third{0.974} & \best{0.978} & \second{0.974} & 0.955 & 31.52 & 44.18 & 57.74 & \best{52.75} & \second{53.32} & \third{56.10} & 62.08  \\  
Motion-Medium & 31.40 & 30.04 & \second{28.16} & \best{29.07} & \third{27.47} & 24.93 & 24.80 & 0.980 & 0.980 & 0.970 & \second{0.975} & \best{0.981} & \third{0.971} & 0.966 & 25.12 & 31.33 & 50.03 & \second{45.28} & \third{48.12} & 56.43 & \best{33.66}  \\  
Motion-Hard & 29.05 & 28.49 & \second{26.10} & \best{27.55} & \third{25.16} & 24.54 & 22.93 & 0.972 & 0.976 & 0.959 & \third{0.967} & \second{0.976} & \best{0.976} & 0.964 & 41.35 & 40.13 & 77.54 & \third{69.75} & 71.25 & \second{63.35} & \best{53.42}  \\ \hline 
Deformation-Simple  & 31.63 & 28.01 & 28.09 & \best{29.63} & \third{28.18} & 27.42 & \second{28.30} & 0.981 & 0.972 & 0.968 & \third{0.975} & \best{0.976} & \second{0.975} & 0.974 & 29.02 & 42.62 & 48.17 & \second{41.68} & 48.18 & \third{45.44} & \best{23.70}  \\  
Deformation-Medium & 30.01 & 29.65 & \second{29.77} & \best{30.52} & \third{29.22} & 26.29 & 26.60 & 0.972 & 0.975 & 0.971 & \second{0.974} & \best{0.979} & \third{0.972} & 0.963 & 41.14 & 37.18 & \second{39.36} & \third{43.51} & 46.95 & 52.69 & \best{29.12}  \\  
Deformation-Hard & 29.79 & 30.80 & \second{27.19} & \best{28.11} & \third{24.77} & 21.93 & 21.48 & 0.967 & 0.973 & 0.954 & \third{0.957} & \best{0.969} & \second{0.958} & 0.934 & 46.09 & 44.83 & \best{70.04} & \second{70.16} & \third{81.14} & 84.59 & 83.36  \\ \hline 
Texture-Simple & 30.53 & 31.39 & \third{27.85} & \best{30.45} & \second{29.13} & 25.36 & 27.39 & 0.978 & 0.988 & 0.974 & \second{0.984} & \best{0.988} & 0.979 & \third{0.979} & 36.02 & 23.53 & 58.78 & \third{43.09} & \second{41.53} & 53.69 & \best{24.72}  \\  
Texture-Medium & 30.85 & 31.33 & \third{28.50} & \best{30.53} & \second{29.62} & 22.46 & 27.40 & 0.978 & 0.982 & 0.968 & \second{0.977} & \best{0.984} & 0.959 & \third{0.971} & 29.32 & 27.04 & 47.33 & \second{37.99} & \third{41.46} & 75.08 & \best{25.01}  \\  
Texture-Hard & 29.16 & 28.23 & \second{26.73} & \best{27.36} & \third{25.69} & 19.98 & 24.78 & 0.966 & 0.956 & 0.942 & \third{0.947} & \best{0.963} & 0.946 & \second{0.950} & 36.48 & 58.55 & 79.68 & \third{79.17} & \second{77.36} & 94.60 & \best{34.66}  \\ \hline 
Interaction-No & 31.31 & 31.90 & \best{29.05} & \second{28.71} & \third{27.77} & 23.82 & 26.77 & 0.978 & 0.985 & \third{0.972} & \second{0.975} & \best{0.984} & 0.971 & 0.971 & 34.00 & 23.65 & \third{50.36} & 56.07 & \second{49.79} & 67.40 & \best{30.00}  \\  
Interaction-Simple & 31.55 & 32.13 & \third{29.09} & \best{30.12} & \second{29.56} & 25.93 & 28.59 & 0.982 & 0.987 & 0.976 & \second{0.981} & \best{0.987} & 0.977 & \third{0.978} & 27.58 & 21.79 & \third{45.55} & 46.48 & \second{41.69} & 57.18 & \best{22.00}  \\  
Interaction-Medium & 28.82 & 27.15 & \third{25.59} & \best{25.72} & \second{25.65} & 22.02 & 23.51 & 0.967 & 0.968 & 0.955 & \third{0.956} & \second{0.975} & \best{0.997} & 0.953 & 43.33 & 52.52 & \third{73.05} & 85.50 & \second{68.03} & 92.90 & \best{54.39}  \\  
Interaction-Hard & 30.03 & 29.29 & \second{28.09} & \best{28.25} & \third{25.00} & 22.92 & 23.87 & 0.972 & 0.977 & \third{0.962} & \second{0.964} & \best{0.972} & 0.961 & 0.951 & 43.64 & 39.71 & \second{57.97} & \third{63.42} & 71.00 & 81.49 & \best{57.38}  \\ \hline 
Overall & 30.39 & 29.68 & \second{27.85} & \best{28.86} & \third{27.31} & 24.11 & 25.53 & 0.975 & 0.976 & 0.964 & \third{0.970} & \best{0.978} & \second{0.970} & 0.962 & 35.74 & 37.47 & 58.12 & \second{56.53} & \third{56.91} & 67.76 & \best{41.04} \\ \hline

\end{tabular}
}
\caption{\textbf{Benchmark results on novel view synthesis task.} State-of-the-art methods' performance of novel test views on seen poses in each benchmark split. We abbreviate \ngp~\cite{muller2022instant} as `NGP', \neus~\cite{wang2021neus} as `NS', \anerf~\cite{su2021nerf} as `AN', \nv~\cite{lombardi2019neural} as `NV', \nb as `NB'~\cite{peng2020neural}, \aniN~\cite{peng2021animatable} as `AnN', and \hn~\cite{weng2022humannerf} as `HN'. Cell color \legendsquare{colorbest}~\legendsquare{colorsecond}~\legendsquare{colorthird} indicate the best, second best, and third best performance in the same split respectively. We exclude the static methods NGP and NS during ranking and separate them with dash lines.}~\label{tab:novelview}
\vspace{-5ex}
\end{center}
\end{table*}

\begin{figure*}[hbtp]
\vspace{-4ex}
\centering
\includegraphics[width=0.96\linewidth]{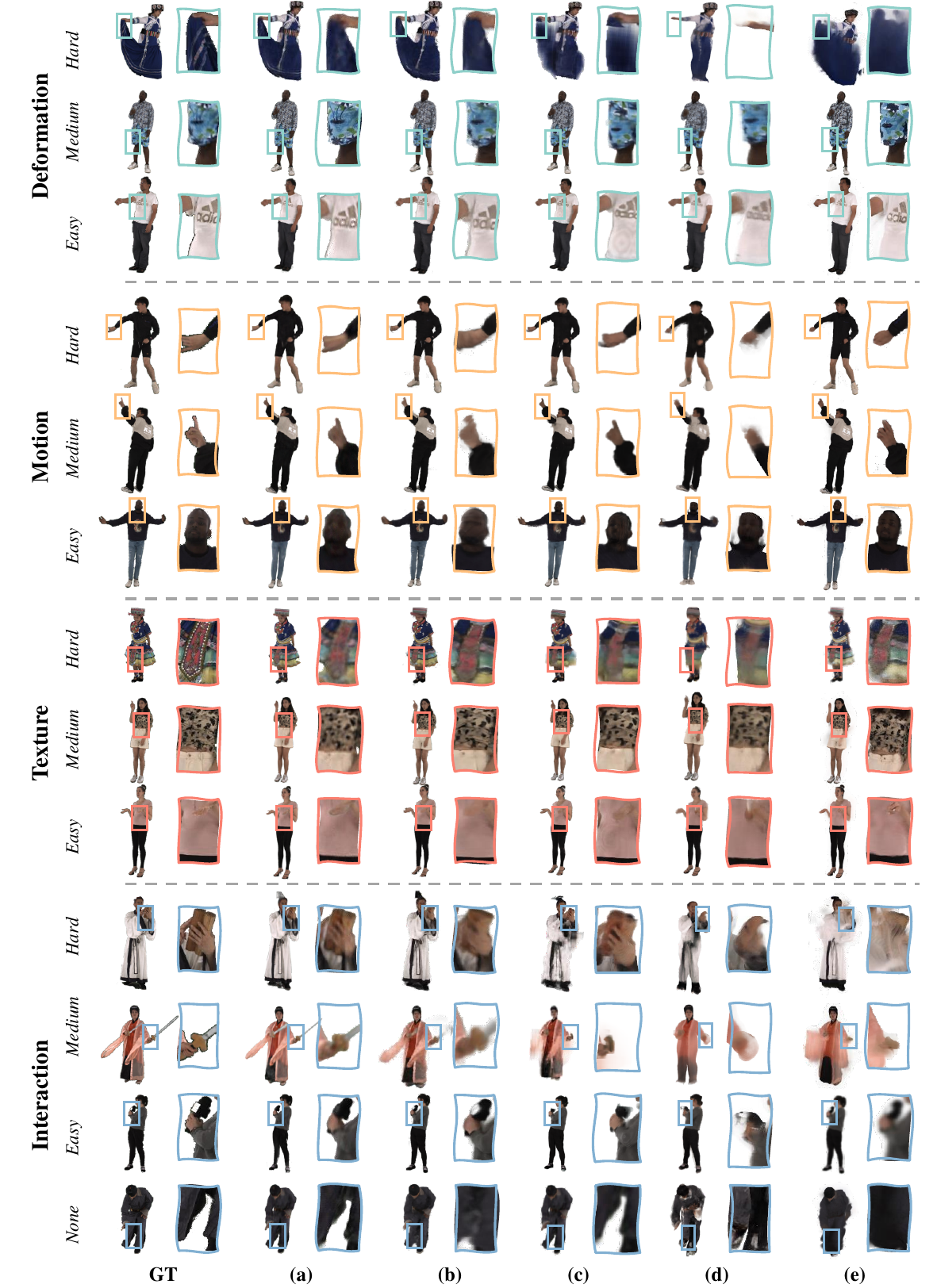}
\vspace{-2ex}
\caption{\small{\textbf{Novel view synthesis results on each data split}. From top to bottom, we illustrate the rendering results generated by \textbf{(a-e)}. \nv~\cite{lombardi2019neural}, \anerf~\cite{su2021nerf}, \nb~\cite{peng2020neural}, \aniN~\cite{peng2021animatable}, \hn~\cite{weng2022humannerf}. Please zoom in for better visualization.} }\label{fig:novelview_supp}
\end{figure*}

\begin{figure*}[hbtp]
\vspace{-5ex}
\centering
\includegraphics[width=0.96\linewidth]{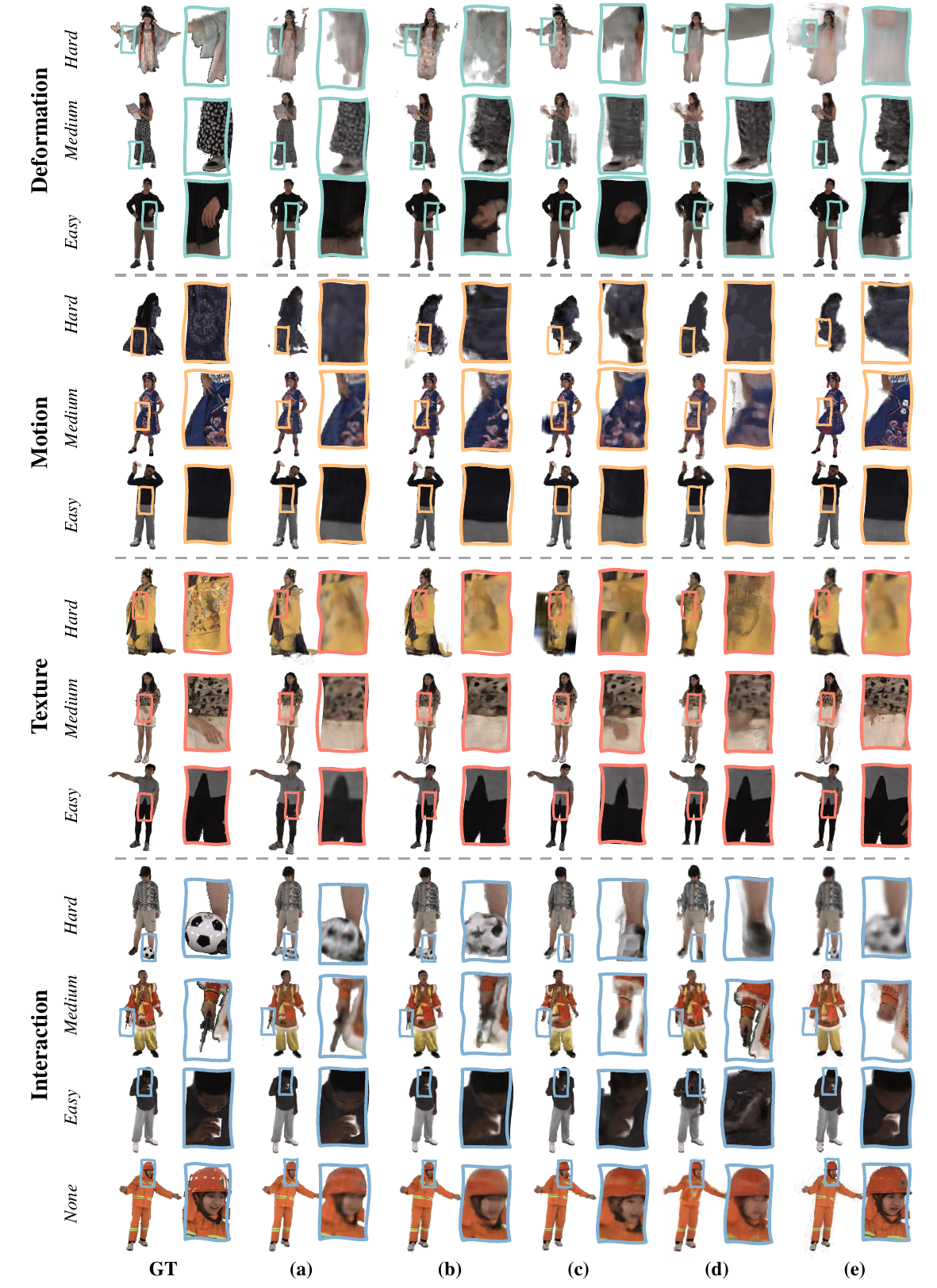}
\vspace{-2ex}
\caption{\small{\textbf{Novel pose animation results on each data split}. From top to bottom, we illustrate the reposing results generated by \textbf{(a-e)}: \nv~\cite{lombardi2019neural}, \anerf~\cite{su2021nerf}, \nb~\cite{peng2020neural} \aniN~\cite{peng2021animatable}, and \hn~\cite{weng2022humannerf}. Please zoom in for better visualization.} }\label{fig:novelpose_supp}
\end{figure*}

\noindent \textbf{Detailed Results.}
As a supplement to the result analysis in the main text, we present more detailed results and analysis in this subsection. Detailed quantitative results across our testing splits are listed in Tab.~\ref{tab:novelview}, which correspond to the bubble charts in Fig.~\ref{fig:novel_view}. We also illustrate the qualitative results in Fig.~\ref{fig:novelview_supp}. 
From the ranking in Tab.~\ref{tab:novelview}, we can observe that \anerf~\cite{su2021nerf}, \nb~\cite{peng2020neural} and \hn~\cite{weng2022humannerf} achieve the best numbers in most of the test splits in terms of PSNR, SSIM and LPIPS~\cite{zhang2018perceptual}. As the module designs of these methods might play vital roles in such distinct results, we further analyze the phenomenon by unfolding their conceptual differences as follows: \textit{\nv}~\cite{lombardi2019neural} adopts a VAE~\cite{kingma2013auto}-style neural rendering framework that encodes and decodes both the affinity transformation field and rendering volume from sparse view references. Such a paradigm absorbs the strength of VAE that compresses input multi-view features into one compact latent representation space, which follows the Gaussian assumption. Thus it could generalize well on novel views (achieving top-three performance in PSNR) with acceptable rendering quality. Whereas, such a framework also inherits the smooth problem of VAE that leads to not sharp enough qualitative results.
~\textit{\anerf}~\cite{su2021nerf} is a conditioned NeRF that utilizes local joint ordinate information of query points. It samples a small box center at a random point in the foreground and adds a proportion of background points to regularize the empty space, using only mean square error loss (MSE). This strategy enforces the network to encode the dynamic NeRF with local bone coordinates, which is efficient in the human foreground region. The overall novel view synthesis ability of \anerf~\cite{su2021nerf} is appealing, especially in PSNR. 
However, due to the sparsity characteristic of skeleton representation, \anerf~tends to generate dilated artifacts (see \textit{Motion-Medium} and \textit{Motion-Hard} in Fig.~\ref{fig:novelview_supp}), more obvious with novel pose in off-body parts of \textit{Interaction-Hard} case in Fig.~\ref{fig:novelpose_supp}.
\textit{\nb}~\cite{peng2020neural} and \textit{\aniN}~\cite{peng2021animatable} first compute a \threeD~bounding box from SMPL, then train/infer on the reprojected \threeD~box region and fill the outer region with the background color. Thus, their SSIM scores are typically greater than other methods that infer the whole image. Whereas, the bounding box only helps the network consider the main body and ignore the object, which leads to both methods can not reconstruct large interacting objects (as illustrated in the sword part of \textit{Interaction-Medium} case in Fig.~\ref{fig:novelview_supp}) \textit{\hn}~\cite{weng2022humannerf} combines the strength of previous methods' design, it samples squared boxes on reprojected \threeD~bounding box, and trained model with a perceptual~\cite{johnson2016perceptual} loss and MSE loss. This results in best LPIPS performance, as well as the best visual performance with sharp texture in qualitative results. However, as~\hn~\cite{weng2022humannerf} designs a human motion prior that is Gaussian distribution along body parts or bones. This prior may lead to training failure on loose clothing and interactive objects, as illustrated in the \textit{Deformation-Hard} and \textit{Interaction-Medium} cases in Fig.~\ref{fig:novelpose_supp}. 

\subsubsection{Novel Pose Animation}\label{sec:sup:benchmark-animation}
\noindent \textbf{Detailed Settings}
We conduct novel pose experiments on dynamic methods on the same models in novel view synthesis. Specifically, by training on the first $80\%$ frames of each case, we test the animatable model with input of the pose sequences extracted from the last $20\%$ frames with a 15-frame skip. Depending on the pose-condition scheme of different methods, the test input can be divided into two categories, the SMPL parameters and the image features. We use the same testing view and rendering resolution as the novel view synthesis experiment.

\noindent \textbf{Detailed Results.} We present the detailed novel pose animation results on $13$ testing splits in Tab.~\ref{tab:animation} in the main paper, and show the qualitative samples of results in Fig.~\ref{fig:novelpose_supp} in this subsection. Similar to novel view synthesis task, \anerf~\cite{su2021nerf} achieves the best PSNR performance, \nb~\cite{peng2020neural} has the best SSIM score, and \hn~\cite{weng2022humannerf} gets the best LPIPS~\cite{zhang2018perceptual}. 
Differently, \nv~\cite{lombardi2019neural}'s performance decrease by a large margin, especially in \textit{Motion} splits. One of the underlying reasons is that the affinity field learning in the \nv~\cite{lombardi2019neural} only relies on the latent code that is learned from multiview images and regularized by KL-divergence. Such a methodology is tied in a global warping manner, which might be relatively less affected by factors like global deformation in distance within a short movement change, but is vulnerable to unseen local motion (\textit{e.g.,} wrong head pose of \textit{Interaction-Medium} case in Fig.~\ref{fig:novelpose_supp}, and strained border of actor's shirt in \textit{Texture-Easy} case in Fig.~\ref{fig:novelpose_supp} ). Moreover, the design cannot preserve global scale in unseen poses, due to the wrong prediction of global affine transform sometimes (\textit{e.g., }the zoom scale of the actor in \textit{Texture-Easy} case in Fig.~\ref{fig:novelpose_supp})).
The methods~\cite{su2021nerf,peng2020neural,peng2021animatable,weng2022humannerf} with the explicit human pose information as input can typically generate reasonable animation results in terms of the local first motion, as shown in Fig.~\ref{fig:novelpose_supp}. Whereas, we draw the other major conclusion that current methods fail to model \textit{Deformation} and \textit{Interaction} properly. The typical examples shown in Fig.~\ref{fig:novelpose_supp} are the loose cloth case of \textit{Deformation-Hard}, and the interactive objects in \textit{Interaction-Medium} and \textit{Interaction-Hard} cases. How to properly model non-rig or out-of-body motions while preserving the advantages from explicit body representations is worth great pondering.

\begin{figure*}[ht]
\vspace{-3ex}
\centering
\includegraphics[width=0.90\textwidth]{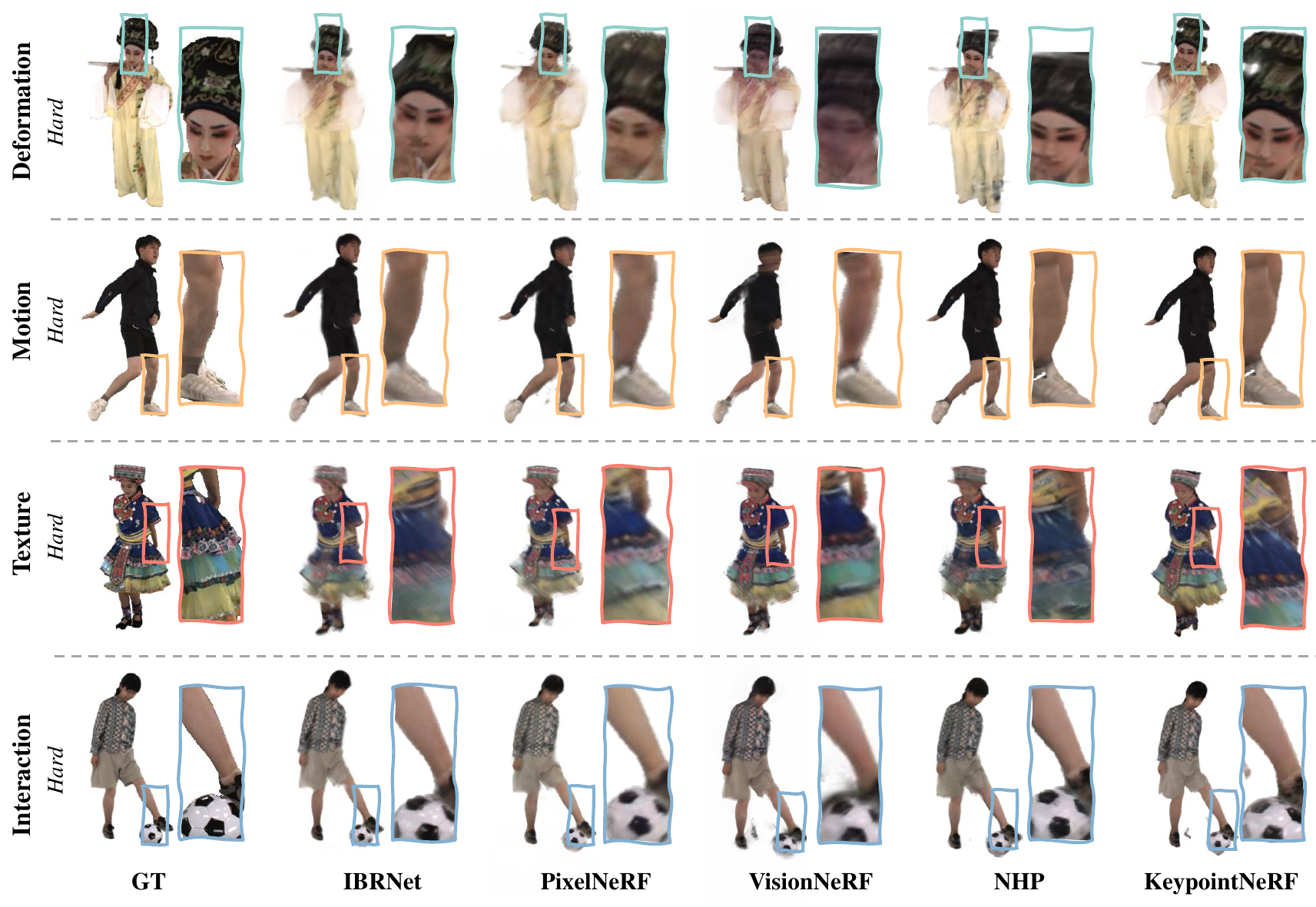}
\vspace{-2ex}
\caption{\small{\textbf{Novel ID synthesis results on each data split}.} Splits with hard difficulty are visualized to illustrate the robustness of generalizable methods on hard cases.}
\label{fig:figure-novel-id-exp-classes}
\vspace{-4ex}
\end{figure*}

\subsubsection{Novel Identity Rendering}\label{sec:sup:benchmark-generalize} 
\noindent \textbf{Detailed Settings.}
For novel identity synthesis, we review five selected state-of-the-art methods~\cite{yu2020pixelnerf,wang2021ibrnet,mihajlovic2022keypointnerf,1998Subdivision,kwon2021neural}, and described their modification in Sec.~\ref{sec:sup:methods-generalizable}.
The testing set is the same 39 testing sequences in the novel view and the novel pose benchmark. The training set consists of $400$ sequences with full coverage of all categories and difficulties. We select four balanced views as source views. These source view images are cropped and resized into $512\times512$ resolution (same with the official implementation in~\cite{mihajlovic2022keypointnerf,cheng2022generalizable,kwon2021neural}). For category-agnostic methods~\cite{yu2020pixelnerf,wang2021ibrnet,lin2023visionnerf}, we only provide segmentation and camera parameters during training and testing. For methods with human prior, we also input the fitted \smplx~or \threeD~keypoints. 
We train models on the full $60$ views in the training identity sequences. For inference, we evaluate the unseen identities on the same $18$ test views used in novel view and novel pose tasks, but on full sequences with frame skip at $45$.
All methods are trained under the same 8-V100 machine environment, and evaluated on a single V100. 

\noindent \textbf{Detailed Results.} 
In the main text, we draw the conclusion that generalization methods use human prior~\cite{mihajlovic2022keypointnerf,kwon2021neural} is more robust than category-agnostic methods~\cite{yu2020pixelnerf,wang2021ibrnet,lin2023visionnerf} according to the results reported in Tab.~\ref{tab:novel-id-sup}. To further illustrate this conclusion, we compare the qualitative results in \textit{Hard} level of each data factor in Fig.~\ref{fig:figure-novel-id-exp-classes}. Human prior methods generally render better images with more precise human shape and texture compared to category-agnostic methods, especially in the \textit{Texture-Hard} and \textit{Deformation-Hard} cases. Moreover, in addition to the influence of conceptual difference between image blending strategies to direct radiance prediction in the main text, we also compare the generalization of image feature extractors between \pixel~\cite{yu2020pixelnerf} and \vision~\cite{lin2023visionnerf}. \vision~\cite{lin2023visionnerf} uses a similar structure of \pixel~\cite{yu2020pixelnerf}, but mainly incorporates the local CNN-based image encoder with a global vision transformer. Such a design achieves better visual quality on average with sharper texture details and higher scores in both Fig.~\ref{fig:figure-novel-id-exp-classes} and Tab.~\ref{tab:novel-id-sup}, since the transformer is capable to learn more global coherence features across source views.


\begin{table*}[b]
\begin{center}
\vspace{-2ex}
\resizebox{\linewidth}{!}{
\begin{tabular}{c|l|ccccc|ccccc|ccccc} 
\hline
{\parbox[t]{3mm}{\multirow{16}{*}{\rotatebox[origin=c]{90}{\textbf{Novel View Synthesis} \enspace \enspace \enspace \enspace}}}} & \multirow{2}*{\textbf{Splits}} & \multicolumn{5}{c|}{\textbf{PSNR$\uparrow$}} & \multicolumn{5}{c|}{\textbf{SSIM$\uparrow$}} & \multicolumn{5}{c}{\textbf{LPIPS*$\downarrow$}} \\ 
& & \multicolumn{1}{c}{NV} & \multicolumn{1}{c}{AN} & \multicolumn{1}{c}{NB} & \multicolumn{1}{c}{AnN} & \multicolumn{1}{c|}{HN} & \multicolumn{1}{c}{NV} & \multicolumn{1}{c}{AN} & \multicolumn{1}{c}{NB} & \multicolumn{1}{c}{AnN} & \multicolumn{1}{c|}{HN} & \multicolumn{1}{c}{NV} & \multicolumn{1}{c}{AN} & \multicolumn{1}{c}{NB} & \multicolumn{1}{c}{AnN} & \multicolumn{1}{c}{HN} \\ \hline
& Motion Simple & 23.31 & 24.32 & \second{24.77} & \third{24.40} & \best{26.04} & 0.948 & 0.954 & \third{0.972} & \best{0.975} & \second{0.973} & 84.02 & 66.33 & \third{70.53} & \second{57.75} & \best{32.26}  \\  
& Motion Medium & 23.79 & 24.18 & \third{22.89} & \second{23.01} & \best{23.84} & 0.955 & 0.959 & \second{0.972} & \best{0.973} & \third{0.963} & 73.66 & 59.10 & \third{72.82} & \second{70.43} & \best{37.83}  \\  
& Motion Hard & 21.69 & 23.31 & \third{21.45} & \second{22.84} & \best{25.02} & 0.943 & 0.952 & \second{0.971} & \third{0.970} & \best{0.972} & 104.23 & 75.17 & \third{83.73} & \second{78.06} & \best{35.29}  \\ \cline{2-17} 
& Deformation Simple  & 24.56 & 24.85 & \third{24.45} & \second{25.27} & \best{27.59} & 0.951 & 0.956 & \third{0.964} & \second{0.964} & \best{0.971} & 61.39 & 46.81 & \third{72.32} & \second{56.19} & \best{30.46}  \\  
& Deformation Medium & 25.40 & 25.36 & \best{25.29} & \second{24.07} & \third{24.06} & 0.953 & 0.955 & \best{0.972} & \second{0.965} & \third{0.959} & 57.19 & 49.92 & \third{76.15} & \best{60.26} & \second{61.82}  \\  
& Deformation Hard & 22.89 & 23.08 & \second{21.34} & \third{20.18} & \best{22.80} & 0.931 & 0.935 & \best{0.964} & \second{0.948} & \third{0.937} & 96.88 & 84.66 & \third{110.75} & \best{98.65} & \second{104.05}  \\ \cline{2-17} 
& Texture Simple & 23.58 & 24.00 & \third{24.28} & \second{24.31} & \best{25.24} & 0.959 & 0.964 & \best{0.980} & \third{0.974} & \second{0.974} & 76.06 & 58.66 & \third{64.69} & \second{54.79} & \best{32.99}  \\  
& Texture Medium & 24.41 & 25.08 & \second{25.01} & \third{20.37} & \best{26.41} & 0.953 & 0.959 & \best{0.975} & \third{0.962} & \second{0.967} & 63.33 & 49.54 & \second{67.54} & \third{82.32} & \best{32.62}  \\  
& Texture Hard & 22.39 & 22.69 & \second{21.89} & \third{21.45} & \best{25.95} & 0.920 & 0.926 & \second{0.951} & \third{0.950} & \best{0.958} & 110.36 & 115.52 & \third{103.13} & \second{86.19} & \best{26.35}  \\ \cline{2-17} 
& Interaction No & 24.83 & 25.23 & \second{24.71} & \third{20.70} & \best{25.28} & 0.961 & 0.964 & \best{0.980} & \third{0.964} & \second{0.966} & 68.06 & 49.58 & \second{67.65} & \third{79.91} & \best{45.11}  \\  
& Interaction Simple & 24.67 & 25.38 & \third{25.39} & \second{25.47} & \best{25.95} & 0.963 & 0.968 & \best{0.982} & \second{0.977} & \third{0.971} & 61.90 & 43.24 & \third{59.19} & \second{57.79} & \best{34.74}  \\  
& Interaction Medium & 22.11 & 23.23 & \third{21.67} & \second{21.84} & \best{22.84} & 0.938 & 0.943 & \best{0.965} & \second{0.961} & \third{0.948} & 99.50 & 80.25 & \third{94.61} & \second{91.78} & \best{69.95}  \\  
& Interaction Hard & 23.55 & 24.06 & \second{21.93} & \third{21.15} & \best{22.18} & 0.939 & 0.944 & \best{0.961} & \second{0.959} & \third{0.941} & 84.15 & 70.00 & \third{96.80} & \second{94.30} & \best{87.81}  \\ \cline{2-17} 
& Overall & 23.63 & 24.21 & \second{23.47} & \third{22.70} & \best{24.86} & 0.947 & 0.952 & \best{0.970} & \second{0.965} & \third{0.962} & 80.06 & 65.29 & \third{79.99} & \second{74.49} & \best{48.56} \\ \hline 

{\parbox[t]{2mm}{\multirow{16}{*}{\rotatebox[origin=c]{90}{\enspace\enspace\enspace \enspace\enspace\textbf{Novel Pose Animation}}}}}
& Motion Simple & 21.17 & \third{24.05} & \second{24.31} & 21.34 & \best{25.62} & 0.941 & 0.952 & \second{0.972} & \third{0.953} & \best{0.972} & 93.60 & \second{66.80} & \third{76.33} & 82.37 & \best{32.45}  \\  
& Motion Medium & 19.40 & \best{21.64} & \second{21.50} & 20.53 & \third{21.04} & 0.941 & \third{0.947} & \best{0.968} & 0.936 & \second{0.951} & 100.36 & \third{78.29} & 83.66 & \second{61.69} & \best{54.61}  \\  
& Motion Hard & 19.09 & \second{21.39} & \third{20.64} & 18.85 & \best{23.58} & 0.938 & 0.948 & \second{0.967} & \third{0.950} & \best{0.967} & 112.72 & \second{80.41} & \third{87.99} & 95.97 & \best{40.50}  \\ \cline{2-17} 
& Deformation Simple  & 20.73 & \second{23.89} & \third{23.43} & 23.00 & \best{26.17} & 0.938 & 0.950 & \second{0.962} & \third{0.960} & \best{0.966} & 87.29 & \second{55.29} & 81.29 & \third{67.70} & \best{35.39}  \\  
& Deformation Medium & 22.43 & \second{24.89} & \best{25.19} & \third{23.12} & 22.76 & 0.943 & 0.951 & \best{0.971} & \third{0.952} & \second{0.955} & 70.79 & \best{52.05} & 75.68 & \second{61.72} & \third{70.44}  \\  
& Deformation Hard & 19.13 & \third{20.83} & \second{21.34} & 18.52 & \best{21.41} & 0.920 & 0.923 & \best{0.964} & \second{0.941} & \third{0.929} & 131.86 & \second{108.27} & \third{110.74} & \best{91.70} & 129.75  \\ \cline{2-17} 
& Texture Simple & 20.44 & \third{23.42} & \best{24.27} & 23.07 & \second{24.20} & 0.950 & 0.962 & \best{0.980} & \third{0.972} & \second{0.972} & 87.66 & \second{59.22} & \third{65.08} & 77.59 & \best{36.64}  \\  
& Texture Medium & 23.16 & \third{24.56} & \second{25.14} & 21.23 & \best{26.63} & 0.950 & 0.957 & \best{0.977} & \third{0.957} & \second{0.967} & \third{69.64} & \second{51.30} & 72.68 & 69.96 & \best{30.81}  \\  
& Texture Hard & 20.23 & \second{21.90} & \third{21.65} & 17.93 & \best{25.66} & 0.911 & 0.921 & \second{0.951} & \third{0.932} & \best{0.956} & 136.26 & 124.49 & \second{111.66} & \third{116.84} & \best{24.88}  \\ \cline{2-17} 
& Interaction No & 21.45 & \best{24.90} & \second{24.37} & 22.08 & \third{23.52} & 0.953 & \third{0.962} & \best{0.979} & 0.948 & \second{0.962} & 88.23 & \second{53.08} & \third{69.11} & 70.76 & \best{50.70}  \\  
& Interaction Simple & 22.18 & \third{25.05} & \second{25.32} & 22.77 & \best{25.99} & 0.958 & 0.967 & \best{0.982} & \third{0.970} & \second{0.971} & 71.61 & \second{43.02} & \third{59.29} & 72.53 & \best{32.95}  \\  
& Interaction Medium & 19.83 & \best{22.83} & \third{21.33} & 20.60 & \second{22.40} & 0.931 & 0.943 & \best{0.966} & \second{0.959} & \third{0.947} & 107.04 & \second{76.08} & 94.86 & \third{92.98} & \best{66.90}  \\  
& Interaction Hard & 20.55 & \best{23.06} & \second{21.93} & 21.02 & \third{21.42} & 0.927 & \third{0.937} & \best{0.961} & \second{0.945} & 0.934 & 110.91 & \best{81.78} & 96.79 & \second{92.63} & \third{95.35}  \\ \cline{2-17} 
& Overall & 20.75 & \second{23.26} & \third{23.11} & 21.08 & \best{23.88} & 0.939 & 0.948 & \best{0.969} & \third{0.952} & \second{0.958} & 97.54 & \second{71.54} & 83.47 & \third{81.11} & \best{53.95}\\ \hline

\end{tabular}
}
\vspace{-2ex}
\caption{\textbf{Benchmarks with sparse view training.} We abbreviate \nv~\cite{lombardi2019neural} as `NV', \anerf~\cite{su2021nerf} as `AN', \nb~\cite{peng2020neural} as `NB', \aniN~\cite{peng2021animatable} as `AnN' and \hn~\cite{weng2022humannerf} as `HN'. Cell color \legendsquare{colorbest}~\legendsquare{colorsecond}~\legendsquare{colorthird} indicate the best, second best, and third best performance in the same split respectively.}~\label{tab:benchmark-sparse}
\vspace{-7ex}
\end{center}
\end{table*}

\subsubsection{Benchmarks with Sparse View Training}
Compare with dense view human synthesis, rendering humans from sparse views or even with a monocular image setting, take a step further in narrowing the domain gap between structured in-door data capture and unstructured open-world data collection. Relaxing the requirement on structured data could
help towards portable human avatar generation and may further enable applications like bullet-time rendering from online videos. Thus, aside from the main benchmarks, we also evaluate the state-of-the-art methods' performances under a spare-view training setting.

\noindent \textbf{Setting.} Similar to the dense novel view and novel pose experiments, we retrain separate models for each dynamic human rendering method. The key difference of the sparse view settings from the dense ones is that, each model is trained with only four balanced views, namely Camera $1$, $13$, $25$, and $37$. The other setting details are the same as the dense novel view and novel pose benchmarks.


\begin{figure*}[htbp]
   \centering
   \begin{subfigure}[b]{0.45\textwidth}
     \centering
     \includegraphics[width=\textwidth]{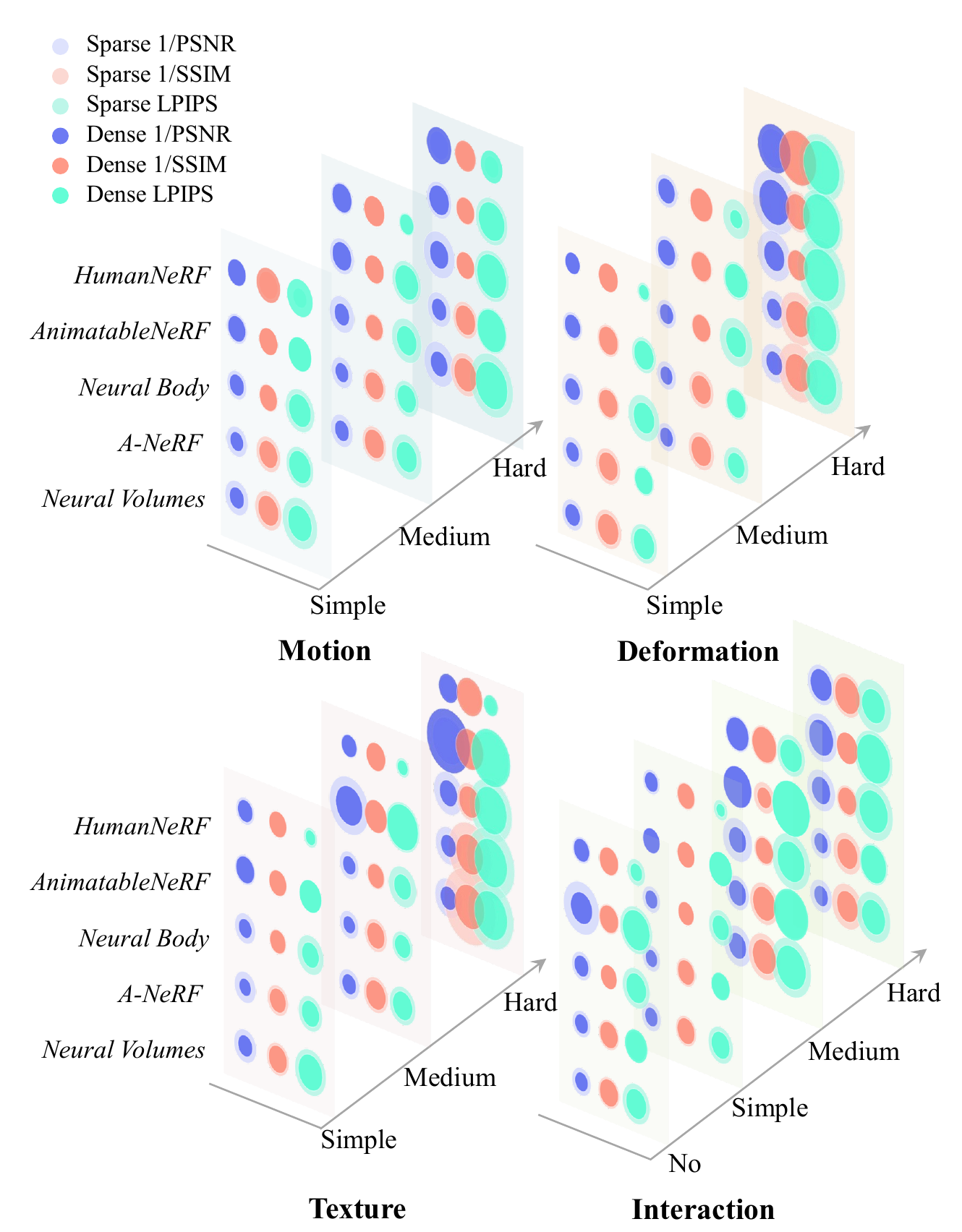}
     \caption{Novel view synthesis with different training view numbers.}
     \label{fig:sparse-novel-view}
   \end{subfigure}
   \hfill
   \begin{subfigure}[b]{0.45\textwidth}
     \centering
     \includegraphics[width=\textwidth]{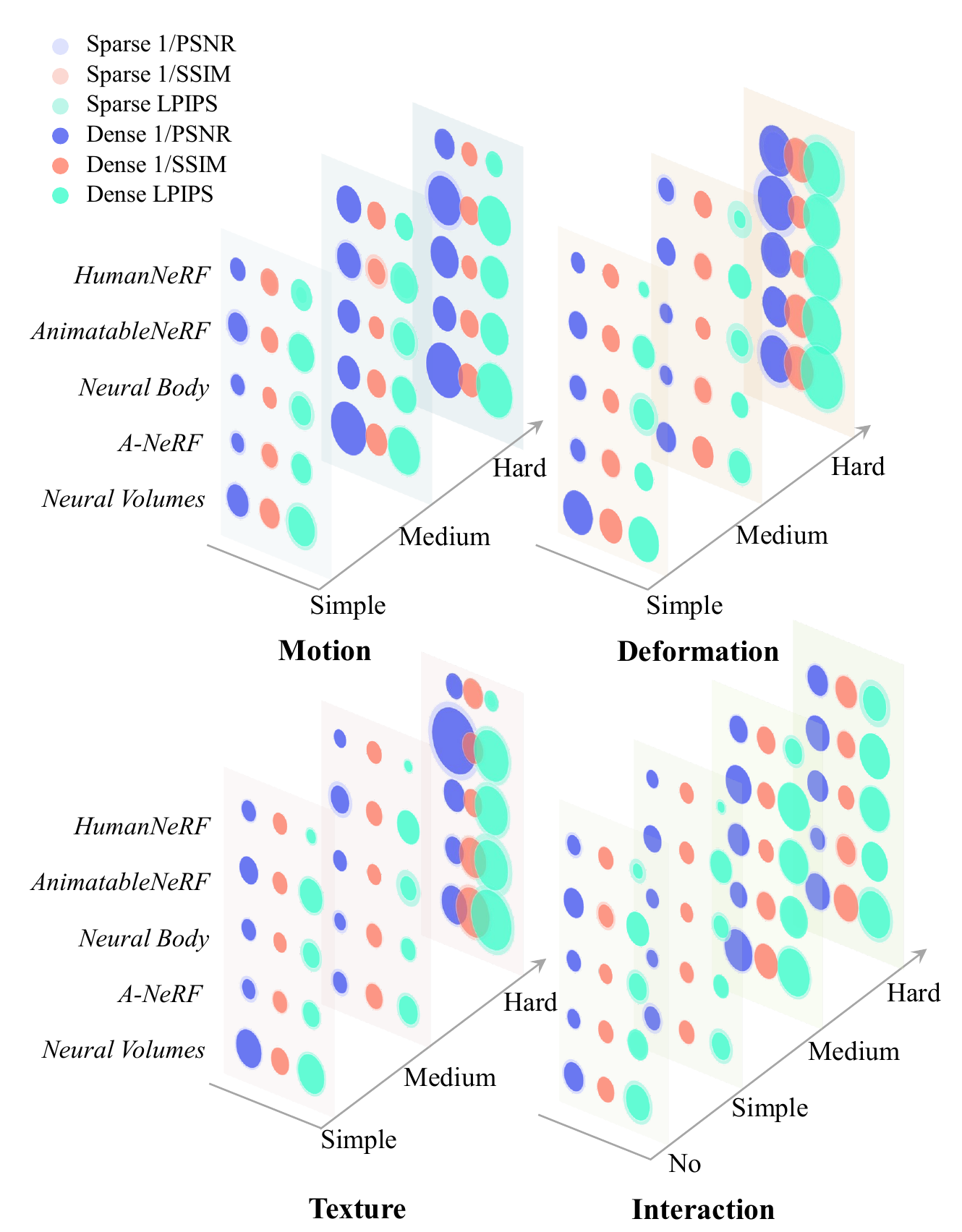}
     \caption{Novel pose animation with different training view numbers.}
     \label{fig:sparse-novel-pose}
   \end{subfigure}
   \vspace{-0.25cm}
    \caption{\textbf{Visualization of quantitative comparison between dense view training and sparse view training.} For (a) novel view and (b) novel pose tasks, we compare the dense view training (42 views) and sparse view (4 views) training on different data splits. Bubbles in dark tones are with dense view training, and other ones in light tones are with sparse view training. Noted that the scale used here is different from Fig.\ref{fig:novel_view} for better comparison visualization.}
    \label{fig:sparse-benchmark}
    \vspace{-3ex}
\end{figure*}


\noindent \textbf{Results.} We list the quantitative results of both sparse novel view synthesis and sparse novel pose animation in Tab~\ref{tab:benchmark-sparse}, and compare the difference between dense and sparse view training in Fig~\ref{fig:sparse-benchmark}. For the sparse novel view synthesis task, we can observe that \hn~\cite{weng2022humannerf} achieves the best PSNR and LPIPS metric in most of the splits, and \nb~\cite{peng2020neural} gets the best SSIM performance. Originally sparse-designed methods, \nb~\cite{peng2020neural},~\aniN~\cite{peng2021animatable}, and \hn~\cite{weng2022humannerf} ranked top-3 in all evaluation metrics, this phenomenon is totally different from dense results in Tab.~\ref{tab:novelview}. The performance gap between sparse view settings and the dense ones can be easily observed from the inflating bubbles in Fig.~\ref{fig:sparse-novel-view}. In the figure, the bubbles with darker colors refer to the performances under dense view settings, and the ones with lighter colors refer to the results under sparse view settings.  The underlying reason for the phenomenon lies in the natural difficulty in sparse neural field supervision -- the fewer training observations require a network to have the more powerful capability on learning multi-view relationships (both interpolation and extrapolation) and proper hallucinating, to approximate precise geometry. \nb~\cite{peng2020neural},~\aniN~\cite{peng2021animatable}, and \hn~\cite{weng2022humannerf} all adopt strong human priors with SMPL mesh, blend weights, and motion priors. Thus, they are more robust to sparse observations. In contrast, \anerf~\cite{su2021nerf} integrates only skeleton prior that is sparse in human shape representation, and the category-agnostic method \nv~\cite{lombardi2019neural} relies on dense observations to overfit to a particular distribution. These two methods' performances drop significantly when given fewer views during training. Similar to the observation from dense novel view benchmarks, the current dynamic human method performs unsatisfactorily when \textit{Deformation} and \textit{Texture} difficulty increase. Due to the complex texture and off-body non-rigidity, the results in the sparse setting further enlarge the gap with the dense one. In contrast with the phenomena in sparse novel view synthesis, the quantitative results on the sparse novel pose animation task  (Tab.~\ref{tab:benchmark-sparse}) show similar trends compared to the dense view setting. Specifically, \hn~\cite{weng2022humannerf} and \nb~\cite{peng2020neural} perform the best among the three metrics. \anerf~\cite{su2021nerf} shows better pose generalization ability compared to \aniN~\cite{peng2021animatable}, considering the fact that \aniN~\cite{peng2020neural} performs better in novel view synthesis tasks given seen human poses. When comparing the differences between dense and sparse settings, quantitative results display a relatively smaller performance drop. The phenomenon reveals that, the training observations from both dense and sparse view settings are not adequate enough for the benchmark methods to learn a compact dynamic field for unseen poses.

\begin{figure*}[t]
\vspace{-4ex}
\centering
\includegraphics[width=0.98\textwidth]{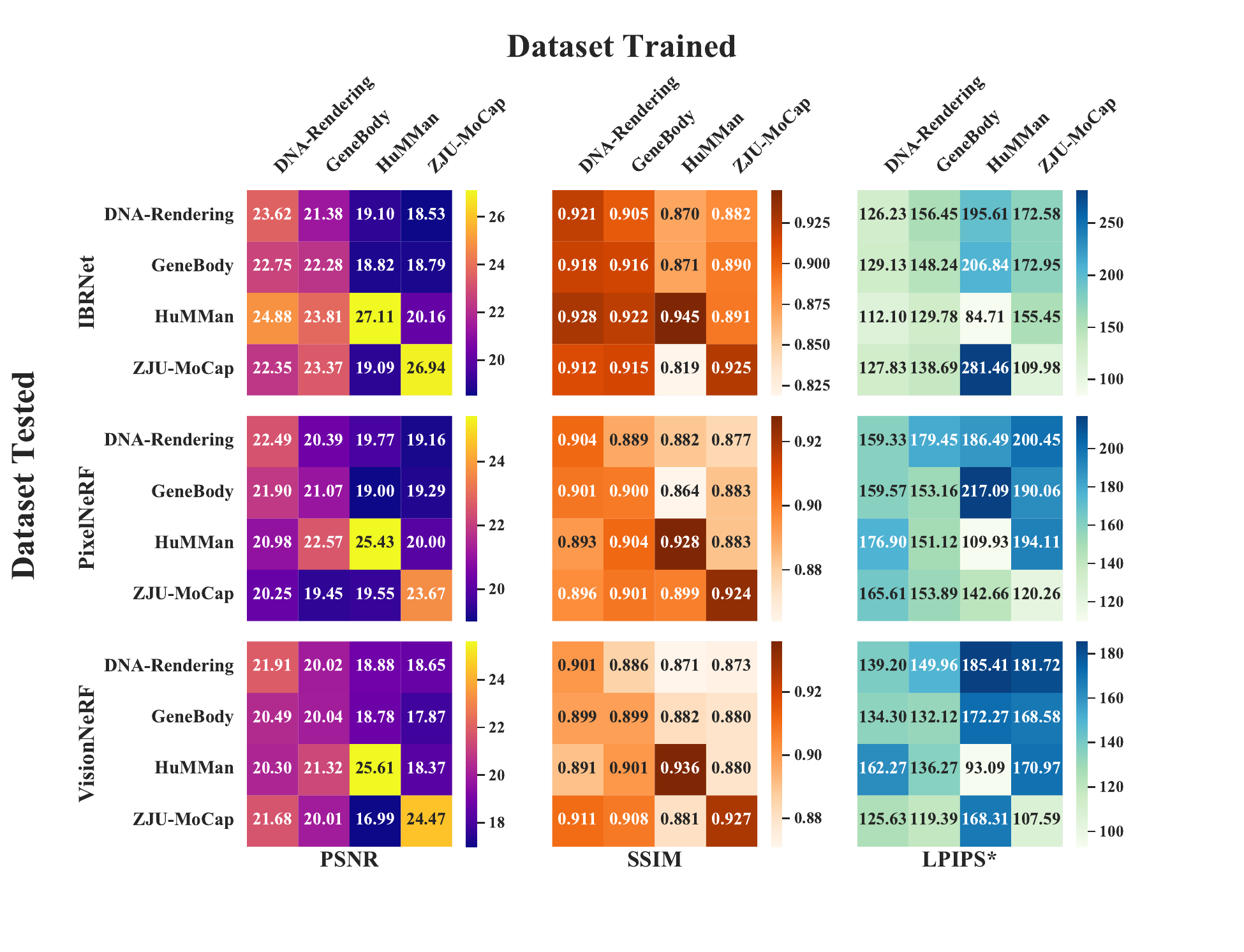}
\vspace{-6ex}
\caption{\small{\textbf{Cross-dataset affinity map for category-agnostic generalization methods}. We crossly evaluate models trained on each dataset and plot their performance on testing splits on each dataset. The PSNR, SSIM, and LPIPS* are plotted in separate matrices.}}
\vspace{-3ex}
\label{fig:cross-affinity-full}
\end{figure*}

\section{Cross-dataset Details}\label{sec:sup:cross}
In this section, we provide more details as a supplement to the cross-dataset evaluation mentioned in the main paper. Specifically, we first describe the criteria for selecting the compared datasets, and review the key attributes of these datasets in Sec.~\ref{sec:sup:cross-dataset}. Then, we introduce the implementation and setting details in Sec.~\ref{sec:sup:cross-setting}. We discuss the results in Sec.~\ref{sec:sup:cross-results}. Finally, we analyze the impact of color consistency on multi-camera datasets in Sec.~\ref{sec:sup:cross-dataset-view}.

\subsection{Compared Datasets}\label{sec:sup:cross-dataset}

To evaluate the potential of the proposed dataset on boosting algorithms' generalizability from the data engineering aspect, we compare the proposed dataset with the most commonly used human-centric multiview datasets on the generalizable neural rendering task. For a fair comparison, we select datasets with foreground segmentation annotations and dense camera views. Thus, several well-known datasets are not suitable for this evaluation. For example. Human3.6M~\cite{ionescu2013human3} only contains four RGB cameras, CMU Panoptic~\cite{joo2015panoptic} and AIST++~\cite{aist-dance-db,li2021ai} lack official segmentation annotation{\footnote{Some human rendering methods~\cite{kwon2021neural,chen2023uv} use their own tools to generate mask, we exclude these mask sources for fairness.}}. We select \zjumocap~\cite{peng2020neural}, \humman~\cite{cai2022humman} and \genebody~\cite{cheng2022generalizable} for comparison. In this subsection, we discuss their main features and their adaption to generalizable methods.

\noindent\textbf{\zjumocap}~\cite{peng2020neural} is currently the most widely used dataset in human neural rendering domain. It contains 10 multiview performance sequences, with accurate camera calibration, human segmentation as well as SMPL annotation. The main drawback of this dataset is the lack of diversity in clothing and motion and without human objection interaction. As mentioned in Sec.~\ref{sec:annotation} in the main paper, color consistency design might be ignored in \zjumocap~\cite{peng2020neural}, where obvious color differences can be observed between neighboring cameras, as shown in Fig.~\ref{fig:cross-view-stats-example}. When training generalizable models on this dataset, we adopt the official splits and follow the implementation in \kptnerf~\cite{mihajlovic2022keypointnerf}.

\noindent\textbf{\humman}~\cite{cai2022humman} is a human action dataset with data captured under $10$ synchronized Kinect RGB-D cameras. It contains 400k sequences and 500 human actions which emphasize muscle-related movements. The clothing diversity is marginal where most subjects wear sports and daily costumes, and there is no human-object interaction either. As the source images come from the Kinect sensor, they might be stuck in low-quality, and obvious color differences can be found in \humman~\cite{cai2022humman} dataset. Note that the full dataset is still unreachable, we train models on the released version,  with its official list that contains a training split with 317 sequences and a testing split with 22 sequences. Noted that different from other datasets, where cameras are organized in a world coordinate near the origin that axis alignment with the real world, \humman~\cite{cai2022humman} uses a coordinate system relative to the first camera. Thus, we make a rigid transformation to eliminate the coordinate system difference.

\noindent\textbf{\genebody}~\cite{cheng2022generalizable} is a recent multi-view human performance capture dataset, which contains relatively wide diversity coverage among clothing, motion, and interactions. It captures human performance with 48 synchronized 5MP cameras with a low proportion-in-view of the main subject, where the average height of the human bounding box is around 600 pixels. We use its official splits with 40 training sequences and 10 testing sequences.

\begin{figure}[hb]
\vspace{-2ex}
\centering
\includegraphics[width=0.9\linewidth]{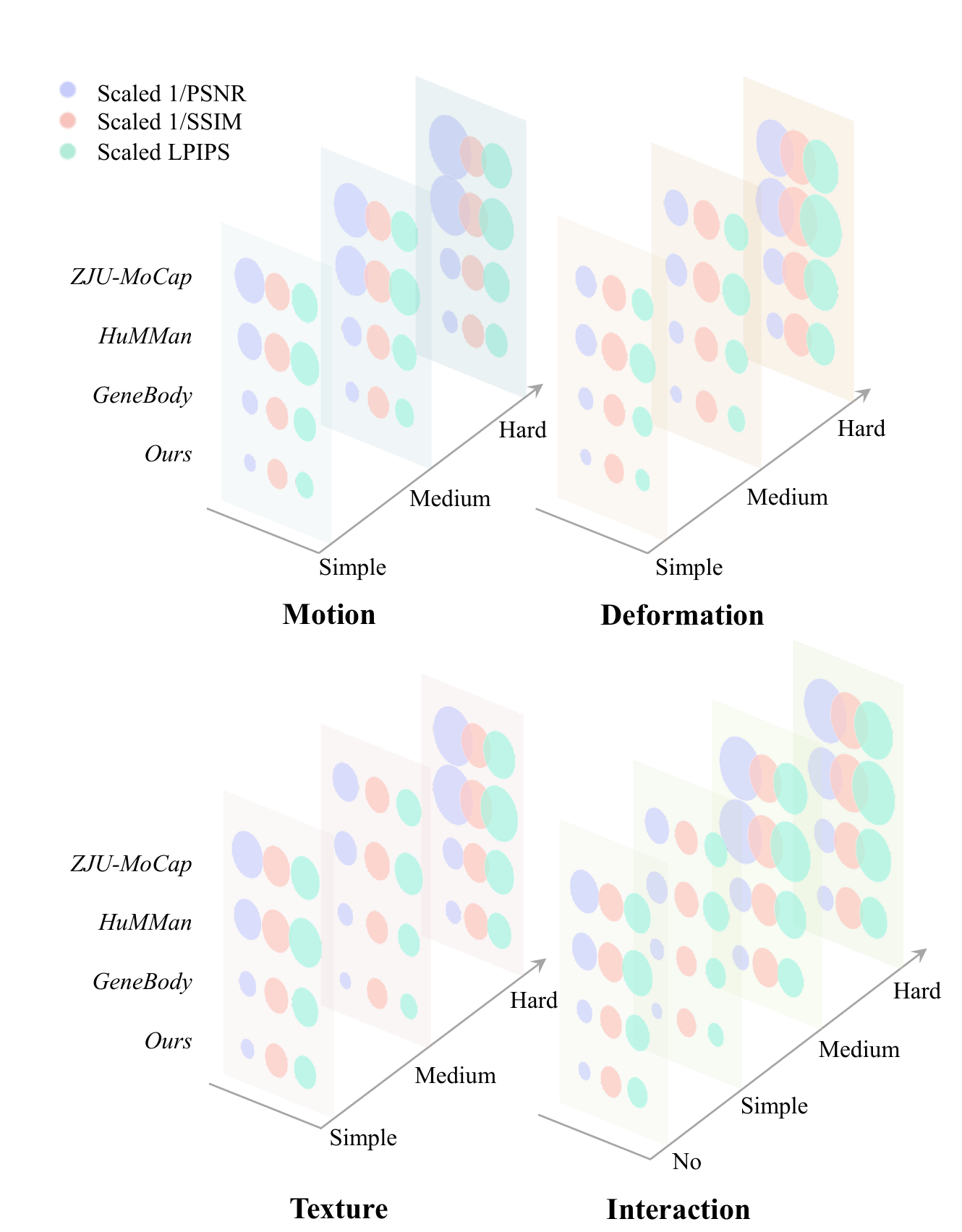}
\caption{\small{\textbf{Cross-dataset evaluation on our DNA-Rendering data splits}. We visualize the performance of models trained on different datasets on the proposed dataset's splits.} }\label{fig:cross_split}
\vspace{-2ex}
\end{figure}

\subsection{Detailed Settings}\label{sec:sup:cross-setting}
Multiple factors might affect the impact assessment across different datasets on neural rendering tasks, {\textit{eg.,}} the proportion of subjects in camera views, data scale, annotation accuracy, source view selection, training status, {\textit{etc.}} Our main goal is to investigate where the \textit{diversity} of the proposed dataset can benefit the generalization of human rendering. We conduct the experiments in the following settings. $1)$ In order to ensure the fairness of dataset comparison, we need to unify several input conditions, {\textit{e.g.,} number and resolution of source views, \textit{etc.} Meanwhile, we only investigate the category-agnostic generalizable methods, namely \pixel~\cite{yu2020pixelnerf}, \ibr~\cite{wang2021ibrnet} and \vision~\cite{lin2023visionnerf}, to avoid difference of input and accuracy of human prior in multiple datasets. 
$2)$ Evaluating the whole object-centric images with background removal produces a large rendering metric gap if the center subjects' proportions in camera views differ a lot. Thus, different from the novel pose benchmark in Sec.~\ref{sec:benchmark} and Sec.~\ref{sec:sup:benchmark-generalize} where a half resolution is used, we crop the subjects out from the original image in all datasets with square bounding boxes and resize them to $512\times512$ resolution for both source views and target views{\footnote{Note that this setting leads to the absolute values in cross-dataset evaluation worse than the ones in novel identity task, due to the larger proportion of human foreground.}}. 
$3)$ As the discussed datasets are all captured in dense circling camera arrays, we manually select four balanced views as the reference views at the same height that have a roughly 90-degree interval. 
$4)$ During training, we use the same learning rate over all datasets and stop the training process with the same global step. Each model is trained on one 8-V100 machine with distributed data-parallel stopping at $200k$ iterations. 
$5)$ Finally, we train and evaluate all the models based on the official splits of each dataset. For the comparable data volume 
magnitude of test samples, we size the data volume of test frames or test views on each dataset with $\langle$~$45$ frame-skip, 18 test views~$\rangle$ on \genebody~and \datasetname~, $\langle$~$45$ frame-skip, 12 uniform sampled test views~$\rangle$ on~\zjumocap~, and $\langle$~$8$ frame-skip, 6 test views~$\rangle$ on \humman, respectively.

\subsection{Additional Results}\label{sec:sup:cross-results}
In the main paper, we present the average PSNR performance over all three general scene methods, \textit{i.e.,} \pixel~\cite{yu2020pixelnerf}, \ibr~\cite{wang2021ibrnet}, and \vision~\cite{lin2023visionnerf}. Here, we present the performances of these methods individually (Fig.~\ref{fig:cross-affinity-full}), and illustrate the qualitative results (Fig.~\ref{fig:cross-ibr-400k}). We unfold the result analysis in terms of in-domain, and cross-domain in Sec.~\ref{sec:sup:cross-indomain} and Sec.~\ref{sec:sup:cross-crossdomain} respectively. A discussion on the impact of color consistency is discussed in Sec.~\ref{sec:sup:cross-dataset-view}.

\begin{figure*}[t]
\vspace{-0.5cm}
\centering
\includegraphics[width=0.98\textwidth]{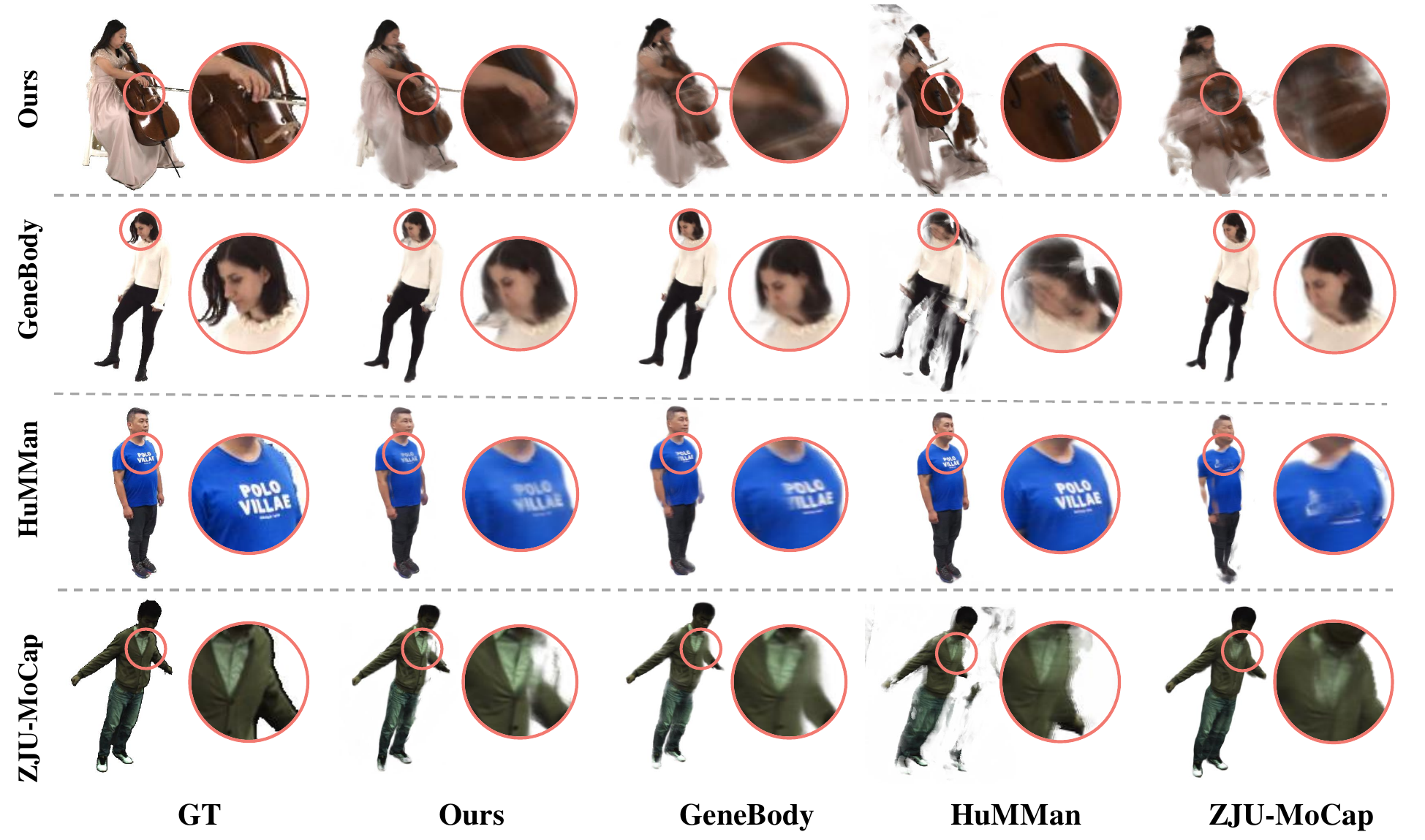}
\vspace{-2ex}
\caption{{\textbf{Qualitative cross-dataset results}. We demonstrate samples from different datasets (left labels) generated by IBRNet~\cite{wang2021ibrnet} models trained on different datasets (bottom labels).}}
\vspace{-3ex}
\label{fig:cross-ibr-400k}
\end{figure*}

\subsubsection{In-domain}\label{sec:sup:cross-indomain}

In-domain refers to the problem of evaluating models with the trainset and testset sharing the same underlying data distribution. We observe two key phenomena: 
$1)$ models trained on datasets with low data diversity achieve better in-domain results. As shown in the diagonal elements of the matrices in Fig.~\ref{fig:cross-affinity-full}, for in-domain generalization performance, methods trained on~\humman~\cite{cai2022humman} and \zjumocap~\cite{peng2020neural} achieve the best and second performances with relatively appealing metric values. In contrast, their in-domain performances on~\genebody~and~\datasetname~ are worse than the other two datasets. 
Noting that test sets in \zjumocap~and \humman~only contain cases with textureless clothing and easy motion illustrated in Fig.~\ref{fig:cross-ibr-400k}, which are easy cases in terms of data difficulty.
Besides, the nature of textureless data and easy human geometry is relatively friendly to category-agnostic generalization methods that conduct multiview image feature aggregation in a common manner without geometry prior.
$2)$ Larger data volume and diversity boosts in-domain performance. Concretely, like the proposed dataset, \genebody~\cite{cheng2022generalizable} contains a train and test split with a wide distribution of clothing, accessories, and motion, while with far less data volume and diversity compared to \datasetname (about $10\%$ data volume of our dataset). 
Despite both test sets of \genebody~and proposed dataset containing cases with even distribution in multiple difficulties, all three methods demonstrate our boost on in-domain performance.

\subsubsection{Cross-domain}\label{sec:sup:cross-crossdomain}
Cross-domain refers to directly evaluating the pre-trained model on one dataset to the test split of another dataset, which is represented by the off-diagonal elements in Fig.~\ref{fig:cross-affinity-full}. We expand the result analysis in two folds: $1)$ datasets with large variations of data attributes and difficulties boost the cross-domain generalization. Higher performance degradation can be observed in off-diagonal elements in each row in \zjumocap~\cite{peng2020neural} and \humman~\cite{cai2022humman} in Fig.~\ref{fig:cross-affinity-full}. Besides, even when \zjumocap~\cite{peng2020neural} trained model test on cross-domain case with easy clothing and motion, the cross-domain performance is still far from acceptable, refer to left most blue T-shirt case in Fig.~\ref{fig:cross-ibr-400k}. On the contrary, \genebody~\cite{cheng2022generalizable} and \datasetname~ experience a very marginal degradation when evaluating other test sets.
$2)$ Larger data volume and diversity can also boost cross-domain robustness. As mentioned in the in-domain part, \datasetname~is $10$ times than \genebody~in terms of data volumes, such improvement helps increase the model's generalization ability in considering margin. In the overwhelming majority of first-row elements, models with the proposed dataset get the best cross-domain results and even perform better than in-domain results of \genebody.

\noindent \textbf{Cross-domain on \datasetname~Splits.} To further investigate the performance of different dataset-trained models' performance on different human performance dimensions and difficulties, we visualize their performance on \datasetname~splits in Fig.~\ref{fig:cross_split}. We plot the in-domain result of our dataset-trained model as a reference bar. \genebody~\cite{cheng2022generalizable} has a relatively wide range of dimensions across our dimension and it achieves the best rendering quality among all splits. On the other hand \humman~\cite{cai2022humman} and \zjumocap~\cite{peng2020neural}, only contain easy clothing, motion, and interaction, the performance on more difficult splits drop significantly espec compared to \genebody~\cite{cheng2022generalizable}.


\begin{figure*}[th]
\vspace{-2ex}
   \centering
   \begin{subfigure}[b]{0.23\textwidth}
     \centering
     \includegraphics[width=\textwidth]{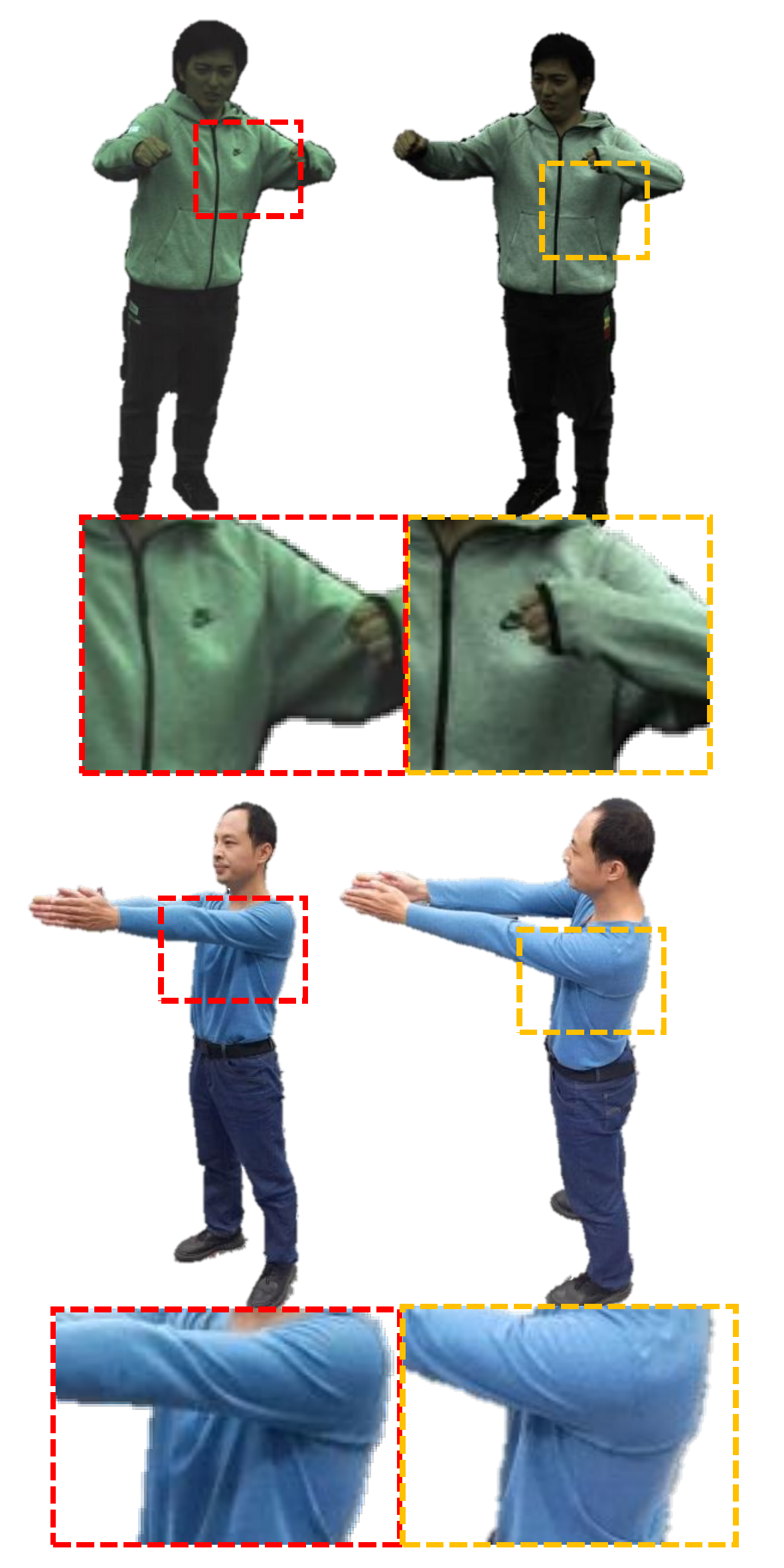}
     \caption{}
     \label{fig:cross-view-stats-example}
   \end{subfigure}
   \hfill
   \begin{subfigure}[b]{0.25\textwidth}
     \centering
     \includegraphics[width=\textwidth]{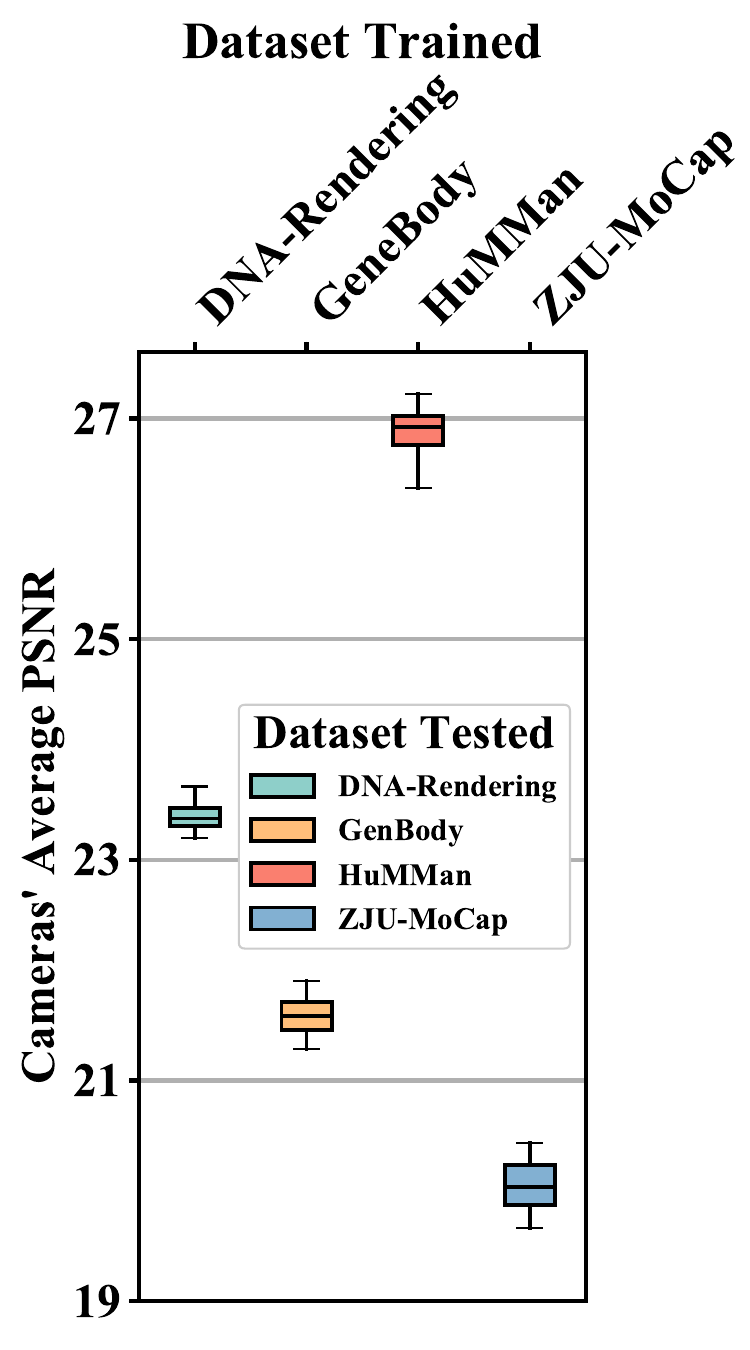}
     \caption{}
     \label{fig:cross-view-stats-indomain}
   \end{subfigure}
   \hfill
   \begin{subfigure}[b]{0.50\textwidth}
     \centering
     \includegraphics[width=\textwidth]{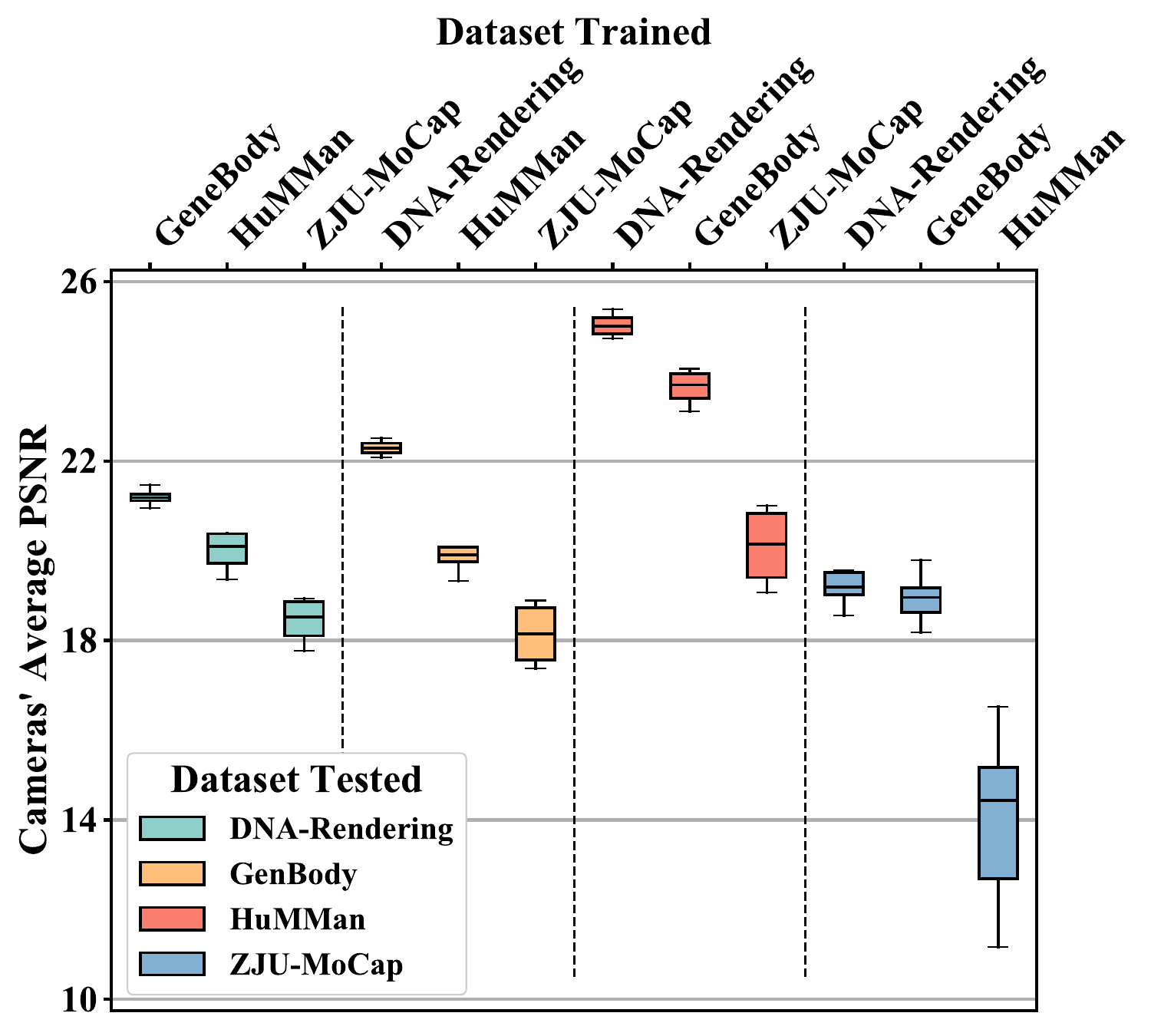}
     \caption{}
     \label{fig:cross-view-stats-crossdomain}
   \end{subfigure}
   \vspace{-1ex}
    \caption{\textbf{Cross-dataset evaluation across views.} (a) Examples of the color differences between neighboring cameras in \zjumocap~\cite{peng2020neural} (top) and \humman~\cite{cai2022humman} (bottom). (b) \textbf{In-domain} statistics on view average PSNR when trained and tested on the same dataset. (c) \textbf{Cross-domain} statistics on view average PSNR when trained and tested on different datasets. In both (b) and (c), the horizontal axis means different datasets trained, and the color of the box separates the datasets tested. The line in the middle indicates the median value, the box indicates the lower to the higher quartile, and the whiskers indicate the range of average PSNR across views.}
    \label{fig:cross-view-stats}
    \vspace{-3ex}
\end{figure*}


\subsection{Impact of Color Consistency} \label{sec:sup:cross-dataset-view}
To further analyze the impact of color consistency of training dataset in generalizable rendering, we unfold the stats across views on the cross-dataset results. Due to the different groundtruth in different views, it is hard to draw any conclusion from any single frame. Thus, we expand the average PSNR across camera views and analyze the statistics. Noted that we only select the test views which have very close angle distances from the nearest source view, to erase the performance gap from the viewpoints.
The average PSNR across testing cameras is plotted in Fig.~\ref{fig:cross-view-stats}. More concretely, we visualize the in-domain statistics of cameras' average PSNR in Fig.~\ref{fig:cross-view-stats-indomain}. When training and testing a model on the same dataset, the camera color distinction will remain constant. Models trained on the datasets that cannot ensure color consistency across views (illustrated in Fig.~\ref{fig:cross-view-stats-example}) might treat the color difference of different views as the view-depend effect and memorize it. The variance of cameras' average performance in such datasets is slightly higher than \genebody~\cite{cheng2022generalizable} and the proposed dataset.
Cross-domain generalization span on views is also plotted in Fig.~\ref{fig:cross-view-stats-crossdomain}. Different from in-domain statistics, when generalizing on other datasets, models trained on datasets with color inconsistency all suffer a major average performance dropping, and the variance of camera performance becomes even larger. This phenomenon is very noticeable on cross-evaluation between \zjumocap~\cite{peng2020neural} and \humman~\cite{cai2022humman}. While models trained on the proposed dataset as well as \genebody~\cite{cheng2022generalizable}, have very small view performance variations between each other. The increased view variation on \zjumocap~\cite{peng2020neural} and \humman~\cite{cai2022humman} is due to the nature color difference on their groundtruth. In a nutshell, with the best color consistency, the proposed dataset can benefit the community by providing high-quality data with faithful probing capability across views.

\section{Future Work}\label{sec:sup:future}

\noindent \textbf{Leaderboard.}
In the current human-centric rendering community, researchers from different institutions use different datasets and experimental settings to evaluate the performances of their algorithms. There is no agreement across institutions to benchmark human rendering methods under the same criterion yet. To reduce such divergence and align the standards, we proposed a large-scale diverse dataset \datasetname, and construct a complete benchmark in three human-centric rendering tasks. Benchmarks are evaluated on different data splits on different factors and difficulties. Additionally, we conduct a cross-dataset evaluation which demonstrates the proposed dataset can benefit the community from its diversity and coverage. In \datasetname, all actors are signed with agreements before data collection. Thus, all of the data is with a Creative Commons license and free for use under certain usage agreements. In the future, we will host a web-based leaderboard, and release the easy-to-run tools to the community to better reduce the divergence.

\noindent \textbf{Robust Human-centric Matting Refinement.}
In the annotation pipeline, we use Grab-cut~\cite{rother2004grabcut} to refine bad results of CNN-based methods, yet it is still not perfect. Since per-frame human labeling is impractical due to the volume of captured data, we involve human checks over segmentation results, and only cases without major artifacts in the whole sequence across views will be released.  We tried 3D methods to further refine results, \textit{e.g}., using \ngp~\cite{muller2022instant} to train with valid views and infer the bad views. However, the results are not appealing enough (blurry edges). We will further investigate more robust tools. These challenges could also benefit matting research. We believe with the development of relevant techniques, matting robustness will be improved in the near future.

\noindent \textbf{New benchmarks.}
In this paper, we set up three task benchmarks as a kick-off of the DNA-Rendering dataset. Our datasets could potentially be used in many other tasks related to human-centric rendering, such as garment modeling/animation and human shape completion. We encourage and welcome the community to join us to unlock more downstream tasks.  

\balance

\end{document}